\documentclass{article}
\usepackage{graphicx} 

\usepackage[letterpaper,top=2cm,bottom=2cm,left=2.5cm,right=2.5cm,marginparwidth=1.75cm]{geometry}
\usepackage{authblk}
\usepackage{amsmath,amssymb, bm,amsthm}
\usepackage{graphicx}
\usepackage{courier}
\usepackage{comment}
\usepackage{color}
\usepackage{algpseudocode}
\usepackage{algorithm}

\newcommand \HardwareModel{A~{\it RC-Spike-Equivalent Hardware models}}
\newcommand \Proof{B~{\it Proofs}}
\newcommand \TTFS{C~{\it Simulation results for TTFS-SNNs}}
\newcommand \SpikeNoise{D~{\it Effects of spike timing noise on RC-Spike models}}
\newcommand \StepsInTestPhase{E~{\it Effects of DSTD steps in test phase}}
\newcommand \WeightScaling{F~{\it Weight scaling optimization in test phase}}
\newcommand \CircuitDetail{G~{\it Circuit details}}

\title{Harnessing Nonidealities in Analog In-Memory Computing Circuits: A Physical Modeling Approach for Neuromorphic Systems}
\author[1,2]{Yusuke Sakemi}
\author[2,3]{Yuji Okamoto}
\author[4]{Takashi Morie}
\author[1,5,6]{Sou Nobukawa}
\author[7]{Takeo Hosomi} 
\author[1,2]{Kazuyuki Aihara}

\affil[1]{Research Center for Mathematical Engineering, Chiba Institute of Technology, Narashino, Japan}
\affil[2]{International Research Center for Neurointelligence (WPI-IRCN), The University of Tokyo, Tokyo, Japan}
\affil[3]{Graduate School of Medicine, Kyoto University, Kyoto, Japan}
\affil[4]{Graduate School of Life Science and Systems Engineering, Kyushu Institute of Technology, Kitakyushu, Japan}
\affil[5]{Department of Computer Science, Chiba Institute of Technology, Narashino, Japan}
\affil[6]{Department of Preventive Intervention for Psychiatric Disorders, National Institute of Mental Health, National Center of Neurology and Psychiatry, Tokyo, Japan}
\affil[7]{NEC Corporation, Kawasaki, Japan}

\date{\today}

\begin{document}

\maketitle

\begin{abstract}
Large-scale deep learning models are increasingly constrained by their immense energy consumption, limiting their scalability and applicability for edge intelligence.  
In-memory computing (IMC) offers a promising solution by addressing the von Neumann bottleneck inherent in traditional deep learning accelerators, significantly reducing energy consumption. 
However, the analog nature of IMC introduces hardware nonidealities that degrade model performance and reliability. 
This paper presents a novel approach to directly train physical models of IMC, formulated as ordinary-differential-equation (ODE)-based physical neural networks (PNNs). 
To enable the training of large-scale networks, we propose a technique called differentiable spike-time discretization (DSTD), which reduces the computational cost of ODE-based PNNs by up to 20 times in speed and 100 times in memory. 
We demonstrate that such large-scale networks enhance the learning performance by exploiting hardware nonidealities on the CIFAR-10 dataset. 
The proposed bottom-up methodology is validated through the post-layout SPICE simulations on the IMC circuit with nonideal characteristics using the sky130 process. 
The proposed PNN approach reduces the discrepancy between the model behavior and circuit dynamics by at least an order of magnitude. 
This work paves the way for leveraging nonideal physical devices, such as non-volatile resistive memories, for energy-efficient deep learning applications. 
\end{abstract}

\section{Introduction}
Deep learning is a state-of-the-art methodology in numerous domains, including image recognition, natural language processing, and data generation \cite{Dong2021survey}. 
The discovery of scaling laws in deep learning models \cite{Bahri2024explaining} has motivated the development of increasingly larger models, commonly referred to as foundation models \cite{Openai2024gpt4, Dubey2024llama3}. 
Recent studies have shown that reasoning tasks can be improved through iterative computations during the inference phase \cite{OpenAI_o1}. 
While computational power continues to be a major driver of artificial intelligence (AI) advancements, the associated costs remain a significant barrier to broader adoption across diverse industries \cite{Patterson2021carbon}. 
This issue is especially critical in edge AI systems, where energy consumption is constrained by the limited capacity of batteries, making the need for more efficient computation paramount \cite{Murshed2021machine}.

One promising strategy to enhance energy efficiency is fabricating dedicated hardware. 
Since matrix-vector multiplication is the computational core in deep learning, parallelization greatly enhances computational efficiency \cite{Krizhevsky2012imagenet}. 
Moreover, in data-driven applications such as deep learning, a substantial portion of power consumption is due to data movement between the processor and memory, commonly referred to as the von Neumann bottleneck \cite{Horowitz2014computing}. 
Consequently, a variety of hardware accelerators have been developed, including graphical processing units (GPUs) optimized for data centers \cite{Dally2023hardware}, tensor processing units (TPUs) \cite{Jouppi2017in}, and application-specific integrated circuits (ASICs) \cite{Sze2017}.

Among the various AI hardware architectures proposed to date, in-memory computing (IMC) has garnered significant attention for its potential to achieve the highest levels of energy efficiency. 
IMC eliminates the von Neumann bottleneck by performing computations directly within memory, thereby demonstrating superior power efficiency \cite{Duan2024memristor, Aguirre2024hardware}. 
With some exceptions \cite{Fujiwara2022fully}, IMC is typically implemented as an analog computing system that sums currents flowing through resistive memory in accordance with Kirchhoff’s laws. 
Various resistive memory technologies have been utilized to implement IMC, including static random access memory (SRAM) \cite{Valavi2019tile,  Wang2023charge, Yoshioka2024capacitor, Tagata2024double}, floating gate memory \cite{Bavandpour2019energy, Xiao2022accurate}, resistive random access memory (ReRAM) \cite{Wan2022compute, Dalgaty2024mosaic}, and phase-change memory (PCM) \cite{Gallo2023core}.

Despite its potential, designing IMC circuits to achieve both high energy efficiency and reliability remains challenging. 
Owing to the nonideal characteristics of analog circuits, discrepancies  may arise between software-trained artificial neural networks (ANNs) and their hardware implementations, leading to inaccurate inferences \cite{Roy2020inmemory}. 
The primary sources of these inaccuracies are process variation and nonlinearity in circuit characteristics. 
Process variation refers to the fact that even circuits fabricated from the same design data exhibit performance variations , sometimes substantially. 
Nonlinearity occurs even in the absence of process variation and represents a pure modeling error inherent to the  behavior of analog components.  
Nonlinearity becomes more pronounced in advanced semiconductor fabrication processes \cite{Razavi}.

These  traits are referred to as nonideal because AI hardware research typically follows a top-down approach, where a model is first defined, and the hardware is then developed to efficiently execute it.  
Although algorithm-hardware co-design is often adopted, primarily focusing on model downsizing techniques such as quantization, pruning, and distillation \cite{Krestinskaya2024neural}, 
complex behaviors inherent to analog hardware systems are not easily captured. 
If these complex behaviors could be incorporated directly into the models, 
the issue of nonideal characteristics may be addressed.  
The human brain, which outperforms current AI hardware in terms of energy efficiency, inherently embodies such a bottom-up approach. 
For instance, neurons and synapses exhibit complex, nonlinear physiological characteristics, and properties of individual neurons vary \cite{Byrne2014molecules}. 
Furthermore, as the brain develops, it dynamically adjusts its physiological characteristics, such as the excitation-inhibition (E-I) balance \cite{Ito2016gaba}, to create an efficient learning system. 

Neuromorphic engineering is one such bottom-up approach \cite{Frenkel2023bottom}. 
Neuromorphic engineering focuses on mimicking biological neural systems, such as implementing silicon retinas by leveraging the dynamics of metal-oxide-semiconductor field-effect transistors (MOSFETs) in their subthreshold regions \cite{Frenkel2023bottom}. 
However, they typically employed biologically plausible learning rules, such as Hebbian learning. 
As a consequence, they were unable to rival the high performance of modern deep learning models empowered by backpropagation algorithms.

Another emerging bottom-up approach is physics-aware training (PAT) \cite{Wright2022deep}. 
PAT enables the training of physical neural networks (PNNs) \cite{Momeni2024training} by modeling any physical system, such as analog circuits or optical systems, as a deep neural network. 
However, applying PAT to IMC circuits is challenging owing to the significant data requirements needed to model input--output relations when adjusting weights. 
Further optimization of IMC circuits would require even more extensive data, rendering this method impractical for IMC. 
Therefore, to design IMC circuits using a bottom-up approach, a more efficient modeling and training framework is needed.

In this study, we address the nonlinearity problem by modeling the IMC circuits as PNNs that precisely capture the complex analog dynamics of an IMC circuit. 
Unlike PAT, our PNNs are modeled in the form of ordinary differential equations (ODEs), 
and each neuron in the PNN is formulated as a spiking neuron with a reversal potential. 
However, because these PNNs are based on ODEs, the training process is inefficient, scaling with the square of the input dimension, which makes it difficult to apply to large-scale network architectures. 
To overcome this challenge, we introduce differentiable spike-time discretization (DSTD), which significantly improves computational speed and memory efficiency by 10--100 times and demonstrates the ability to train convolutional neural networks on the CIFAR-10 dataset \cite{CIFAR10}. 
We also show that this method can be applied to spiking neural networks (SNNs) based on time-to-first-spike (TTFS) coding.

Finally, to demonstrate the utility of this bottom-up design approach, we designed an IMC circuit with nonideal characteristics using the sky130 process \cite{sky130}. 
Post-layout simulations show that the modeling error from this bottom-up approach is less than one-tenth of that from the conventional top-down approach.
\section{Results}

\begin{figure*}
\centering
\includegraphics[clip, width=\textwidth]{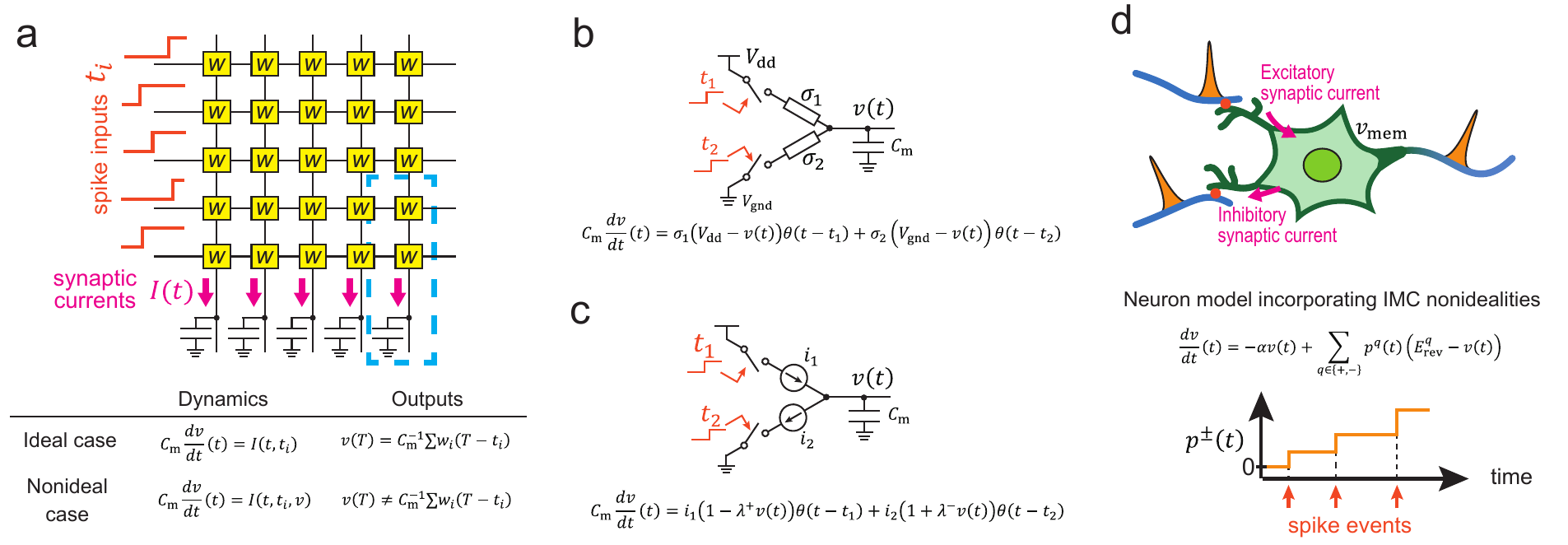}
\caption{{\bf Comparison of the dynamics between IMC circuits and biological neurons } 
{\bf a.} Schematic of charge-domain IMC Circuits. 
Input signals are delivered through horizontal lines in the form of spikes. 
Upon receiving these spikes, synaptic currents are induced along the vertical lines due to interactions between the spike signals and the memory elements, denoted as `W' in the figure. 
The currents are integrated and converted into voltages at capacitors. 
Circuit examples within the regions outlined by dashed blue lines are illustrated in panels {\bf b} and {\bf c}.
{\bf b.} Dynamics of IMC circuits with resistors and transistor switches. 
At time $t_1$, a transistor switch is activated, allowing current to flow through a resistor with conductance $\sigma_1$. 
Subsequently, at time $t_2$, another transistor switch is turned ON, similarly permits current to pass through a resistor of conductance $\sigma_2$. 
The resulting current induces a change in the voltage across the capacitor, $v(t)$. 
According to Ohm's law, the current magnitude depends on the capacitor voltage $v(t)$.  
{\bf c.} Dynamics of IMC circuits with current sources and transistor switches. 
While an ideal current source provides a constant current that is independent of the capacitor voltage $v(t)$, real-world current sources exhibit a behavior in which the current varies linearly with the capacitor voltage. 
This nonideal behavior can be characterized by the parameter $\lambda^\pm$.
{\bf d.} Dynamics of membrane potentials in IMC-aware neuron models. 
The net activity of receptors for excitatory current $p^+(t)$ and inhibitory current $p^-(t)$ exhibits stepwise changes in response to incoming spikes.  
Synaptic currents are governed by Ohm's law, incorporating reversal potentials  $E_\text{rev}^+$ and $E_{\text{rev}}^-$. 
This neuron model, referred to as the IMC-aware neuron model, captures the dynamics observed in the IMC circuits described in \textbf{a} and \textbf{b}. 
}
\label{fig:Equivalent_models}
\end{figure*}

The IMC circuit is structured as a crossbar array, as depicted in Fig. \ref{fig:Equivalent_models} {\bf a}. 
In this configuration, the input signals propagate horizontally along the axon or word lines, while the current flows vertically through the dendrite or bit lines \cite{Roy2020inmemory}. 
The magnitude of the current is determined by the state of the resistive memory elements, denoted as `W'. 
The matrix-vector multiplication (MVM) operations within the crossbar array can be executed through various approaches, which are broadly classified into the current domain and the charge domain \cite{Valavi2019tile}.
In current-domain IMC, computations are performed by summing currents, while in charge-domain IMC, the current is stored in a capacitor, where the result is held in the form of an electric charge. 
By reducing the amount of current needed for calculations, charge-domain IMC typically achieves higher energy efficiency \cite{Cai2019fully, Yamaguchi2021energy, Bavandpour2019energy, Valavi2019tile, Saito2020igzo}. 
Additionally, its dynamics resembles those of the biological neurons, which accumulate synaptic currents in the membrane capacitance. 
Consequently, this study focuses on the charge-domain IMC owing to its potential advantages.

Figure \ref{fig:Equivalent_models} {\bf b} illustrates the analog dynamics of an IMC circuit when using a synapse composed of a resistor and transistor switch. 
This configuration is known as a 1-transistor-1-resistor (1T1R) memory topology \cite{Hayakawa2015highly, Ankit2019puma}. 
When a spike signal is input, the switch turns ON, allowing current to flow. 
The amount of current is controlled by the conductance of the resistor, denoted as  $\sigma_i$, which corresponds to the weight of the network. 
The current alters the voltage across the capacitor, $v(t)$. 
According to Ohm's law, the current flowing through the resistor depends on the capacitor's voltage. 
A similar dynamic behavior can also be observed in a 1-diode-1-resistor (1D1R) topology \cite{Liu2014ReRAM, Zhou2021high}.
Fig. \ref{fig:Equivalent_models} {\bf c} represents an IMC circuit configured with current sources as synapses. 
Although an ideal current source does not depend on the capacitor voltage $V_\text{mem}$, actual current sources exhibit some dependency on this voltage \cite{Wang2018time,Bavandpour2019energy}. 
This dependency can be approximated linearly with the introduction of factors $\lambda^\pm$. 
In a later section, we present the circuit design and simulations based on this topology. 

Biological neurons receive spike signals from other neurons via synapses, as shown in Fig. \ref{fig:Equivalent_models} {\bf d}. 
A leaky integrate-and-fire (LIF) model is well-known for reproducing the dynamics of biological neurons in a computationally efficient manner \cite{Gerstner2014neuronal}. 
When a spike signal arrives at a synapse, it prompts the release of neurotransmitters from the presynaptic neuron, which are then detected by receptors on the postsynaptic neuron. 
This interaction opens ion channels on the postsynaptic neuron, resulting in the generation of an ionic current.
The magnitude of this current is proportional to the difference between the membrane potential and reversal potential for specific ions, which is determined by the ion concentrations inside and outside a cell \cite{Byrne2014molecules}. 
In other words, the current generated by receiving spike signals can be described in terms of two factors: one dependent on the opening and closing of ion channels and the other on the membrane potential.

We introduce the IMC-aware neuron model, which simplifies the nonlinear dynamics of ion channels using stepwise functions. 
Crucially, the dynamics of the IMC-aware neuron model is equivalent to that of the circuit depicted in Fig. \ref{fig:Equivalent_models} {\bf b} and {\bf c}. 
For further details on this equivalence, refer to SI. \HardwareModel.

The reversal potential $E_\text{rev}^\pm$ can be considered as a nonideal characteristic in conventional IMC circuit design 
because when $|E_\text{rev}^\pm| \rightarrow \infty$, the voltage after a period of current accumulation becomes a weighted sum of the inputs, which is core computation of ANNs.  
Conversely, when $|E_\text{rev}^\pm|$ is finite, the results deviate. 
To mitigate this effect, various methods are employed, such as using operational amplifiers to set $v(t)$ as virtual ground \cite{Cai2019fully, Xiao2022accurate} or reducing the amplitude of the capacitor voltage \cite{Wang2018time, Bavandpour2019energy}. 
However, these solutions often involve trade-offs, such as increased circuit area and reduced power efficiency, and fail to provide a fundamental resolution.

To overcome this issue, we model the complex analog dynamics using an ODE-based PNN by introducing an IMC-aware neuron model, which incorporates the effects of $E_\text{rev}^\pm$. 
Because the PNNs are ODE systems, 
its training is computationally inefficient. 
To address this, we introduce DSTD, which drastically enhances the training speed. 
Finally, we designed an IMC circuit using the sky130 process \cite{sky130} and show that the PNN approach significantly reduces modeling errors compared to traditional mapping approaches.

\subsection{Fast training by differentiable spike-time discretization}

In this study, we investigate the characteristics of deep neural networks where each neuron is modeled as an IMC-aware neuron model.  
The time evolution of the IMC-aware neuron model is described as follows: 
\begin{align}
\frac{d}{dt}v(t) &= - \alpha v(t) + \sum _{q\in \{+,-\}}  p^q (t) \left( E_\text{rev}^q - v(t) \right)   \\
&= - \left( \alpha  + \sum _{q\in\{+,-\}}  p^q(t) \right) v(t) + \sum _{q\in \{+,-\}}  p^q(t) E_\text{rev}^q
\end{align}
where $\alpha$ is the leak constant of the neuron.  
$E_\text{rev}^+$ and $E_\text{rev}^-$ are the reversal potentials associated with the excitatory (positive) and inhibitory (negative) synapses, respectively. 
Synaptic currents flow depending on the activity of the channel gates, represented as $p^+(t)$ for excitatory currents and $p^-(t)$ for inhibitory currents. 
This activity exhibits a stepwise behavior as a function of the input spike timing, 
corresponding to synaptic currents that lack decay.  
Consider a scenario where $N$ input spikes are given to this neuron within a normalized time interval $[0,1]$. 
The times at which these spikes occur are $T_0 < T_1  < \dots, T_{M-1} < T_M=1$, and some spikes may coincide.
In this case, the membrane potential at $t=1$ can be calculated analytically (see Methods section):
\begin{align}
v(1) = v(0) + \sum _{i=0}^{M-1} \frac{g(t_i)}{f(t_i)} \left(Z_i - Z_{i-1}\right), \label{eq:neuron_general}
\end{align}
where we defined the following variables
\begin{align}
&Z_i = e^{-\sum _{j=k}^{M} f(t_j)\left(t_{j+1} - t_j\right)}, \\
&f_i = \alpha + \sum_{q\in\{+,-\}} p^q(t_i),~
g_i = \sum_{q\in\{+,-\}} p^q(t_i)E_\text{rev}^q. 
\end{align}
When $E_\text{rev}^+\rightarrow \infty$ and $E_\text{rev}^- \rightarrow - \infty$, $v(1)$ converges to a weighted sum of input spike times $t_i$, which is the ideal case for charge-domain computing \cite{Bavandpour2019energy} (see Methods section). 
However, in practical cases where these conditions are not met, Eq. (\ref{eq:neuron_general}) is complex, 
which costs $\mathcal{O}(MN)$ of computations because $p^\pm(t)$ is a summation of input spikes. 
Here, $\mathcal{O}$ represents the big O notation. 
Technically, different spike times in continuous time do not overlap, which leads to a computational complexity of $\mathcal{O}(N^2)$. 
This is much larger than the $\mathcal{O}(N)$ complexity of a neuron model in ANNs; consequently, scaling the model to larger sizes is difficult \cite{Sakemi2022spiking}. 
To address this issue, we propose DSTD as follows.

DSTD combines the benefits of continuous-time systems, which can differentiate spike timings \cite{Mostafa2018supervised}, with the computational efficiency of discrete-time systems \cite{Kheradpisheh2019s4nn}. 
Let the spike time from neuron $j$ in layer $l-1$ be denoted as $t_j^{(l-1)}$. 
Note that we limit the number of spikes produced by a neuron to one.  
During processing in layer $l$, the spike time representation is approximated and converted into a discrete-time representation at the discrete time points $\{T^{(l)}_m \}_{m=1,2,\dots, M}$, as follows:
\begin{flalign}
\{s_{j0}^{(l-1)}, s_{j1}^{(l-1)}, \dots, s_{jM}^{(l-1)} \} = \text{DSTD} \left( t_j^{(l-1)}\right) \label{eq:discrete_spikes}
\end{flalign}
Here, $s_{jm}^{(l-1)}$ is a spike variable that indicates whether the spike from neuron $j$ in layer $l-1$ occurred at time $T^{(l)}_m$. 
Owing to this approximation, the input spike can only occur at one of the $M$ discrete time points, which reduces the required memory for membrane potential calculations from $\mathcal{O}(N^2)$ to $\mathcal{O}(NM)$, as discussed earlier.
Importantly, the discrete-time spike variable $s_{jm}^{(l-1)}$ is not a typical all-or-none value but can take a real value within $[0,1]$, making gradient calculation possible.

\begin{figure*}
\centering
\includegraphics[clip, width=13cm]{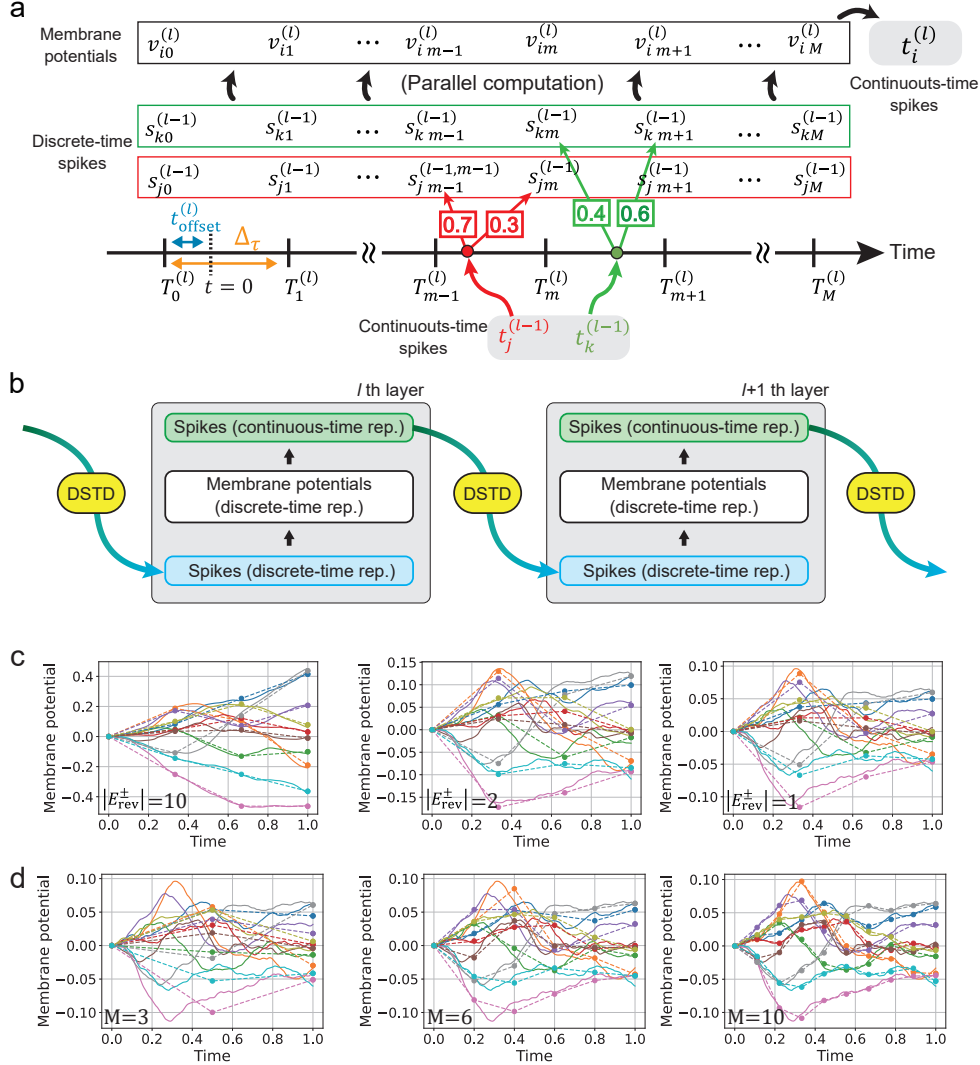}
\caption{{\bf Computation with differentiable spike-time discretization (DSTD)}. 
{\bf a.} Illustration of calculating the firing time of a neuron in the $l$th layer when spikes are input from the $j$th and $k$th neurons in the preceding $l-1$th layer at times $t_j^{(l-1)}$ and $t_k^{(l-1)}$, respectively, using DSTD. 
First, the discrete time points $T^{(l)}_m$ are determined based on an offset time $t_\text{offset}^{(l)}$ and a time interval $\Delta _\tau$. 
The discrete spike variables $s_{im}^{(l)}$ at these time points are calculated from the continuous-time spikes using DSTD. 
Using these discrete-time spikes, the membrane potential at each discrete time point is computed via an analytical solution. 
Based on this membrane potential, the firing time of the neuron is computed.
{\bf b.} The computational method using DSTD for a multilayer model. 
The output of each layer is a continuous-time spike signal, which is then converted into a discrete-time spike signal by DSTD. 
This discrete spike signal serves as input to the next layer. 
{\bf c.} Examples of the time evolution of the membrane potentials in a single-layer network without learning. 
The input consists of 1000 random spikes generated from a uniform distribution over the interval $[0, 1]$. The weights are  in a random initial state. The results for three different reversal potentials ($|E_\text{rev}^\pm|=10,~2,$ and $1$) are shown in separate graphs. 
In each figure, the solid line represents the exact trajectory of the membrane potential, while the dashed line 
 and plotted symbols represent the approximate membrane potential calculated using DSTD with steps $M=4$ when $t_\text{offset}$ is set to 0.
{\bf d.} Same as {\bf c}, but the membrane potentials are plotted for different values of $M$ in different graphs. 
The $|E_\text{rev}^\pm|$ is set to 1. 
}
\label{fig:proposed}
\end{figure*}

Figure \ref{fig:proposed} {\bf a} illustrates the processing when two spike inputs, $t_j^{(l-1)},~t_k^{(l-1)}$, are present in the $l$th layer. 
$t_\text{offset}^{(l)}$ denotes the difference between $t=0$ and $T_0^{(l)}$, which plays an important role in the subsequent sections. 
In this scenario, because the spike $t_j^{(l-1)}$ is received within the time interval $[T^{(l)}_{m-1}, T^{(l)}_m]$, 
only the spike variables $s_{j~m-1}^{(l-1)}$ and $s_{jm}^{(l-1)}$,  corresponding to time $T^{(l)}_{m-1}$ and $T^{(l)}_m$, respectively, take non-zero values. 
Similarly, because $t_k^{(l-1)}$ exists within the time interval $[T^{(l)}_m, T^{(l)}_{m+1}]$, 
only the spike variables $s_{km}^{(l-1)}$ and $s_{k~m+1}^{(l-1)}$, corresponding to time $T^{(l)}_m$ and $T^{(l)}_{m+1}$, respectively, take non-zero values. 
From these discrete-time spike variables, the membrane potential $v_{im}^{(l)}$ at each discrete time step can be computed in parallel. 
Subsequently, the firing time $t_i^{(l)}$ of the output spike in continuous time is calculated based on these membrane potentials. 
Figure \ref{fig:proposed} {\bf b} illustrates the computational flow for a multilayer network. 
Continuous-time spikes, either from the input layer or from each hidden layer, are transformed into discrete-time representations via DSTD. 
Upon receiving the discrete-time spikes, each layer computes the membrane potential at discrete time steps and, based on that, produces output spikes in continuous-time representation. 
As all computations are differentiable, the model is trainable using the backpropagation algorithm \cite{Baydin2018automatic}. 

Figure \ref{fig:proposed} {\bf c} presents the time evolution of the membrane potentials for varying values of the reversal potentials $E_\text{rev}^\pm$. 
This evolution results from the input of 1000 uniformly distributed random spikes over the time interval [0,1], with each spike assigned a random weight. 
The solid line in the figure represents the exact time evolution, while the dashed line illustrates the approximate time evolution obtained with DSTD steps of four ($M=4$). 
Because  $E_\text{rev}^+$ and $E_\text{rev}^-$ serve as the upper and lower bounds of the membrane potential, respectively, 
the range of the membrane potentials at time $1$ narrows as $|E_\text{rev}^\pm|$ decreases. 
Note that as $|E_\text{rev}^\pm|$ decreases, the approximation error in the DSTD method increases. 
Nevertheless, the DSTD method achieves reasonable accuracy even when the number of DSTD steps is four.
Figure \ref{fig:proposed} {\bf d} compares the exact temporal evolution of the membrane potentials (solid lines) for $E_\text{rev}^+=1$ and $E_\text{rev}^-=-1$ with the approximate evolution obtained for different values of DSTD steps $M$ (dashed lines). 
For $M=3$, although there are minor discrepancies in the final membrane potential, the overall agreement is reasonable. 
At $M=10$, the error is reduced to a negligible level.

\begin{figure*}
\centering
\includegraphics[clip, width=11cm]{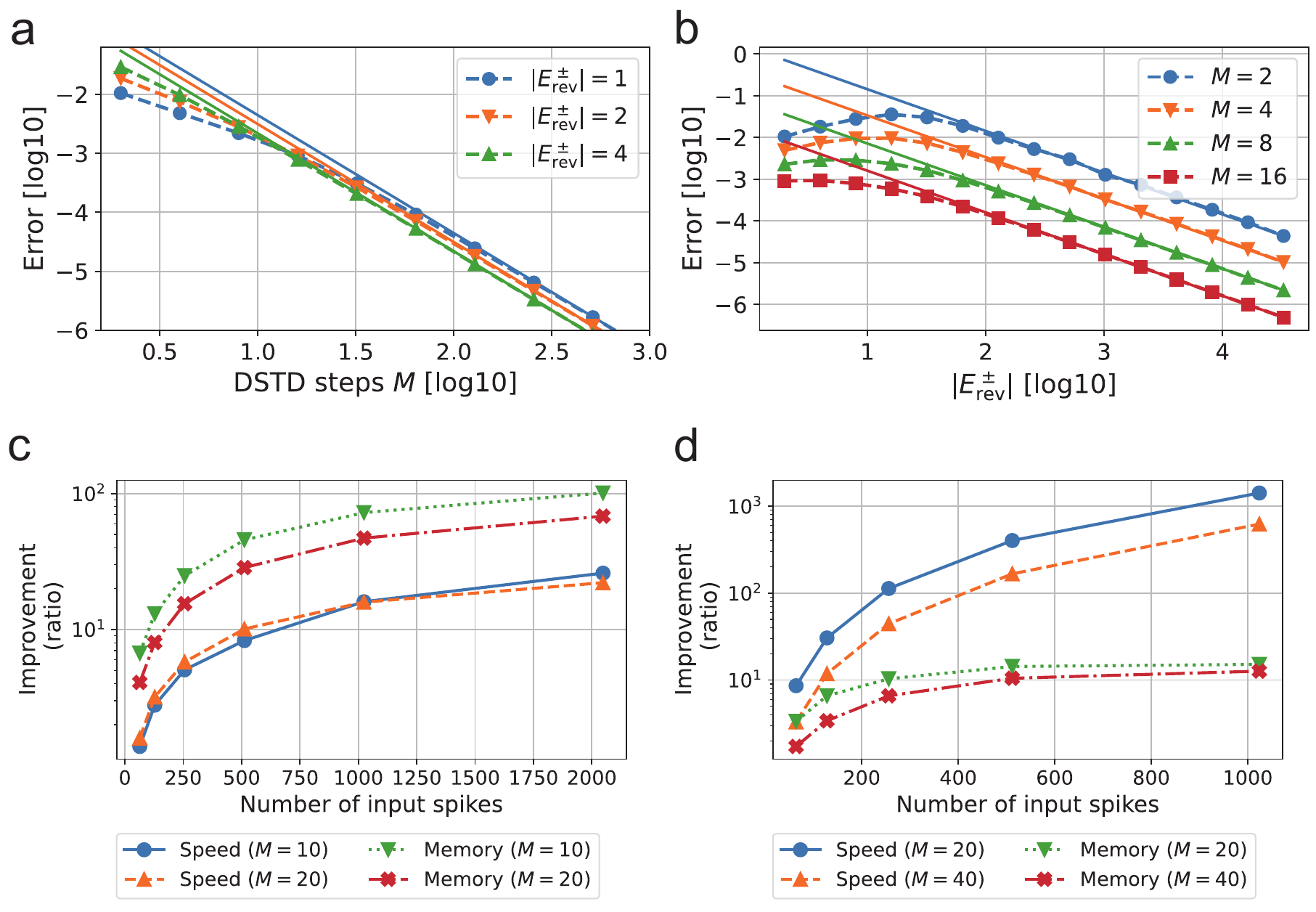}
\caption{{\bf Basic properties of DSTD.}    
{\bf a, b.} Errors between the membrane potential values at time 1 when using the exact solution and the approximated solution with DSTD. 
The membrane potential is obtained from a single-layer, untrained network consisting of 10 neurons, with spike inputs. 
The data consist of 1000 samples, with each sample containing 1000 input spikes. 
The input spike times are uniformly distributed over the interval $[0,1]$, and the synaptic weights are randomly assigned. 
In {\bf a.}, a double logarithmic plot illustrates how the error decreases as the number of DSTD steps $M$ increases, for various values of 
$|E_\text{rev}^\pm|$ (with $E_\text{rev}^+ = - E_\text{rev}^-$). 
In {\bf b.}, the error is plotted as a function of $|E_\text{rev}^\pm|$ (with $E_\text{rev}^+ = - E_\text{rev}^-$), for different numbers of DSTD steps $M$. 
In both {\bf a} and {\bf b}, dashed lines denote experimental results, while solid lines denotes theoretical predictions.    
{\bf c, d.} These figures depict the computational efficiency for the case of RC-Spike models ({\bf c}) and TTFS-SNN models ({\bf d}) when the number of input spikes is varied for a single-layer network consisting of 1000 neurons. The dataset consists of 1000 samples, with a batch size of 100. The input spike times for each sample are uniformly distributed over the interval $[0,1]$. 
}
\label{fig:DSR_properties}
\end{figure*}

Figures \ref{fig:DSR_properties} {\bf a} and {\bf b} illustrate the convergence of the error $\delta _\text{error} = \langle (|v_i^d(1) - \hat{v}_i^d(1)| \rangle _{i,d}$ as the number of DSTD steps $M$ increased and as $|E_\text{rev}^\pm|$ became larger. 
Here, $v_i^d(1)$ represents the exact membrane potential of the $i$th neuron for the $d$th data point at time $1$, 
and $\hat{v}_i^d(1))$ is the corresponding approximation obtained with DSTD at time $1$. 
The notation $\langle \cdot \rangle _{i,d}$ denotes averaging over neurons and data points. The input data consist of spike trains of dimension 1000, uniformly distributed in the range $[0,1]$, with 1000 data samples. 
Under various conditions, an increase in the number of DSTD steps $M$ and a large $E_\text{rev}^+ (=-E_\text{rev}^-)$ resulted in a uniform error reduction. 
This behavior can be mathematically expressed as $ |v(1) - \hat{v}(1)| = \mathcal{O}(M^{-2} |E_\text{rev}^\pm|^{-1})$. 
We provide a proof of this property in SI. \Proof. 
For sufficiently large values of $M$ and $|E_\text{rev}^\pm|$, 
 the numerical results aligned closely with the theoretical predictions.

The use of DSTD significantly reduces memory usage and computation time in the presence of reversal potentials. 
In Fig. \ref{fig:DSR_properties} {\bf c},  we compare the computational performance, in terms of time and peak GPU memory usage, of a single-layer network with and without the DSTD method. 
This network can be viewed as a single-layer version of the RC-Spike model \cite{Sakemi2022spiking}, which incorporates reversal potentials into ANNs (see Fig. \ref{fig:RC_Spike_VGG7} {\bf a} and Methods section for detail). 
Experiments were conducted on an NVIDIA V100 SXM2 GPU (16 GB HBM2) using the PyTorch framework \cite{pytorch}.
The figure shows the training time per epoch and peak GPU memory consumption, with a network comprising 1000 neurons and a dataset with 1000 samples, utilizing a batch size of 100. 
The results reflect variations in the number of input spikes, where spike timings were uniformly distributed within the time interval $[0,1]$. 
The performance improvements are shown in Fig. \ref{fig:DSR_properties} {\bf d}. 
When the number of DSTD steps is 10, the peak memory usage was reduced by  a factor of 100, while the computation time was reduced by a factor of 30. 
Figures \ref{fig:DSR_properties} {\bf d} depict the results for the TTFS-SNN model. 
TTFS-SNN is a type of SNNs in which neurons encode information using a TTFS coding scheme, firing at most once per neuron \cite{Bohte2002error, Mostafa2018supervised}. 
In TTFS-SNN models, computation is more complex than RC-Spike models owing to the firing events triggered upon reaching a threshold value  (see Fig. \ref{fig:RC_Spike_VGG7} {\bf a} and Methods section for detail). 
For a DSTD step count of 20, 
the peak memory usage was reduced by a factor of 10, and the computation time by a factor of 1000.

\subsection{Reversal potentials are not nonidealities}

\begin{figure*}
\centering
\includegraphics[clip, width=\textwidth]{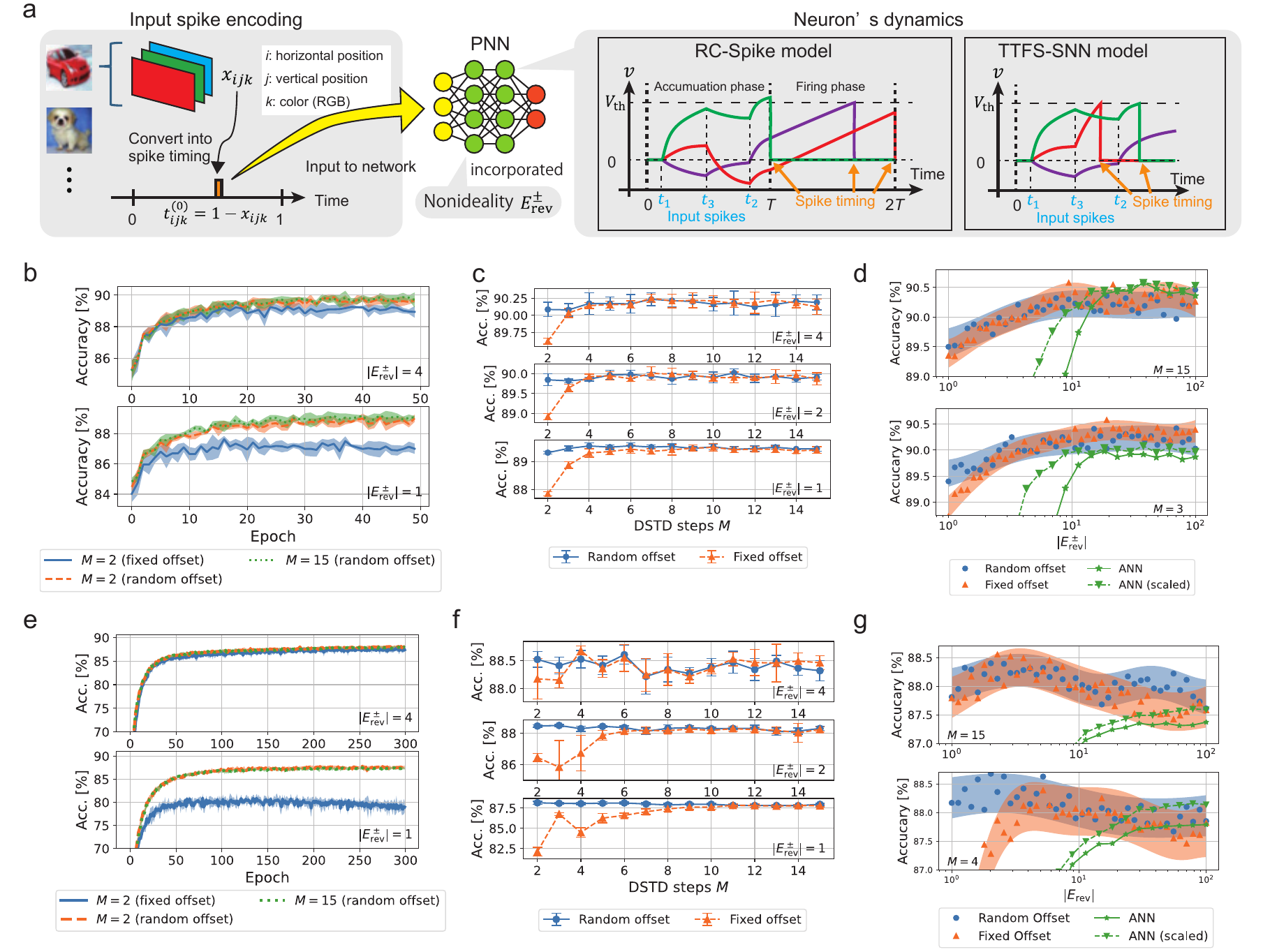}
\caption{{\bf Learning results for RC-Spike models using DSTD.} 
{\bf a.} The experimental setup. Each data point of a dataset is represented as a three-dimensional tensor $x_{ijk}$. 
These features are converted into single input spikes, $t_{ijk}^{(0)}=1-x_{ijk}$, 
which are processed by a PNN, which incorporates IMC nonidealities $E_\text{rev}^\pm$. 
We employed the RC-Spike model as a PNN. 
The RC-Spike model operates in two phases: the accumulation phase and  firing phase. During the accumulation phase, the neuron membrane potential evolves over time as it receives spikes from the preceding layer. In this example, three input spikes arrive at times 
$t_1$, $t_2$, and $t_3$. 
Between receiving consecutive spikes, the membrane potential follows an exponential decay curve (RC-decay), influenced by the reversal potential. 
In the firing phase, the membrane potential increases linearly, and a spike is generated when it exceeds the firing threshold. 
Similar experiments are conducted for the case of TTFS-SNN models, and the results are summarized in {\it SI.\TTFS Simulation results for TTFS-SNNs}. 
{\bf b-d.} Experimental results for fully connected RC-Spike models (784-400-400-10) on the Fashion-MNIST dataset. 
In {\bf b}, learning curves for $|E_\text{rev}^\pm| = 4$ (upper) and $|E_\text{rev}^\pm| = 1$ are shown.  
The error bars represents the standard deviation across five networks with different initial weights. 
Each panel shows the results for different values of DSTD steps $M$. 
Additionally, we show the cases where the offset time $t_\text{offset}$ was randomized (random offset) and fixed (fixed offset) for each mini-batch. 
In {\bf c}, accuracies obtained from training the model with different DSTD steps $M$ are shown. 
For each $M$, results for $|E_\text{rev}^\pm|=4,~2,$ and $1$ are presented from top to bottom. 
Each panel compares the model performance with and without the introduction of a random offset. 
Each data point represents the mean and standard deviation of the training results from five models, each initialized with different random weights.
In {\bf d}, recognition accuracies of models trained with 15 DSTD steps (upper) and 3 DSTD steps (lower) as a function of $|E_\text{rev}^\pm|$. 
In each panel, the effects of introducing a random offset are compared. 
The 95\% confidence intervals are estimated using Gaussian process regression. 
The green line represents the performance of a model trained with $|E_\text{rev}^\pm|=100$ (equivalent to an  ANN). 
The green dashed line represents the performance of the ANN case, 
but the positive and negative weights are optimally scaled for specific $|E_\text{rev}^\pm|$ values in the test phase. 
{\bf e-g}, Experimental results for convolutional RC-Spike models on the CIFAR-10 dataset. 
The experimental setup of {\bf e}, {\bf f}, and {\bf g} are same as {\bf b}, {\bf c}, and {\bf d}, respectively. 
}
\label{fig:RC_Spike_VGG7}
\end{figure*}

DSTD facilitates the training of large-scale neural networks in which neurons possess reversal potentials.   
These reversal potentials can be regarded as nonideal characteristics inherent to conventional charge-domain IMC circuits. 
Next, we investigate how the reversal potentials, $E_\text{rev}^\pm$, influence learning performance. 
Additionally, we examine the impact of the approximations introduced by DSTD on learning dynamics. 
Figure \ref{fig:RC_Spike_VGG7} {\bf a} illustrates the experimental setup.
We used standard image recognition datasets: Fashion-MNIST \cite{Xiao2017fashion} and CIFAR-10 \cite{CIFAR10}. 
For both datasets, each data point can be represented as a three-dimensional tensor $x_{ijk}$, 
where each element of a tensor is converted into an input spike and fed into the PNN for training. 
See Methods section for detail. 

We utilized the RC-Spike model as a PNN to directly assess the effect of $E_\text{rev}^\pm$.  
As depicted in Fig. \ref{fig:RC_Spike_VGG7} {\bf a}, the RC-Spike model operates in two temporal phases: 
an accumulation phase, where the membrane potential evolves in response to input spikes, and a firing phase, during which spikes are generated \cite{Sakemi2022spiking}. 
The membrane potential undergoes exponential decay between successive input spikes, a behavior driven by the reversal potentials. Notably, when $|E_\text{rev}^\pm|\rightarrow\infty$, the model's behavior converges to the ideal operation of charge-domain computing \cite{Bavandpour2019energy, Yamaguchi2021energy}, where the output of each layer corresponds precisely to a weighted sum with a hard sigmoid activation function. 
Consequently, the RC-Spike model continuously interpolates between ideal and nonideal behaviors of charge-domain computing through the parameter $E_\text{rev}^\pm$. 
For detailed mathematical formulations, refer to SI.\HardwareModel.  
A similar set of experiments was conducted with TTFS-SNN. 
In TTFS-SNN, a neuron fires at most once for each data.  
Because the spike timing carries information, TTFS-SNN is an important class of SNNs that realize temporal coding. The experimental results regarding TTFS-SNN are summarized in SI.\TTFS. 
In the following experiments, additive Gaussian noise with a mean of $0$ and a standard deviation of $\sigma_\text{spike}$ was applied to the output spikes, including during the inference phase. 
This is because, when the values of $|E_\text{rev}^\pm|$ are small, the network tends to minimize the membrane potential to avoid the influence of $|E_\text{rev}^\pm|$. 
However, from a circuit implementation perspective, this is undesirable as both membrane potentials and output spikes have finite precision. 
To consider this limited precision, spike noise was introduced. 
We set $\sigma _\text{spike}=0.01$ throughout this study. 
Refer to SI.\SpikeNoise~ for details on the results with various levels of spike noise. 
Furthermore, DSTD was also applied during the test phase. 
To ensure reliable evaluations, a large number of DSTD steps were utilized during the test phase, as detailed in SI.\StepsInTestPhase. 

The training results of a fully connected network with two hidden layers (784-400-400-10) on the Fashion-MNIST are shown in Figs. \ref{fig:RC_Spike_VGG7} {\bf b-d}. 
Figure \ref{fig:RC_Spike_VGG7} {\bf b} presents the learning curves of the training process of a fully connected network with two hidden layers (784-400-400-10) on the Fashion-MNIST dataset, which was performed using the Adam optimizer with a mini-batch size of 32 over 50 epochs (see Methods section). 
The errors represent the standard deviation across five independently initialized networks. 
The upper panel shows the results for $|E_\text{rev}^\pm|=4$, while the lower panel corresponds to $|E_\text{rev}^\pm|=1$. 
Each panel compares the results for DSTD steps $M=2$ and $M=15$. 
Additionally, two conditions were compared: one where the offset time for DSTD, $t_\text{offset}^{(l)}$ (see Fig. \ref{fig:proposed} {\bf a}), was randomized across mini-batches (random offset), and the other where it was fixed at $t_\text{offset}=0$ (fixed offset). 
In all the conditions, the learning process was stable. 
However, for $M=2$ under the fixed offset condition, recognition performance decreased significantly, particularly for $|E_\text{rev}^\pm|=1$. 
By contrast, even with $M=2$, the random offset condition achieved an accuracy comparable to that observed for $M=15$.

Figure \ref{fig:RC_Spike_VGG7} {\bf c} presents the recognition performance results when varying the number of DSTD steps, $M$. 
Each panel, from top to bottom, shows the results for $|E_\text{rev}^\pm|=4$, $|E_\text{rev}^\pm|=2$, and $|E_\text{rev}^\pm|=1$. 
In each panel, the results are shown for the case where the time offset $t_\text{offset}^{(l)}$ was randomized (random offset) and where $t_\text{offset}^{(l)}=0$ (fixed offset). 
The error bars for each plot represent the standard deviation across five models initialized with different random weights. 
In all cases, the recognition accuracy saturated when the number of DSTD steps, $M$, reached approximately 10. 
Furthermore, in the random offset condition,  performance saturated with fewer DSTD steps, which could be attributed to the reduction in the gradient bias caused by DSTD approximation. 
Notably, this effect was more pronounced when $|E_\text{rev}^\pm|$ was small, corresponding to the case of stronger nonideal characteristics. 

Figure \ref{fig:RC_Spike_VGG7} {\bf d} illustrates the change in recognition performance as $|E_\text{rev}^\pm|$ was varied. 
The upper panel shows results for DSTD steps of $M=15$, while the lower panel corresponds to $M=3$. 
In both the random offset and fixed offset cases, a decrease in $|E_\text{rev}^\pm|$, which reflects an increase in nonideal characteristics, resulted in performance degradation. 
However, even in the most extreme case where $|E_\text{rev}^\pm|=1$, the performance drop was at most 1 point. 
Moreover, in the random offset case, the performance degradation was mitigated compared to the fixed offset condition, particularly for DSTD steps $M=3$. 
We also compared the approach of modeling nonideal characteristics using ODE-based PNNs (proposed approach) with the conventional ANN-to-hardware mapping approach. 
The network trained with $|E_\text{rev}^\pm|=100$ represents a model that minimally incorporates nonideal characteristics, akin to an ANN. 
The inference results of this network under various $E_\text{rev}^\pm$ values are represented as green solid lines (referred to as ANN). 
Additionally, we present the results in which the positive and negative weights of the network were optimized during inference to incorporate the effect of the reversal potential (referred to as scaled ANN). 
Specifically, the values of a scaling vector $(\alpha^+, \alpha^- )$, where positive and negative trained weights are respectively scaled by  $\alpha^+$ and $\alpha^-$, are optimized for each value of $|E_\text{rev}^\pm|$ during the inference phase. 
The detail of the optimization procedure is described in SI.\WeightScaling.
Although the scaled ANN achieved higher recognition performance than the standard ANN, its performance degraded significantly when $|E_\text{rev}^\pm|<4$. 
This result indicates that nonideal characteristics modeled by the reversal potential are difficult to capture accurately using a conventional ANN.

We conducted similar experiments with a larger-scale convolutional network on the CIFAR-10 dataset \cite{CIFAR10}. 
The network architecture was similar to Ref. \cite{Simonyan2015very}:
Conv(64)-Conv(64)-PL-Conv(128)-Conv(128)-PL-Conv(256)-Conv(256)-PL-512-10, 
where Conv($N$) represents a convolution layer with a kernel size of $3\times3$, number of output channels $N$, and stride $1$, 
while PL represents a pooling layer with kernel size of $2\times 2$, stride $2$. 
The learning curve is presented in Fig. \ref{fig:RC_Spike_VGG7} {\bf e}. 
As shown, stable learning was achieved under various DSTD steps $M$ and different values of $|E_\text{rev}^\pm|$. 
Figure \ref{fig:RC_Spike_VGG7} {\bf f} illustrates the variation in recognition performance as $M$ was altered under different $|E_\text{rev}^\pm|$ conditions. 
Similar to the results in Fig. \ref{fig:RC_Spike_VGG7} {\bf c}, recognition performance was maximized when $M \gtrsim 10$. 
In the case of fixed offsets, recognition performance decreased when the number of DSTD steps was small. 
However, with the introduction of random offsets, maximum performance was achieved even at $M = 2$.

Figure  \ref{fig:RC_Spike_VGG7} {\bf g} shows the impact on recognition performance as $|E_\text{rev}^\pm|$ was varied, with DSTD steps set to $M = 4$ or $15$. 
As with the MLP results, recognition performance degraded when $|E_\text{rev}^\pm|$ was extremely small (approximately $1$ point), particularly when $M$ was small. 
However, this degradation was mitigated with the introduction of random offsets. 
Unlike the MLP network, which exhibited a monotonic increase in performance with increasing $|E_\text{rev}^\pm|$, the convolutional network demonstrated peak performance at specific $|E_\text{rev}^\pm|$ values.
This result indicates that reversal potentials $E_\text{rev}^\pm$, which have traditionally been considered nonidealities in circuit design, can be exploited to improve model performance if these characteristic are properly incorporated into models.

\subsection{Examination of PNN approach by Circuit Simulations}

As depicted in Fig. \ref{fig:Equivalent_models}, the effects of the reversal potential in neuron models corresponds to the nonideal characteristics observed in IMC circuits. 
In the previous section, we introduced a method for accelerating models that incorporate this nonideal characteristic (reversal potential $E_\text{rev}^\pm$) when computed on digital platforms such as GPUs. 
Through numerical experiments, we demonstrated that the proposed method prevents degradation in learning performance caused by these nonidealities and, in some cases, enhances learning performance.
In this section, we evaluate the efficacy of the proposed method in a post-layout circuit simulation. 
We designed an IMC circuit based on the architecture illustrated in Fig. \ref{fig:Equivalent_models} {\bf c}. 
Figure \ref{fig:CircuitOverview} {\bf a}  presents the layout of the designed circuit (RC-Spike circuit), which was designed using the open-source processing design kit (PDK) for the sky130 process \cite{sky130}. 

The circuit consists of two layers: a hidden layer and an output layer, both sharing an identical structure. 
Each layer includes synapse circuits and neuron circuits, which emulate five synapses and five neurons, respectively. 
Within each layer, the synapse circuit receives input spikes and delivers synaptic current to the neuron circuit. 
The neuron circuit accumulates the synaptic current in a capacitor, converting it into voltage, and subsequently generates spikes, which are transmitted to the synapse circuit of the subsequent layer (refer to the inset in Fig. \ref{fig:CircuitOverview} {\bf a}).
Figure \ref{fig:CircuitOverview} {\bf b} illustrates the overall operation of the circuit. 
The upper panel shows the waveform of the digital control signals applied to the circuit. 
These digital signals, provided externally, control the circuit's operation. 
The lower panel outlines the phase transitions within the circuit. 
Each layer operates cyclically, alternating among the reset, accumulation, and firing phases. 
Each phase persists for a duration of $T^\text{circ}$. 
Since the timing of these phases is shifted for each layer, the circuit operates in a pipelined manner, enabling efficient processing \cite{Yamaguchi2021energy, Bavandpour2019energy}.

Figure \ref{fig:CircuitOverview} {\bf c} presents the schematic diagram of the synapse circuit, illustrating a case with two inputs and two outputs. 
When a spike pulse $V_j^\text{spike}$ arrives (transitioning from 0 V to the supply voltage $V_\text{dd}$, with the opposite transition for $\overline{V_j^\text{spike}}$), the corresponding selector metal-oxide-semiconductor field-effect transistors (MOSFETs) (NS$_{ij}$ and PS$_{ij}$) are activated, allowing current to flow. 
The magnitude of this current is determined by the weight MOSFETs (NM$_{ij}$ and PM$_{ij}$), which operate as nonideal current sources.  
In this study, the current of the weight MOSFETs is controlled by applying a bias voltage externally. 
However, alternative memory technologies, such as SRAM \cite{Yamaguchi2021energy} or non-volatile memories such as floating-gate devices \cite{Bavandpour2019energy}, can also be used to set the current.
Figure \ref{fig:CircuitOverview} {\bf d} shows the schematic of the neuron circuit, which consists of input and output components. 
The input component charges a capacitor using the synaptic current, while the output component generates the spike signal. 
The neuron circuit operates in three distinct phases: the reset, accumulation, and firing phases. 
During the reset phase, $V_\text{reset}$ is activated, fixing the voltage across the membrane capacitor $C_\text{m}$ (the membrane potential) to the resting potential $V_0$. 
Simultaneously, node $V_\text{out}$ is grounded, bringing the output spike $V_i^\text{spike}$ to the ground potential. 
In the accumulation phase, both $V_\text{reset}$ and $V_\text{phase}$ are deactivated, allowing synaptic current to flow into (or out of) the capacitor, which is constructed with a metal-insulator-metal (MIM) structure. 
During the firing phase, $V_\text{reset}$ remains deactivated while $V_\text{phase}$ is activated. 
In this phase, the discharger circuit operates to lower the membrane potential. 
Simultaneously, the sensing inverter monitors the membrane potential, triggering an output spike when the potential falls below a certain threshold voltage. 
For further details on the circuit characteristics, refer to SI.\CircuitDetail.

To process data using the RC-Spike circuit, it is essential to precisely replicate the software-trained model on the hardware through a process known as {\it model-to-hardware mapping}. 
The mapping and simulation methodology is illustrated in Fig. \ref{fig:CircuitOverview} {\bf e}. 
The procedure for mapping an ANN model to hardware involved first training the model on the Iris dataset \cite{Fisher1936, scikit-learn}, followed by converting the trained model's weights into synaptic current values (the ANN-to-IMC mapping). 
When mapping the PNN model (in this case, we used a RC-Spike model) to hardware (the PNN-to-IMC mapping), the reversal potentials, $E_\text{rev}^\pm$, corresponding to the circuit's nonideal characteristics were first evaluated via circuit simulations. 
These reversal potentials were then used to construct the RC-Spike model. 
The remaining steps were the same as those used in ANN model mapping.
Note that the designed circuit exhibited significant nonideal characteristics, corresponding to reversal potentials of $E_\text{rev}^+ = 2.80$ and $E_\text{rev}^- = -1.53$. 
During the model-to-hardware mapping, a uniform scaling of the currents, corresponding to network weights, was required to mitigate the effects of parasitic capacitances present in the circuit. 
Specifically, the values of a scaling vector $(\alpha^+, \alpha^- )$, where positive and negative trained currents are respectively scaled by  $\alpha^+$ and $\alpha^-$, respectively, are needed to be  optimized for achieving the best results in SPICE simulation. 
In the case of the ANN-to-IMC mapping, both scaling elements $\alpha^+$ and $\alpha^-$ are separately scaled (2-dimensional scaling) to also mitigate the effects of circuit nonidealities, corresponding to reversal potentials. 
Conversely, in the PNN-to-IMC mapping, only a global scaling was applied (1-dimensional scaling, where $\alpha^+ = \alpha^-$),  to only account the effects of parasitic capacitance.
For a more detailed description of the mapping methods and the circuit and model parameters, refer to Methods section and SI.\WeightScaling.  

Figure \ref{fig:CircuitOverview} {\bf f} presents the results of the model and the circuit simulation for the ANN-to-IMC mapping case.  
In the hidden layer (upper panel) and output layer (lower panel), the reset, accumulation, and firing phases are masked in green, blue, and yellow, respectively. 
We set $T^\text{circ}=1~\mu$s.
The time evolution of the membrane potential for the mathematical model is depicted with an orange dashed line, while the firing time is represented in the firing phase by a reset of the membrane potential to the resting potential ($V_0=1.3$ V). 
The time evolution of the membrane potential in the circuit simulation is shown with a solid black line, and the spike signals are indicated by a blue dash-dotted line.
The input data correspond to the instance among 50 test data points where the root mean square error (RMSE) between the firing times in the output layer in the model and those in the circuit simulation was the largest. 
As seen in the figure, the firing times in the output layer were not consistent between the model and SPICE simulation results. 
This discrepancy highlights the ANN model' inability to fully capture the effects of the circuit's intrinsic nonidealities $E_\text{rev}^\pm$. 
Figure  \ref{fig:CircuitOverview} {\bf g} shows the results for the PNN-to-IMC mapping in a similar manner. 
Unlike the case with the ANN-to-IMC mapping, the results of the mathematical model and  SPICE simulation were almost entirely consistent.

Figure  \ref{fig:CircuitOverview} {\bf h} presents the distribution of the timing discrepancies between the firing times observed in the mathematical model and those obtained from SPICE simulations at the output layer. 
This histogram is based on 50 test samples from the Iris dataset. 
The upper and lower panels illustrate the results for the ANN-to-IMC mapping and PNN-to-IMC mapping, respectively. 
Both histograms exhibited a peak at zero, but the spread of the histogram is noticeably broader in the ANN-to-IMC mapping compared to that in the PNN-to-IMC mapping. 
The RMSE of the firing time differences was 39.04 ns for the ANN-to-IMC mapping and 1.97 ns for the SNN-to-IMC mapping, demonstrating a reduction in error by more than an order of magnitude. 
These results indicate that incorporating nonideal circuit characteristics, such as reversal potentials, into the model can reduce the modeling error substantially.

\section{Conclusion}

In this study, we incorporated the dynamics of IMC circuits with nonideal characteristics into a PNN model, represented as an ODE. 
Specifically, we showed that the nonideal property, wherein synaptic currents are dependent on the magnitude of the membrane potential, can be mathematically formulated using a neuron model that includes reversal potentials, $E_\text{rev}^\pm$. 
We called this model the RC-Spike \cite{Sakemi2022spiking}. 
As the RC-Spike model provides an analytical solution, it can be utilized for learning. 
However, the computational complexity is proportional to the square of the number of input spikes, posing a challenge for scaling the network. 
To mitigate this, we introduced DSTD, which successfully reduces the computational cost to linear scaling with respect to the number of input spikes, similar to conventional ANNs. 
This advancement enables the efficient training of large-scale neural networks.
Numerical simulations revealed that the DSTD method did not require a large number of discretization steps. 
Particularly, in the RC-Spike model, offset randomization of discrete time $t_\text{offset}$ for each mini-batch allowed effective learning with as few as four DSTD steps. 
In experiments using the convolutional RC-Spike model trained on the CIFAR-10 dataset, performance was maximized when $|E_\text{rev}^\pm|\sim 3$, a region where nonideal characteristics were present. 
These findings suggest that, by leveraging the analog dynamics as modeled in PNNs, circuit characteristics that were previously regarded as nonideal could be harnessed, turning them into an advantage for model performance.

DSTD is closely related to the learning algorithm of SNNs, as implied by the introduction of spike variables. 
Recent studies have shown that by adopting deep learning algorithms, particularly error backpropagation algorithms, multilayer SNNs can be trained efficiently \cite{Dampfhoffer2023backpropagation, Eshraghian2023training}. 
Successful backpropagation-based SNN training methods include the surrogate gradient method and timing-based methods. 
The surrogate gradient method is a technique that enables backpropagation and is particularly designed for discrete-time SNNs \cite{Neftci2019surrogate}.
In these discrete-time SNNs, binary spike variables indicate the occurrence of spikes at each discrete time step \cite{Wu2018spatio}. 
By approximating the small changes in spikes caused by membrane potential fluctuations with a suitable surrogate function, the entire network can be trained using backpropagation through time (BPTT) \cite{Neftci2017event, Huh2018gradient, Zenke2018super, Wu2018spatio}. 
The surrogate gradient method has demonstrated high performance in numerous studies \cite{Rathi2020enabling,   Kim2021revisiting, Yin2021accurate}. 
Although some studies have demonstrated the theoretical validity of the surrogate gradient \cite{Zenke2021remarkable, Gygax2024elucidating}, the computation of the gradient remains heuristic. 
Moreover, mapping discrete-time SNNs onto continuous-time analog hardware is not straightforward \cite{Cramer2022surrogate}. 
For instance, reducing discretization errors requires shortening the time step, which increases the number of steps needed for training and consequently escalates training costs.

The timing-based learning generally targets continuous-time SNNs and imposes the constraint that each neuron can fire at most once \cite{Bohte2002error}, refered to as TTFS coding.  
In cases where an analytical solution for spike timing can be obtained \cite{ Mostafa2018supervised, Comsa2020temporal, Goltz2021fast, Sakemi2023supervised}, error backpropagation becomes straightforward. 
This learning rule allows for exact gradient computation and implements temporal coding, an information processing mechanism based on spike timing. 
However, Owing to computational efficiency challenges, the timing-based learning has not been widely applied to large-scale networks, and its use in training models capable of solving modern benchmarks like CIFAR-10 remains limited \cite{Zhou2021temporal, Sakemi2023sparse}. 
Recently, extensions of the timing-based method that relax the constraint of single spike firing have been proposed \cite{Wunderlich2021event, Yamamoto2024can}. Additionally, Kheradpisheh et al. \cite{Kheradpisheh2019s4nn} implemented timing-based learning in discrete time, while Kim et al. \cite{Kim2020unifying} proposed a method that combines the surrogate gradient method with the timing-based learning in discrete-time systems. However, the gradient computation in these discrete-time SNNs remains heuristic. 

DSTD can be viewd as a hybrid method that combines the computational efficiency of discrete-time SNNs used in the surrogate gradient method with the precise gradient computation found in continuous-time SNNs employed by the timing-based method. 
While DSTD discretizes spikes in time, similar to the surrogate gradient method, it differs in that the spike variables are represented by real values instead of binary ones. 
These values depend on the original continuous spike timing, allowing for differentiation with respect to time. 
Additionally, by increasing the number of steps, DSTD can exactly match ODEs, making the relationship between approximate and exact gradients explicit (see SI.\Proof).

In this study, we developed a fast learning method by utilizing the analytical solutions of ODEs. 
ODE systems can also be trained using the adjoint method without relying on analytical solutions \cite{Chen2018neural, Huh2018gradient, Wunderlich2021event}. 
However, the adjoint method requires computationally expensive ODE solvers, making it difficult to train large-scale networks.
To solve ODEs efficiently, a well-known method exists that discretizes time with relatively coarse step sizes and connects them using analytical solutions \cite{Rotter1999exact}. 
However, this approach assumes that input spikes occur at specified time steps, leading to discrepancies with the exact solution when using coarse step sizes \cite{Rotter1999exact, Tian2020exponential}. Moreover, because it handles discrete time, it cannot capture gradients concerning spike timing displacements. 
The DSTD method enabled the use of this simulation technique as a training method by approximating spike variables as real numbers in discrete time. 
Furthermore, by randomizing the time step offsets in each mini-batch, we successfully reduced the impact of discretization errors on the loss function.

We designed and simulated a small-scale circuit to evaluate the effectiveness of the PNN approach in real hardware. 
The simulations compared the performance of mapping ANN models to hardware (the ANN-to-IMC mapping) with that of PNN models (the PNN-to-IMC mapping). 
The results demonstrated that the proposed method reduced the modeling error to less than one-tenth of that observed with the ANN-to-IMC mapping. 
These findings suggest that the adoption of PNN models that incorporate the reversal potentials omit the necessity of introducing complex circuits to avoid the effects of nonideal characteristics \cite{Cai2019fully, Xiao2022accurate, Xiao2022accurate, Wang2018time, Bavandpour2019energy}. 

We focused on the membrane potential dependence of current as a nonideal characteristic in this study. 
In actual circuits, various nonidealities arise depending on the memory type and architecture \cite{Roy2020inmemory}. 
For instance, in ReRAM, issues such as IR drop and sneak currents pose significant challenges \cite{Chen2024scaling}. 
Beyond deterministic characteristics, real circuits are also affected by stochastic phenomena such as process variation and noise, making reliable large-scale analog circuit design difficult \cite{Roy2020inmemory}. 
Efforts to mitigate these nonidealities have been made by many research groups \cite{Krestinskaya2024neural}. 
Joshi et al. \cite{Joshi2020accurate} minimized the effect of nonidealities by retraining models while incorporating Gaussian noise. 
Cao et al. \cite{Cao2022nonidealities} and Rasch et al. \cite{Rasch2023hardware} optimized parameters such as mapping scale and quantization to minimize inference loss in hardware. 
Cramer et al. \cite{Cramer2022surrogate} used a ``chip-in-the-loop'' approach, incorporating a portion of circuit dynamics into the learning process to better align the model with hardware operation. 

While these methods integrate hardware characteristics, they employ a top-down approach, where the learned model is an ANN or SNN that is mapped onto hardware. By contrast, the present study directly incorporated nonidealities into the learning process by training on the hardware's intrinsic dynamics. 
Despite these advancements, our method only incorporates a subset of nonideal characteristics. To address a broader range of nonidealities, more complex models need to be trained.
One possible approach is to incorporate nonidealities that do not harm learning efficiency into the model while addressing other nonidealities with conventional methods \cite{Krestinskaya2024neural}. 
Although this study adopted a model-based approach by representing analog dynamics with ordinary differential equations, model-free approaches offer an alternative \cite{Wright2022deep}. 
While model-free methods are versatile, they require large datasets and are challenging to apply to many non-volatile memory types. 
A promising direction for practical applications could involve combining the versatility of model-free approaches with the advantages of model-based methods \cite{Raissi2019physics, Okamoto2024learning}.

\section{Methods} \label{ss:methods}

\subsection{Neuron model incorporating IMC nonidealities} 
We modeled the analog dynamics in the IMC circuit as a non-leaky integrate-and-fire model with reversal potentials, refered to as the IMC-aware neuron model. 
See SI.\HardwareModel~to confirm the equivalency between them. 
We considered a feedforward neural network. 
The time evolution of the membrane potential of the $i$th neuron in the $l$th layer is given by 
\begin{align}
\frac{d}{dt}v_i^{(l)}(t) &= - \alpha v_i^{(l)}(t) + \sum _{q\in \{+,-\}}  p^{q(l)}_i (t) \left( E_\text{rev}^q - v_i^{(l)}(t) \right)  \label{eq:neuron_first}
\end{align}
where $\alpha (\ge 0)$ is the leak constant. 
This model possesses two reversal potentials $E_\text{rev}^+$ and $E_\text{rev}^-$ $\left( E_\text{rev}^+ > E_\text{rev}^-\right)$ corresponding to the positive and negative weights, respectively.  
The synaptic currents evoke depending on the activity of the ion channels $p_i^{q(l)}(t)$. 
This activity is given by 
\begin{align}
    &p_i^{+(l)}(t) = \frac{1}{E_\text{rev}^+} \sum_{j\in{\{w_{ij}\ge0\}}}^{N^{(l-1)}} w_{ij} \theta\left(t-t_j^{(l-1)}\right),
    ~  p_i^{-(l)}(t) = \frac{1}{E_\text{rev}^-} \sum_{j\in{\{w_{ij}<0\}}}^{N^{(l-1)}} w_{ij} \theta\left(t-t_j^{(l-1)}\right),
\end{align}
where $\theta\left(\cdot\right)$ is the Heaviside function. 
This synaptic activity corresponds to the assumption that there is no leak in the synaptic current \cite{Mostafa2018supervised, Sakemi2023supervised}. 
Moreover, for simplicity, we adopted the non-leaky neuron model $\alpha=0$ \cite{Sakemi2023supervised,  Zhang2021rectified,Sakemi2023sparse, Maass1998computing}.  
By rearranging variables in Eq. (\ref{eq:neuron_first}), we obtain the following equation
\begin{align}
\frac{d}{dt}v_i^{(l)}(t) = - f_i^{(l)}(t)v_i^{(l)}(t) + g_i^{(l)}(t),  \label{eq:my_neuron}
\end{align}
where we define the following variables 
\begin{align}
    f_i^{(l)}(t)&:=\sum_{q\in \{+,-\}}p_i^{q{(l)}}(t) = \sum _{j=0}^{N^{(l-1)}-1} \beta_{ij}^{(l)} w_{ij}^{(l)}\theta(t-t_j^{(l-1)}), \label{eq:f_t} \\
    g_i^{(l)}(t)&:=\sum_{q\in \{+,-\}}p_i^{q{(l)}}E_\text{rev}^q = \sum _{j=0}^{N^{(l-1)}-1} w_{ij}^{(l)}\theta(t-t_j^{(l-1)}), \label{eq:g_t} \\
    \beta_{ij}^{(l)} &:= \frac{\delta _{w_{ij}^{(l)}\ge 0}}{E_\text{rev}^+} + \frac{\delta _{w_{ij}^{(l)} < 0}}{E_\text{rev}^-}, \label{eq:beta}
\end{align}
where $\delta _x$ is 1 only when $x$ is true, otherwise 0. 
The analytical solution can be obtained with the method of variation of parameters: 
\begin{align}
v_i^{(l)}(t) = e^{-\int _0^tf_i^{(l)}(s) ds} \cdot \left( \int _0 ^t e^{\int _0 ^{s}f_i^{(l)}(s') ds'} g_i^{(l)}(s) ds \right),
\end{align}
where we set $v_i^{(l)}(0)=0$.  
By arranging the spike in the time order  $\{\hat{t}_0^{(l-1)} \leq \hat{t}_1^{(l-1)} \leq \dots \leq \hat{t}_{N^{(l-1)}-1}^{(l-1)}\}$, 
and representing the corresponding weights as $\{ \hat{w}_{ij}^{(l)} \}$,  
we define
\begin{align}
    \hat{t}_j^{(l)} &= t_{\hat{j}}^{(l)}, ~ \hat{w}_{ij}^{(l)} = w_{i\hat{j}}, \\
    \hat{j} &= \arg \min_k \{ t_k^{(l)}| \hat{t}_{i-1}^{(l)} \le t_k^{(l)},~ t_{-1}^{(l)}=0 \}.
\end{align}
With this time-ordered arrangement, the integral becomes a summation, then we obtain the following representation of the membrane potential
\begin{align}
v_{ij}^{(l)}&:=v_i^{(l)}(\hat{t}_j^{(l-1)}) \\
&= \begin{cases}
0, ~\text{for } j=0, \\
\frac{ \sum _{k=0}^{j-1}\frac{g_{ik}^{(l)}}{f_{ik}^{(l)}}\left( F_{ik}^{(l)} -  F_{i~k-1}^{(l)} \right) }{F_{i~j-1}^{(l)}},~\text{for } j>0,
\end{cases} \label{eq:Vmem_rigorous}
\end{align}
where we define
\begin{align}
F_{ij}^{(l)} &:= \begin{cases}
e^{\sum _{q=0}^{j}f_{iq}^{(l)} \left(\hat{t}_{q+1}^{(l-1)} - \hat{t}_q^{(l-1)} \right)~ } \text{ for } j\ge0,\\
1 ~ \text{ for } j=-1,
\end{cases} \label{eq:F_original} \\
f_{ij}^{(l)} &:= f_i^{(l)}(\hat{t}_j^{(l-1)}) = \sum_{k=0}^{j} \hat{w}_{ik}^{(l)} \beta_{ij}^{(l)}, \\
g_{ij}^{(l)} &:= g_i^{(l)}(\hat{t}_j^{(l-1)}) = \sum_{k=0}^{j} \hat{w}_{ik}^{(l)}. 
\end{align}
Detailed derivation can be found in SI.\Proof.

\subsection{RC-Spike model}

The RC-Spike model has two time phases: the accumulation phase and firing phase. 
The time evolution is described by the following differential equations \cite{Sakemi2022spiking} :
\begin{flalign}
\frac{d}{dt} v_i ^{(l)} (t) = \begin{cases}
- f_i^{(l)}(t)v_i^{(l)}(t) + g_i^{(l)}(t) , ~\text{ for } l-1 \le t < l \text{ (accumulation phase)}  \\
1, \text{ for } l \le t < l+1 \text{ (firing phase)}.
\end{cases}  \label{eq:neuron} 
\end{flalign}
The accumulation phase is the same as the neuron model described in (Eq. (\ref{eq:my_neuron})). 
Notably, when $|E_\text{rev}^\pm|\rightarrow \infty$, the final membrane potential $v_i^{(l)}(l)$ at the end of the accumulation phase matches the sum-of-products operation in an artificial neuron model \cite{Cai2019fully, Bavandpour2019energy, Yamaguchi2021energy}. 
Specifically, it corresponds to $\sum_{j=0}^{N^{(l-1)-1}} w_{ij}^{(l)} (l- t_j^{(l-1)})$. 
When $|E_\text{rev}^\pm| < \infty$, the final membrane potential does not match the sum-of-products owing to the nonlinearity \cite{Sakemi2022spiking}. 
The RC-Spike neuron fires when the membrane potential exceeds the firing threshold $V_\text{th}=1$ during the firing phase. 
The firing time is clipped within the interval of the firing phase. 
In the following formulation, we used a shifted version of the accumulation phase and firing phase so that they fall within the interval [0,1] for computational brevity.

With Eq. (\ref{eq:Vmem_rigorous}) and letting $\hat{t}_{N^{(l-1)}}^{(l-1)}=1$, the membrane potential at the end of the accumulation phase is given by 
\begin{align}
v_{i}^{(l)}(1)&:=v_i^{(l)}(\hat{t}_{N^{(l-1)}}^{(l-1)}) \\
&= \frac{ \sum _{k=0}^{N^{(l-1)}-1}\frac{g_{ik}^{(l)}}{f_{ik}^{(l)}}\left( F_{ik}^{(l)} -  F_{i~k-1}^{(l)} \right) }{F_{i~N^{(l-1)}-1}^{(l)}}, \label{eq:Vmem_1} \\
&= \sum _{k=0}^{N^{(l-1)}-1}\frac{g_{ik}^{(l)}}{f_{ik}^{(l)}}\left( Z_{ik}^{(l)} -  Z_{i~k-1}^{(l)} \right), \label{eq:Vmem_1_rigorous}
\end{align}
where we defined
\begin{align}
Z_{ij}^{(l)} &:= \frac{F_{ij}^{(l)}}{F_{i~N^{(l-1)}-1}} \\
&=\begin{cases}
\prod _{q=j+1}^{N^{(l-1)}-1}  e^{-f_{iq}^{(l)} \left(\hat{t}_{q+1}^{(l-1)} - \hat{t}_q^{(l-1)} \right) } ~ \text{ for } j < N^{(l-1)}-1.\\
1 ~ \text{ for } j= N^{(l-1)}-1.
\end{cases} \label{eq:R_original}
\end{align}

The firing time is readily obtained using Eq. (\ref{eq:neuron}) with the membrane potential at the end of the accumulation phase: 
\begin{align}
t_i^{(l)} &=  \text{clip}\left(1- v_i^{(l)}(1)\right),  \label{eq:spike_timing_rc_spike} \\
\text{clip}(x)&= \begin{cases}
0, \text{ if } x <0, \\
x, \text{ if } 0 \le x \le 1, \\
1, \text{ if } 1 < x,
\end{cases}
\end{align}
where we shifted the firing time so that the firing time falls within the interval $[0,1]$. 
Algorithm \ref{alg:RC_Spike} summarizes the calculation method of the firing time at each layer of the RC-Spike model when  input spikes $t_i^{(0)}$ are given. 

\subsection{TTFS-SNN model} 
A neuron in the TTFS-SNN model fires when the membrane potential reaches the firing threshold. 
Unlike the RC-Spike model, it is an event-driven system without phase switching.
This model enables information processing with an extremely small number of spikes by imposing the constraint that each neuron fires at most once \cite{Bohte2002error, Mostafa2018supervised, Kheradpisheh2019s4nn, Comsa2020temporal, Sakemi2023supervised, Goltz2021fast}. 
The firing time can be calculated as follows. 
Let the firing time when it is assumed as $\hat{t}_{j-1}^{(l-1)} < t_i^{(l)} \le \hat{t}_j^{(l-1)}$ be $t_{i~j-1}^{(l)}$. 
In Eq. (\ref{eq:Vmem_rigorous}), by substituting $\hat{t}_j^{(l-1)}$ with $t_{i~j-1}^{(l)}$, 
the firing time can be obtained from the firing condition $v_i^{(l)}(t_{i~j-1}^{(l)})=V_\text{th}$ as follows: 
\begin{align}
 t_{i~j-1}^{(l)} = t_{j-1}^{(l-1)} + \frac{1}{f_{i~j-1}^{(l)}} \ln \left( \frac{\frac{g_{i~j-1}^{(l)}}{f_{i~j-1}^{(l)}}-v_{i~j-1}^{(l)} }{\frac{g_{i~j-1}^{(l)}}{f_{i~j-1}^{(l)}} - V_\text{th}} \right). \label{eq:spike_timing_rigorous}
\end{align}
Then, the actual firing timing is obtained by 
\begin{align}
t_i^{(l)} = \min_j \{t_{i~j-1}^{(l)}| ~\hat{t}_{j-1}^{(l-1)} < t_i^{(l)} \le \hat{t}_j^{(l-1)}, 1\le j \le N^{(l-1)}, \hat{t}_{N^{(l-1)}}^{(l-1)}=\infty \}. \label{eq:spike_timing_min}
\end{align}
We note that when no leak neuron ($\alpha=0$) is used, 
another condition such $v_{ij}^{(l)} \ge V_\text{th} $ can be used instead of $\hat{t}_{j-1}^{(l-1)} < t_i^{(l)} \le \hat{t}_j^{(l-1)}$. 
If the condition of Eq. (\ref{eq:spike_timing_rigorous}) is not satisfied, 
the neuron does not fire, and the firing timing $t_i^{(l)}$ is assigned a value of $\infty$. 
To calculate Eq. (\ref{eq:spike_timing_rigorous}), 
it is necessary to compute the membrane potential (Eq. (\ref{eq:Vmem_rigorous})) at all times when input spikes are received.  
However, $F_{ij}^{(l)}$ of Eq. (\ref{eq:F_original}) is an exponential function, which can easily lead to overflow. 
To prevent this, we can instead compute $\left( F_{ik}^{(l)} -  F_{i~k-1}^{(l)} \right)/F_{i~j-1}^{(l)}$. 
However, because this is a 3-dimensional tensor, it is extremely memory inefficient. 
By using the following recursion
\begin{align}
F_{ij}^{(l)} = e^{f_{ij}^{(l)}\left(\hat{t}_{j+1}^{(l-1)} - \hat{t}_j^{(l-1)} \right)} F_{i~j-1}^{(l)}, 
\end{align}
the membrane potential update equation can be recursively written as follows: 
\begin{align}
v_{ij}^{(l)} &= \frac{\frac{g_{i~j-1}^{(l)}}{f_{i~j-1}^{(l)}}\left(F_{i~j-1}^{(l)} - F_{i~j-2}^{(l)} \right)}{ F_{i~j-1}^{(l)}} 
+ \frac{\sum_{q=0}^{j-2} \frac{g_{iq}^{(l)}}{f_{iq}^{(l)}}\left(F_{iq}^{(l)} - F_{i~q-1}^{(l)} \right) }{ e^{f_{i~j-1}^{(l)}\left(t_{j}^{(l)} - t_{j-1}^{(l)} \right)} F_{i~j-2}^{(l)}     }\\
&=\frac{g_{i~j-1}^{(l)}}{f_{i~j-1}^{(l)}} \left(1 - e^{-f_{i~j-1}^{(l)}\left(t_{j}^{(l)} - t_{j-1}^{(l)} \right)} \right) + e^{-f_{i~j-1}^{(l)}\left(t_{j}^{(l)} - t_{j-1}^{(l)} \right)} v_{i~j-1}^{(l)} \\
&= \frac{g_{i~j-1}^{(l)}}{f_{i~j-1}^{(l)}}  + \left( v_{i~j-1}^{(l)} - \frac{g_{i~j-1}^{(l)}}{f_{i~j-1}^{(l)}} \right) e^{-f_{i~j-1}^{(l)}\left(t_{j}^{(l)} - t_{j-1}^{(l)} \right)}. \label{eq:vmem_recursive}
\end{align}

Algorithm \ref{alg:TTFS} summarizes the computing method of the firing time for each layer when input spikes $t_i^{(0)}$ are given. 

\subsection{Differentiable spike-time discretization}

As explained in previous sections, 
when calculating the firing time, in the case of RC-Spike, 
the element of matrix ($\in \mathbb{R}^{N_\text{out} \times N_\text{in}}$) must be computed (see Eq. (\ref{eq:Vmem_1_rigorous}).) . 
Here, $N_\text{out}$ is the output dimension and $N_\text{in}$ is the input dimension. 
Because each element requires $\mathcal{O}(N_\text{in})$ computations, the total computational cost is $\mathcal{O}(N_\text{in}^2 N_\text{out})$. 
For TTFS-SNN models, the element of matrix ($\in \mathbb{R}^{N_\text{out} \times N_\text{in}}$) must be computed repeatedly $N_\text{in}$ times (see Eq. (\ref{eq:spike_timing_rigorous})). 
Because each element requires $\mathcal{O}(N_\text{in})$ computations, the total computational cost is $\mathcal{O}(N_\text{in}^3 N_\text{out})$. 
As shown in table \ref{tab:computational_cost}, 
the neuron model incorporating the reversal potentials require much more computations compared with ANN models, which prevents the training of large-scale models. 
To mitigate this problem, we introduce differentiable spike-time discretization (DSTD) technique. 

First, consider a situation where the arrival times of the spikes received by the neuron are limited to the following discrete time points \cite{Rotter1999exact}: 
\begin{flalign}
&T^{(l)}_m =  m \Delta _\tau - t_\text{offset}^{(l)},~(m=0,1,\dots, M),  \\
&0 < t_\text{offset}^{(l)} < \Delta _\tau.
\end{flalign}
Here, $\Delta _\tau$ is the discrete-time width, 
and $t_\text{offset}^{(l)}$ represents the offset time. 
In this case, Eq. (\ref{eq:f_t}) and Eq. (\ref{eq:g_t}) can be rewritten as follows:
\begin{align}
    \tilde{f}_{im}^{(l)}:=f_i^{(l)}(T_m^{(l-1)})&= \sum _{j=0}^{N^{(l)}-1} \beta_{ij}^{(l)} w_{ij}^{(l)} S_{jm}^{(l-1)}, \label{eq:f_discrete} \\
    \tilde{g}_{im}^{(l)}:=g_i^{(l)}(T_m^{(l-1)})&= \sum _{j=0}^{N^{(l)}-1} w_{ij}^{(l)}  S_{jm}^{(l-1)}, \label{eq:g_discrete}
\end{align}
where we define
\begin{align}
S_{jm}^{(l-1)}&:=\sum_{q=0}^m s_{jq}^{(l-1)}.
\end{align}
$s_{jq}^{(l)}$ is the $j$th spike variable at the layer $l-1$ at time $T_q^{(l)}$. 
The spike variable is a binary variable, where it takes the value of $1$ when a spike exits at a given time, and $0$ otherwise. 
In this condition, the membrane potential at the end of the accumulation phase can be calculated as
\begin{align}
v_i^{(l)}(1) &= \sum _{m=0}^{M-1}\frac{\tilde{g}_{im}^{(l)}}{\tilde{f}_{im}^{(l)}}\left( \tilde{Z}_{im}^{(l)} -  \tilde{Z}_{i~m-1}^{(l)} \right),   \label{eq:spike_timing_fast}
\end{align}
where we defined
\begin{align}
\tilde{Z}_{im}^{(l)}:=\begin{cases}
\prod _{q=m+1}^{M-1}  e^{\tilde{f}_{iq}^{(l)} \left(\hat{t}_{q+1}^{(l-1)} - \hat{t}_q^{(l-1)} \right)}~ \text{ for } m<M-1. \\
1 ~ \text{ for } m=M-1. 
\end{cases}
\end{align}
Similarly, the firing time for the case of TTFS-SNN models is given by
\begin{align}
t_i^{(l)} = \min_m \{ \tilde{t}_{im}^{(l)}|  \tilde{v}_{im}^{(l)}\ge V_\text{th}, 0 \le 
 m \le M \},  \label{eq:spike_timing_DSTD_TTFS}
\end{align}
where we defined
\begin{align}
\tilde{v}^{(l)}_{im}&:=   v_i^{(l)}\left(T_m^{(l)} \right) \\
&= \frac{\tilde{g}_{i~m-1}^{(l)}}{\hat{f}_{i~m-1}^{(l)}}  + \left( \tilde{v}_{i~m-1}^{(l)} - \frac{\tilde{g}_{i~m-1}^{(l)}}{\tilde{f}_{i~m-1}^{(l)}} \right) e^{-\tilde{f}_{i~m-1}^{(l)}\left(T_{i~m}^{(l)} - T_{i~m-1}^{(l)} \right)}, \label{eq:Vmem_fast} \\
\tilde{t}_{i~m-1}^{(l)} &:= T_{m-1}^{(l)} - \frac{1}{\tilde{f}_{i~m-1}^{(l)}} \ln \left( \frac{\frac{\tilde{g}_{i~m-1}^{(l)}}{\tilde{f}_{i~m-1}^{(l)}}-\tilde{v}_{im}^{(l)} }{\frac{\tilde{g}_{i~j-1}^{(l)}}{\tilde{
f}_{i~j-1}^{(l)}} - V_\text{th}} \right). \label{eq:timing_TTFS_fast} 
\end{align}
In the above computation, we compute the elements of the matrices 
$\tilde{v}_{im}^{(l)},\tilde{f}_{im}^{(l)}, \tilde{g}_{im}^{(l)}, \tilde{Z}_{im}^{(l)} \in\mathbb{R}^{N^{(l)}\times M} $ instead of those of the matrices $v_{ij}^{(l)}, f_{ij}^{(l)}, g_{ij}^{(l)}, Z_{ij}^{(l)} \in \mathbb{R}^{N^{(l)} \times N^{(l-1)}}$, which implies that computational cost is drastically reduced if $M \ll N^{(l-1)}$ (table \ref{tab:computational_cost}). 
However, there are two problems with this approach. 
The first issue is that the spike timing is a continuous value ($t_i^{(l)}\in \mathbb{R}$); therefore, the spike arrival time fundamentally does not inherently exist as a discrete value.
The second problem is that, because the spike timing is discretized, it is not possible to calculate gradients with respect to time. 
DSTD solves these problems by approximating the spike variable $s_{jq}^{(l)}$ as a real number in the following way:
\begin{flalign}
\{s_{j0}^{(l)}, s_{j1}^{(l)}, \dots, s_{j~M-1}^{(l)} \} = \text{DSTD} \left( t_j^{(l)} \middle| \left\{T^{(l)}_m\right\}_{m=0,1,\dots,M} \right).  \label{eq:spike_variables}
\end{flalign}
The DSTD function is differentiable  with respect to time, enabling learning through gradient descent methods. 
In this study, we utilized the DSTD function as follows: 
\begin{flalign} 
s_{jm}^{(l-1)} = \text{max} \left( 0 ,\frac{\Delta _\tau-|T_m^{(l)}-t_j^{(l-1)}|}{\Delta _\tau} \right).  \label{eq:discrete_representation}
\end{flalign}
In the case of RC-Spike, because $T_{M}^{(l)}=1$, the time width at the final time changes depending on the offset time $t_\text{offset}$. 
Therefore, it is necessary to compute it as follows:
\begin{flalign}
s_{jm}^{(l-1)} = \begin{cases}
\text{max} \left( 0 ,\frac{\Delta _\tau-|T_m^{(l)}-t_j^{(l-1)}|}{\Delta _\tau} \right), \text{ for } t_i^{(l)} \le T_{M-1}^{(l)}, \\
\text{max} \left( 0 ,\frac{t_\text{offset}^{(l)}-|T_m^{(l)}-t_j^{(l-1)}|}{t_\text{offset}^{(l)}} \right), \text{ for } t_i^{(l)} > T_{M-1}^{(l)}.
\end{cases}  \label{eq:discrete_representation_RC_Spike}
\end{flalign}

The calculation method for the RC-Spike model using DSTD is shown in Algorithm \ref{alg:Fast_RC_Spike}, and the calculation method for the TTFS-SNN model using DSTD is shown in Algorithm \ref{alg:TTFS_DSR}. 
It should be noted that when using DSTD, sorting the spikes in time order is not required. 
Moreover, the matrix size of $\beta_{ij}^{(l)}$ (Eq. (\ref{eq:beta})) is $N_\text{in}\times N_\text{out}$. However, as this computational cost remains independent of the batch size, it becomes negligible when processing large batch sizes. 
DSTD theoretically guarantees that when the discrete time width $\Delta_\tau$ is sufficiently small, the resulting membrane potential accurately approximates the exact ODEs. 
We proved the following theorem. 

{\bf Theorem 1.}
{\it When a nonleaky neuron receives a number of spikes in the interval of $[0,1]$, 
the error between the exact membrane potential and the approximated membrane potential satisfies}
\begin{align}
    |v(1) - \hat{v}(1)| = \mathcal{O}(\Delta_\tau^2 |E_\text{rev}^\pm|^{-1}),
\end{align}
{\it where $\mathcal{O}(\cdot)$ is the big $\mathcal{O}$ notion.} The more rigorous form of this theorem and its proof are given in SI.\Proof. 

\subsection{Learning algorithms} 
The supervised learning of the RC-Spike model and the TTFS-SNN model was performed using the following cost function: 
\begin{flalign}
C\left(t^{(L)} \middle| ~\kappa \right)  &= \text{Loss} \left(t^{(L)} \middle|~\kappa \right) + \gamma_1 \text{Temp}\left(t^{(L)}\right) \label{eq:cost_function} + \gamma_2 Q, \\
\text{Loss}\left(t^{(L)}\middle|~\kappa\right)  &= \sum _{i=0}^{N^{(L)}-1} \kappa _i \ln S_i \left(t^{(L)}\right),  \\
S_i\left(t^{(L)}\right)&= \frac{\exp\left(\frac{t_i^{(L)}}{\tau _\text{soft}}\right)}{\sum_{j=0}^{N^{(L)}-1} \exp \left(\frac{t_j^{(L)}}{\tau _\text{soft}}\right)},  \\
\text{Temp}\left(t^{(L)}\right) &= \sum _{i=0}^{N^{(L)}-1}\left(t_i^{(L)} - t^\text{ref}\right) ^2,  \\
Q &= \sum_{l=1}^{l=L}\sum _{j\in \Gamma_i^{(l)}}w_{ij}^{(l)}.
\end{flalign}
Here, $C(\cdot)$ is the cost function, $\text{Loss}(\cdot)$ is the classification loss function, 
and $\text{Temp}(\cdot)$ is the temporal penalty term. 
$N^{(L)}$ represents the number of neurons in the output layer, 
and $t^{(L)}=\left(t_0^{(L)}, t_1^{(L)}, \dots, t_{N^{(L)}-1}^{(L)}\right)$ denotes the firing times of the neurons in the output layer. 
$\kappa=\left(\kappa_0, \kappa_1, \dots, \kappa_{N^{(L)}-1} \right)$ is the teacher label, where  the element corresponding to the correct label is 1, and all other elements take a value of 0.
$S_i(\cdot)$ represents the softmax function, and $\tau _\text{soft}$ is a positive real number that performs scaling of the softmax. 
$\gamma_1$ is a positive constant that controls the significance of the temporal penalty term. The temporal penalty term can stabilize the learning process \cite{Sakemi2023supervised}. 
$Q(\cdot)$ is used when training the TTFS-SNN model and stabilizes learning by promoting firing \cite{Sakemi2023sparse}. 
$\Gamma_i^{(l)}$ is a set of indices of spikes that arrive before the $i$th neuron in the $l$th layer fires. 
The strength of this regularization is adjusted by $\gamma_2$. 
Learning was performed by minimizing this cost function using gradient descent with the Adam optimizer \cite{Kingma2014adam} at a learning rate of $\eta$.  
In the numerical simulations of the RC-Spike models, the following hyperparameters were used: $\eta=10^{-4}$,  $\gamma_1=2.6$, $t^\text{ref}=0.9$, $\tau_\text{soft}=0.07$, $\gamma_2=0$, and minibatch size of $32$. 
For the case of TTFS models, we used the following hyperparameters: 
$\eta=2\cdot10^{-4}$,  $\gamma_1=0.02$, $t^\text{ref}=3.2$, $\tau_\text{soft}=0.07$, $\gamma_2=8\cdot 10^{-6}$, and minibatch size of $32$. 

\subsection{Dataset} 
The Fashion-MNIST data consist of 60,000 training images and 10,000 test images.
Each data point is a $28 \times 28$ image with a single channel, and the task was to predict the correct label from 10 categories, such as T-shirts and trousers.
The brightness of each pixel, $x_{ijk}~(i=0,1,...,27,~j=0,1,...,27,~k=0)$, normalized to the range $[0,1]$, was converted into an input spike as follows, with the spike timing defined as 
\begin{flalign}
t_{ijk}^{(0)} = \tau _\text{in} (1-x_{ijk}), \label{eq:input_spike_cnn}
\end{flalign}
where  $\tau_\text{in}$ is the input time scale, which is set to $1$ for all cases. 
For the fully connected neural network cases, 
the input data were flattened as $\hat{x}_i(i=0,1,..., 783)$ and converted into input spikes
\begin{flalign}
t_i^{(0)} = \tau _\text{in} (1-x_i). 
\end{flalign}

The CIFAR-10 dataset consists of 50,000 training data points and 10,000 test data points. 
Each data point is an image with dimensions of $32\times32$ and three channels. 
Each data point is labeled with one of 10 categories such as airplane and automobile, and the task was to predict the correct label. 
The normalized data $x_{ijk}~(i=0,1,...,31,~j=0,1,...,31,~k=0,1,2)$ were converted into input spikes using Eq. (\ref{eq:input_spike_cnn}). 
Additionally, similar to previous research \cite{Zhou2021temporal}, data augmentation techniques such as horizontal flip, rotation, and crop were applied.

\subsection{Hardware simulation} 
When computing the ANN model with the IMC circuit, the ANN model was first trained, and the learned weights were then mapped to the IMC circuit (the ANN-to-IMC mapping). 
We used the RC-Spike model  with $|E_\text{rev}^\pm|=100$ as the ANN model. 
When computing the RC-Spike model with the IMC circuit, 
the reversal potentials $E_\text{rev}^{\pm}$ was first calculated through circuit simulation, 
and based on that, the RC-Spike model was trained. 
Then, the trained weights were mapped onto the IMC circuit (the PNN-to-IMC mapping). 
The relationship between the parameters of the IMC circuit and the model parameters is shown in Table \ref{tab:params_HW}. 
Algorithm \ref{alg:HW_simulation_ANN} and Algorithm \ref{alg:HW_simulation_PNN} summarize the procedures for the cases of the ANN-to-IMC mapping and the PNN-to-IMC mapping, respectively. 

The RC-Spike model and designed IMC circuit comprise a fully connected network with single hidden layer (5-5-5 network).  
We adopted the Iris dataset \cite{Fisher1936, scikit-learn} to train the RC-Spike model. 
The Iris dataset consists of 150 data points, with each data point represented as a 4-dimensional vector. 
The task was to classify each data point into one of three classes. 
To process these data using the RC-Spike model, the value $x_i$ of each dimension of each data point was converted into input spikes as follows:
\begin{align}
t_i^{(0)} = \tau _\text{in} x_i (i = 0, 1, 2, 3).
\end{align}
Additionally, a bias spike was introduced as $t_4^{(0)}=0$. 
To make the mapping procedure efficient, significantly large currents must be avoided. 
In addition, early firing spikes should be prevented, which make circuit design difficult. 
We therefore added the following regularization terms to the cost function: 
\begin{align}
P &= \sum_{l=1}^2 \|w^{(l)}\|_2^2,\\
V &= \sum_{i=0}^4 \left(t_i^{(1)} - 1 \right)^2 + \left( t_i^{(2)} - 1 \right)^2,
\end{align}
with the coefficients $\gamma_P$ and $\gamma_V$ that control the effects of $P$ and $V$, respectively. 
The parameters used in the learning process are shown in table \ref{tab:params_HW}.

\section{Acknowledgements}
This work was partially supported by 
JST PRESTO Grant Number JPMJPR22C5, 
NEC Corporation, 
SECOM Science and Technology Foundation, 
JST Moonshot R\&D Grant Number JPMJMS2021, 
Institute of AI and Beyond of UTokyo, 
the International Research Center for Neurointelligence (WPI-IRCN) at The University of Tokyo Institutes for Advanced Study (UTIAS), 
JSPS KAKENHI Grant Number JP20H05921, 
Cross-ministerial Strategic Innovation Promotion Program (SIP), the 3rd period of SIP , Grant Numbers JPJ012207 and JPJ012425. 
Computational resource of AI Bridging Cloud Infrastructure (ABCI) provided by National Institute of Advanced Industrial Science and Technology (AIST) was used.
Yusuke Sakemi extends gratitude to Prof. Osamu Nomura and the Open Source Silicon Design Community for their invaluable support in circuit design and simulation.

\newpage
\clearpage

\clearpage

\clearpage

\clearpage

\begin{figure*}
\centering
\includegraphics[clip, width=\textwidth]{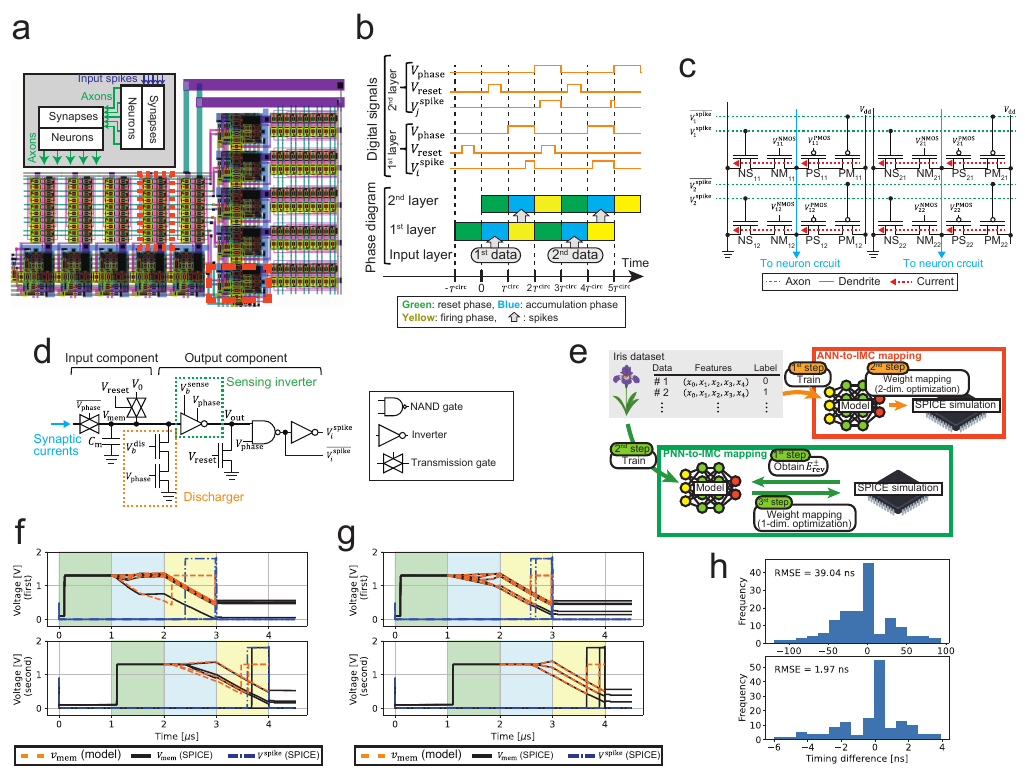}
\caption{{\bf Circuit overview.} 
{\bf a.} Circuit layout of the RC-Spike circuit designed using the sky130 PDK. 
This circuit forms a two-layer network, with each layer composed of five neurons.
The inset shows a schematic overview of the layout. 
One synapse circuit in the second layer and one neuron circuit in the first layer are enclosed by orange dotted and dashed lines, respectively.
{\bf b.} Timing diagram. Each layer of the RC-Spike circuit operates through three phases: the reset phase, accumulation phase, and firing phase.
These phases are controlled by two digital signals: the phase signal and  reset signal.
During the accumulation phase, spike signals are received by each layer, while in the firing phase, the spike signals are transmitted.
{\bf c.} Circuit diagram of the synapse circuit.
The diagram illustrates the case of two inputs and two outputs for simplicity. 
The circuit employs a paired configuration of MOSFETs for weights and selectors, commonly referred to as a 1T1R topology. 
Two types of pairs exist: one composed of N-type MOSFETs (NMOS) for positive weights (denoted as NM$_{ij}$ and NS$_{ij}$) and the other composed of P-type MOSFETs (PMOS) for negative weights (denoted as PM$_{ij}$ and PS$_{ij}$). 
Note that the membrane potential in the circuit is inverted in polarity. 
The MOSFETs responsible for the weights (NM$_{ij}$ and PM$_{ij}$) regulate the current, with the amount controlled by the bias voltage $V_{ij}^{N(P)}$.
In the simulation, the bias voltage is externally supplied.
The selector MOSFETs turn ON when a spike signal $V_i^\text{spike}$ arrives, allowing current to flow into the dendrites.
All MOSFETs used in the synapse circuit have a gate length of 250 nm and a gate width of 1 $\mu\text{m}$. 
{\bf d.} Circuit diagram of the neuron circuit.
The neuron circuit consists of two components: the input component and  output component.
The input component accumulates current from the synapse circuit during the accumulation phase. 
The output component generates a spike during the firing phase.
During the firing phase, the discharger circuit decreases the membrane potential at a constant rate, and the sensing inverter triggers a spike when the membrane potential falls below a threshold value, by inverting its output. 
(continues to the next page)
}
\label{fig:CircuitOverview}
\end{figure*}
\clearpage

\addtocounter{figure}{-1}
\begin{figure} [t!]
  \caption{{\bf (continued)}
  {\bf e.} Outline of methods for reproducing the behavior of ANNs and PNNs, the RC-Spike model in this case, on IMC hardware, termed ANN-to-IMC and PNN-to-IMC mapping, respectively. 
In the ANN-to-IMC mapping, we first train an ANN model (approximated here using $|E_\text{rev}^\pm| = 10^2$ for the RC-Spike model) on the Iris dataset to optimize its weights. 
These weights are subsequently mapped to current values in the synaptic circuits of the RC-Spike circuit based on specific hardware parameters. 
In the PNN-to-IMC mapping, the process begins with determining $E_\text{rev}^\pm$ to construct the PNN model (the RC-Spike model in this case). 
The procedure then follows the same steps as in the ANN-to-IMC mapping. 
To account for the influence of parasitic capacitance in the circuit, current values are appropriately scaled. 
In the case of ANN-to-IMC mapping, both positive and negative currents are scaled and optimized to account for the effects of $E_\text{rev}^\pm$ present in the circuit (2-dimensional scaling) in addition to the effects of parasitic capacitance. 
For the PNN-to-IMC mapping, the absolute values of weights are scaled uniformly without distinguishing between positive and negative values (1-dimensional scaling). 
{\bf f.} Comparison between the results of the mathematical model and the SPICE simulation for the ANN-to-IMC mapping. 
The upper panel illustrates the hidden layer's results, while the lower panel shows the output layer's results. 
In each panel, the reset, accumulation, and firing phases are highlighted in green, blue, and yellow, respectively. 
The time evolution of the membrane potential as predicted by the mathematical model is depicted by the orange dashed line, where firing events are indicated by resetting the membrane potential to its resting value. 
The time evolution of the membrane potential obtained by the SPICE simulation is shown as a black solid line, and the output spike signals are represented by blue dash-dotted lines. 
{\bf g.} Comparison between the mathematical model and SPICE simulation results for the PNN-to-IMC mapping, with the same notation and line styles as described in {\bf b}.
{\bf h.} Histogram of the timing differences in output layer spikes between the mathematical model and SPICE simulation. 
The top and bottom panels present the results for the ANN-to-IMC mapping and PNN-to-IMC mapping, respectively.  
The root mean square error (RMSE) is indicated in the upper-left corner of each panel.
}
\end{figure}
\clearpage

\begin{algorithm}[H]
\caption{RC-Spike}
\label{alg:RC_Spike}
\begin{algorithmic}[1] 
\State Obtain input spikes $t_i^{(0)}$ from data  
\For{all layers ($l=1,...,L$)}
  \State Obtain $\hat{t}_j^{(l-1)}$ by sorting $t_j^{(l-1)}$.
  \State Calculate $v_i^{(l)}(1)$ with Eq. (\ref{eq:Vmem_1}), 
  \State Calculate $t_i^{(l)}$ with Eq. (\ref{eq:spike_timing_rc_spike}). 
\EndFor
\end{algorithmic}
\end{algorithm}

\begin{algorithm}[H]
\caption{TTFS-SNN}
\label{alg:TTFS}
\begin{algorithmic}[1] 
\State Obtain input spikes $t_i^{(0)}$ from data  
\For{all layers ($l=1,...,L$)}
  \State Obtain $\hat{t}_j^{(l-1)}$ sorting $t_j^{(l-1)}$.
  \State Calculate $v_{ij}^{(l)}$ with Eq. (\ref{eq:vmem_recursive})
  \State Calculate $t_{ij}^{(l)}$ with Eq. (\ref{eq:spike_timing_rigorous})
  \State Calculate $t_i^{(l)}$ solving Eq. (\ref{eq:spike_timing_min}) 
\EndFor
\end{algorithmic}
\end{algorithm}

\begin{algorithm}[H]
\caption{RC-Spike with DSTD}
\label{alg:Fast_RC_Spike}
\begin{algorithmic}[1] 
\State Obtain input spikes $t_i^{(0)}$ from data  
\For{all layers ($l=1,...,L$)}
  \State Draw $t_\text{offset}^{(l)}$ from $(0, \Delta _\tau)$
  \State Calculate $s_i^{(l-1)}$ with $t_i^{(l-1)}$ using Eq. (\ref{eq:discrete_representation_RC_Spike}).
  \State Calculate $v_i^{(l)}(1)$ using Eq. (\ref{eq:spike_timing_fast}).
  \State Calculate $t_i^{(l)}$ using Eq. (\ref{eq:spike_timing_rc_spike}).
\EndFor
\end{algorithmic}
\end{algorithm}

\begin{algorithm}[H]
\caption{TTFS-SNN with DSTD}
\label{alg:TTFS_DSR}
\begin{algorithmic}[1] 
\State Obtain input spikes $t_i^{(0)}$ from data  
\For{all layers ($l=1,...,L$)}
  \State Draw $t_\text{offset}^{(l)}$ from $(0, \Delta _\tau)$
  \State Calculate $s_{jm}^{(l-1)}$ with $t_j^{(l-1)}$ using Eq. (\ref{eq:discrete_representation}).
  \State Calculate $\tilde{v}_{ij}^{(l)}$ using Eq. (\ref{eq:Vmem_fast})
  \State Calculate $\tilde{t}_{im}^{(l)}$ using Eq. (\ref{eq:timing_TTFS_fast}).
  \State Calculate $t_i^{(l)}$ solving Eq. (\ref{eq:spike_timing_DSTD_TTFS}). 
\EndFor
\end{algorithmic}
\end{algorithm}

\begin{algorithm}[H]
\caption{Hardware simulation processure with ANN-to-IMC mapping}
\label{alg:HW_simulation_ANN}
\begin{algorithmic}[1] 
\State Train RC-Spike model with $|E_\text{rev}^\pm|=100$.
\State Convert the trained weights into synaptic currents. 
\State Optimize the positive and negative current scales by circuit simulation.
\end{algorithmic}
\end{algorithm}

\begin{algorithm}[H]
\caption{Hardware simulation processure with PNN-to-IMC mapping}
\label{alg:HW_simulation_PNN}
\begin{algorithmic}[1] 
\State Obtain the nonideality of memories $\lambda^{N(P)}$ by circuit simulation.   
\State Train RC-Spike model with $E_\text{rev}^\pm$ calculated from $\lambda^{N(P)}$.
\State Convert the trained weights into synaptic currents. 
\State Optimize the current scale by circuit simulation. 
\end{algorithmic}
\end{algorithm}

\begin{table} 
\begin{center}
\caption{Computational cost of a single layer for various models. 
$N_\text{in}$: input dimension, $N_\text{out}$: output dimension, $M$: discrete time points}
\begin{tabular}[htbp]{llll}
 & ANN & RC-Spike & TTFS \\ \hline  
Conventional & $\mathcal{O}(N_\text{in} N_\text{out})$ & $\mathcal{O} \left( N_\text{in}^2 N_\text{out} \right)$ &  $\mathcal{O} \left( N_\text{in}^3 N_\text{out} \right)$  \\
With DSTD & N/A & $\mathcal{O} \left( M N_\text{in} N_\text{out} \right)$ &  $\mathcal{O} \left( M^2 N_\text{in} N_\text{out} \right) $ 
\end{tabular}
\label{tab:computational_cost} 
\end{center}
\end{table}

\begin{table} 
\begin{center}
\caption{Parameters used in circuit design and simulation}
\begin{tabular}{lll}
HW parameter & explanation & value \\ \hline
$V_\text{dd}$ & supply voltage & 1.8 V \\
$C_\text{m}$ & membrane capacitance & 140 fF \\
$T^\text{circ}$ & time interval & $1$ $\mu$s \\
$V_\text{switch}$ & switching threshold of sensing inverters & 0.428V \\
$V_0$ & resting membrane potential & 1.3 V \\
$V_\text{th}^\text{circ}$ & threshold voltage & $V_\text{0}$ - $V_\text{switch}$(=0.872V) \\
$V_b^\text{sense}$ & bias voltage for sensing inverter & 0.95 V\\
$V_b^\text{dis}$ & bias voltage for discharger circuit & 0.554 V \\
$\lambda^N$ & nonideality coefficient of NMOS current & 0.41 \\
$\lambda^P$ & nonideality coefficient of PMOS current & 0.75 \\
$\lambda^\text{dis}$ & nonideality coefficient of discharging current & 0.177  \\
$I_{ij}^{(l)}$ & synaptic current & $C_\text{m} V_\text{th}^\text{circ}(T^\text{circ})^{-1} w_{ij}^{(l)} $\\
\hline Model parameter & explanation & value \\ \hline  
$E_\text{rev}^+$ & positive reversal potential & $\left(V_\text{th}^\text{circ}\lambda^N\right)^{-1}~(=2.80)$ \\
$E_\text{rev}^-$ & negative reversal potential & $-\left(V_\text{th}^\text{circ} \lambda^P\right)^{-1}~(=-1.53)$ \\
$E_\text{rev}^\text{dis}$ & positive reversal potential in firing phase & $\left(V_\text{th}^\text{circ} \lambda^\text{dis}\right)^{-1}~(=6.44)$ \\\hline
Learning parameter & explanation & value \\ \hline
$\eta$    & learning rate & $10^{-3}$ \\
batch size & batch size & 50 \\
$\tau _\text{soft}$ & softmax scale & 0.07 \\
$\gamma_T$ & temporal penalty & 0.1\\
$\gamma_W$  & weight's 2-norm & $10^{-2}$ \\
$\gamma_V$  & early spike penalty & $0.2$ \\
\end{tabular}
\label{tab:params_HW} 
\end{center}
\end{table}

\newpage
\appendix

\section{RC-Spike-Equivalent Hardware models} \label{ss:hardware_models}

\begin{figure*}
\centering
\includegraphics[clip, width=16cm]{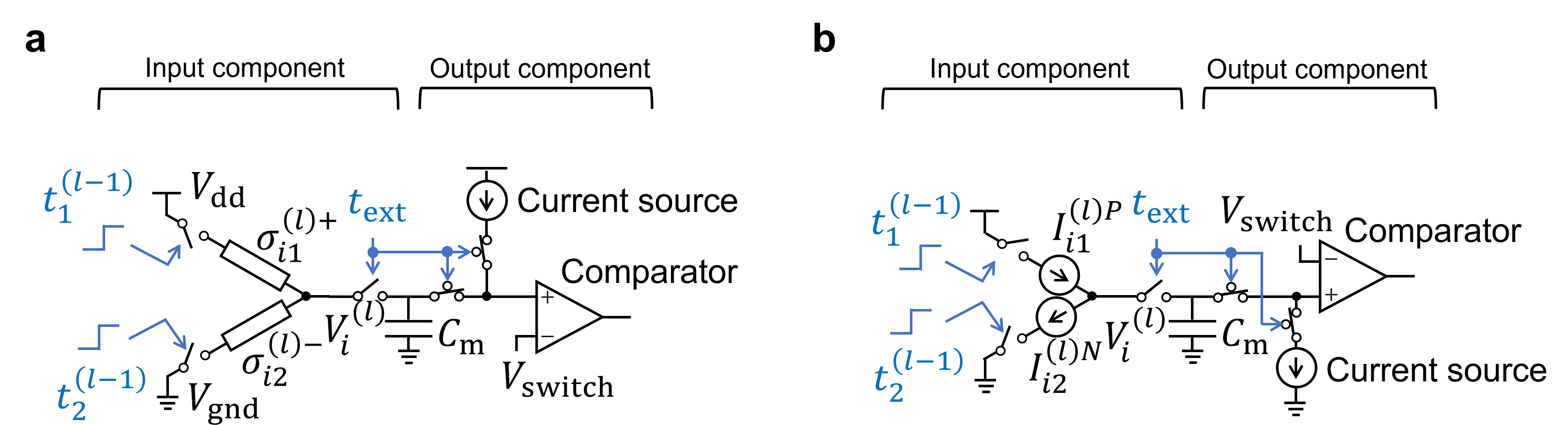}
\caption{{\bf a.} Charge-domain IMC circuits with resistors and transistor switches. 
The network weights are represented by the conductance values of resistors, denoted as $\sigma_{ij}^{(l)\pm}$.
$V_\text{dd}$ is the supply voltage and $V_\text{gnd}$ is the ground voltage. 
{\bf b.} Charge-domain IMC circuits with current sources and transistor switches. 
The network weights are represented by the current value, $I_{ij}^{(l)P}$ for positive value and $I_{ij}^{(l)N}$ for negative value.  
In both cases, two spikes from the $l-1$ layer are recieved at times $t_1^{(l-1)}$ and $t_2^{(l-1)}$. 
Due to the resulting synaptic currents, the membrene potential $V_i^{(l)}$ is altered accordingly. 
}
\label{fig:HW_models}
\end{figure*}

The RC-Spike model incorporates the dependency of synaptic current on membrane potential, which is a non-ideal characteristic inherent to charge-domain in-memory computing (IMC) systems. 
In Fig. \ref{fig:HW_models}, circuit diagrams are presented for charge-domain IMC implementations using resistors and transistor switches, as well as for those employing current sources. In the following subsections, we demonstrate that the behavior of these circuits aligns with that of the RC-Spike model.

\subsection{Resistively coupled synapse model}

According to Ohm's law and Kirchhoff's current law, the membrane potential of the circuit shown in Fig. \ref{fig:HW_models} {\bf a}  evolves over time as follows:
\begin{flalign}
C_\text{m}\frac{d}{dt}V_i^{(l)}(t) = \begin{cases}
\sum _{j=0}^{N^{(l-1)}-1} \sigma _{ij}^{(l)+}\left(V_\text{dd} - V_i^{(l)}(t) \right) \theta\left( t-t_j^{(l-1)}\right) + \sigma _{ij}^{(l)-}\left(V_\text{gnd} - V_i^{(l)}(t)\right) \theta \left( t-t_j^{(l-1)} \right) \\
~~~~~~~~~~~~~ \text{for } (l-1)T^\text{circ} \le t < lT^\text{circ} \text{ (accumulation phase)}, \\
\frac{C_\text{m}V_\text{th}^\text{circ}}{T^\text{circ}},~\text{for } lT^\text{circ} \le t < (l+1)T^\text{circ} \text{ (spike generation phase)},
\end{cases} \label{eq:resistively_coupled_model}
\end{flalign}
where $V_i^{(l)}(t)$ represents the membrane potential of neuron $i$ in layer $l$, corresponding to the voltage across capacitor $C_\text{m}$. 
The time interval $T^\text{circ}$ defines the duration of either the accumulation phase or the firing phase, while $V_\text{th}^\text{circ}$ denotes the firing threshold defined as $V_\text{switch} - V_0$. 
The terms $\sigma _{ij}^{(l)\pm}$ correspond to the electrical conductances (inverse of resistance) representing the synaptic strength from neuron $j$ in layer $(l-1)$ to neuron $i$ in layer $l$. 
The positive ($+$) and negative ($-$) symbols indicate current inflow and outflow, respectively. When $\sigma _{ij}^{(l)+} > 0$, we set $\sigma _{ij}^{(l)-} = 0$, i.e., only current inflow occurs, and vice versa when $\sigma _{ij}^{(l)-} > 0$. 
In IMC circuits, a single weight is typically represented using two memory devices, one for the positive value and the other for the negative value \cite{Sebastian2020memory}.

We show the equivalence of this differential equation to the RC-Spike model. 
First, we scale the voltage and time variables as follows: 
\begin{flalign}
v_i^{(l)}=\frac{\left(V_i^{(l)} - V_0\right)}{V_\text{th}^\text{circ}},
~E_\text{rev}^+=V_\text{dd} - V_0,~ E_\text{rev}^- =V_\text{gnd} - V_0,~t'=\frac{t}{T^\text{circ}}.
\end{flalign}
Here, $V_0$ denotes the resting membrane potential in the hardware (initial voltage value). 
We also assume $V_\text{dd} > V_0$ and $V_\text{gnd} < V_0$. 
By substituting these into Eq. (\ref{eq:resistively_coupled_model}) and reverting the time variable notation from $t'$ to $t$, we obtain 
\begin{flalign}
C_\text{m}V_\text{th}^\text{circ}\frac{d}{T^\text{circ}dt}v_i^{(l)} = \begin{cases}
\sum _{j=0}^{N^{(l-1)}-1} \sigma _{ij}^{(l)+}\left(E_\text{rev}^+ - V_\text{th}^\text{circ}v_i^{(l)}\right) \theta\left( t-t_j^{(l-1)}\right) + \sigma _{ij}^{(l)-}\left(E_\text{rev}^- - V_\text{th}^\text{circ}v_i^{(l)}\right) \theta \left( t-t_j^{(l-1)} \right) \\
~~~~~~~~~~~~~ \text{for } (l-1) \le t < l \text{ (accumulation phase),} \\
\frac{C_\text{m}V_\text{th}^\text{circ}}{T^\text{circ}},~\text{for } l \le t < (l+1) \text{ (firing phase),}
\end{cases} 
\end{flalign}
Simplifying this expression leads to 
\begin{flalign}
\frac{d}{dt}v_i^{(l)} = \begin{cases}
\sum _{j=0}^{N^{(l-1)}-1} \frac{T^\text{circ} E_\text{rev}^+\sigma _{ij}^{(l)+}}{C_\text{m}V_\text{th}^\text{circ}}\left(1 - \frac{V_\text{th}^\text{circ}}{E_\text{rev}^+}v_i^{(l)}\right) \theta\left( t-t_j^{(l-1)}\right) + \frac{T^\text{circ} E_\text{rev}^-\sigma _{ij}^{(l)-}}{C_\text{m}V_\text{th}^\text{circ}}\left(1 - \frac{V_\text{th}^\text{circ}}{E_\text{rev}^-}v_i^{(l)}\right) \theta \left( t-t_j^{(l-1)} \right) \\
~~~~~~~~~~~~~ \text{for } (l-1) \le t < l \text{ (accumulation phase)}, \\
1,~\text{for } l \le t < (l+1) \text{ (spike generation phase)}. 
\end{cases} 
\end{flalign}
By defining the synaptic weights as a transformation of conductances, 
\begin{flalign}
w_{ij}^{(l)\pm}&:=\frac{T^\text{circ} E_\text{rev}^\pm\sigma _{ij} ^{(l)\pm}}{C_\text{m}V_\text{th}^\text{circ}},  \label{eq:conductance_to_weight}
\end{flalign}
and introducing the following parameters, 
\begin{flalign}
\beta ^\pm&:=\frac{V_\text{th}^\text{circ}}{E_\text{rev}^\pm}, 
\end{flalign}
the time evolution of the membrane potential can be expressed as 
\begin{flalign}
\frac{d}{dt}v_i^{(l)} = \begin{cases}
\sum _{j=0}^{N^{(l-1)}-1} w_{ij}^{(l)+}\left(1 - \beta^+ v_i^{(l)}\right) \theta\left( t-t_j^{(l-1)}\right) + w_{ij}^{(l)-}\left(1 - \beta ^- v_i^{(l)}\right) \theta \left( t-t_j^{(l-1)} \right) \\
~~~~~~~~~~~~~ \text{for } (l-1) \le t < l \text{ (accumulation phase)} \\
1,~\text{for } l \le t < (l+1) \text{ (firing phase)}
\end{cases} 
\end{flalign}
Because either $w_{ij}^+$ or $w_{ij}^-$ is zero, we can define  
\begin{flalign}
w_{ij}^{(l)} &= \begin{cases}
    w_{ij}^{(l)+},~\text{for } w_{ij}^{(l)+}\ge0, \\
    w_{ij}^{(l)-},~\text{for } w_{ij}^{(l)+}<0,
\end{cases}\\
\beta_{ij}^{(l)} &:= \begin{cases}
\beta ^+ \text{ for } w_{ij}^{(l)} \ge 0, \\
\beta ^- \text{ for } w_{ij}^{(l)} < 0.
\end{cases} 
\end{flalign}
Thus, the membrane potential evolution simplifies to 
\begin{flalign}
\frac{d}{dt}v_i^{(l)} = \begin{cases}
\sum _{j=0}^{N^{(l-1)}-1} w_{ij}^{(l)}\left(1 - \beta_{ij}^{(l)} v_i^{(l)}\right) \theta\left( t-t_j^{(l-1)}\right)  \\
~~~~~~~~~~~~~ \text{for } (l-1) \le t < l \text{ (accumulation phase)} \\
1,~\text{for } l \le t < (l+1) \text{ (spike generation phase)}, 
\end{cases} 
\end{flalign}
which is consistent with the RC-Spike model. 

\subsection{Non-ideal current source model}

Consider a metal-oxide-semiconductor field-effect transistor (MOSFET) as a current source. 
While the MOSFET operates as a current source in the saturation region, 
it exhibits non-ideal characteristics owing to channel length modulation, as shown below \cite{Razavi}:
\begin{flalign}
I_\text{D}^\text{NMOS} \approx 
I \left( 1 + \lambda^N V_\text{DS} \right), \label{eq:CLM_NMOS}
\end{flalign}
where $I$ is the target current, $V_\text{DS}$ is the drain-source voltage, 
and $\lambda^N$  is a parameter representing the dependence of the current on $V_{DS}$ for the calse of an N-type MOSFET.  
Similarly, for a P-type MOSFET, the current is given by 
\begin{flalign}
I_\text{D}^\text{PMOS} \approx I \left(1 + \lambda^P V_\text{DS}\right). \label{eq:CLM_PMOS}
\end{flalign}
Thus, the MOSFET-based current source exhibits a linear dependence on the drain voltage, which corresponds to the membrane potential $V_i^{(l)}(t)$. 
Given that $V_\text{DS}$ can be regarded as positive (negative) the membrane potential for the case of N(P)-type MOSFETs, the time evolution of the membrane potential is described as follows:
\begin{flalign}
C_\text{m} \frac{d}{dt}V_i^{(l)}(t) = \begin{cases} \sum _j I_{ij}^{(l)P}\left(1 -\lambda ^P (V_i^{(l)}-V_0)\right)\theta\left(t-t_j^{(l-1)}\right) - I_{ij}^{(l)N}\left(1 +\lambda ^N (V_i^{(l)}-V_0) \right) \theta \left(t-t_j^{(l-1)}\right)  \\
-I^\text{dis} \left( 1 - \lambda ^\text{dis} \left(V_i^{(l)} - V_0 \right)\right),
\end{cases}
\end{flalign}
where $I_{ij}^{(l)P}$ represents the current flowing into the capacitor through the P-type MOSFET, and
$I_{ij}^{(l)N}$ represents the current flowing out of the capacitor through the N-type MOSFET. 
These currents are triggered when input spikes arrive, thereby closing the switch. 
In this case, we also consider the non-ideality of discharging current $I^\text{dis}$ represented with $\lambda^\text{dis}$. 
Voltage and time can be rescaled as follows:
\begin{flalign}
v_i^{(l)}=-\frac{ (V_i^{(l)}-V_0)}{V_\text{th}^\text{circ}},~t=T^\text{circ}t'.
\end{flalign}
Note that the membrane potential in the circuit is inverted in polarity. 
Returning to the original time variable $t$, 
the time evolution of the membrane potential is described as 
\begin{flalign}
-C_\text{m} V_\text{th}^\text{circ}\frac{d}{T^\text{circ}dt}v_i^{(l)}(t) &= \begin{cases}
\sum _j I_{ij}^{(l)P}\left(1 +\lambda ^P V_\text{th}^\text{circ}v_i^{(l)}\right)\theta\left(t-t_j^{(l-1)}\right) - I_{ij}^{(l)N}\left(1 -\lambda ^N V_\text{th}^\text{circ} v_i^{(l)} \right) \theta \left(t-t_j^{(l-1)}\right) \\
-I^\text{dis} \left( 1 - \lambda ^\text{dis} V_\text{th}^\text{circ} v_i^{(l)} \right).
\end{cases}
\end{flalign}
After some simplification, this becomes 
\begin{flalign}
\frac{d}{dt}v_i^{(l)} &= \begin{cases}
\sum _j \left[ \frac{T^\text{circ}I_{ij}^{(l)N}}{C_\text{m}V_\text{th}} ( 1 - \lambda ^N V_\text{th}^\text{circ}v_i^{(l)}) - \frac{T^\text{circ}I_{ij}^{(l)P}}{C_\text{m}V_\text{th}^\text{circ}} ( 1 + \lambda ^P V_\text{th}^\text{circ}v_i^{(l)}) \right] \theta (t - t_j^{(l-1)}) \\
\frac{T^\text{circ}I^\text{dis}}{C_\text{m}V_\text{th}} \left( 1 - \lambda ^\text{dis} V_\text{th} v_i^{(l)} \right). 
\end{cases} \label{eq:16}
\end{flalign}
Using the following variable definitions, 
\begin{flalign}
w_{ij}^{(l)+}:= \frac{T^\text{circ}I_{ij}^{(l)N}}{C_\text{m}V_\text{th}^\text{circ}},~
w_{ij}^{(l)-}:= -\frac{T^\text{circ}I_{ij}^{(l)P}}{C_\text{m}V_\text{th}^\text{circ}}  \label{eq:w_vs_I}\\
\beta^+ = V_\text{th}^\text{circ}\lambda ^N, ~ \beta^- = -V_\text{th}^\text{circ}\lambda ^P \label{eq:beta}\\
\alpha =\frac{T^\text{circ}I^\text{dis}}{C_\text{m}V_\text{th}^\text{circ}}   ,~ \beta_\text{charge} = \lambda ^\text{dis} V_\text{th}^\text{circ}, \label{eq:lambda_to_beta}
\end{flalign}
Eq. (\ref{eq:16}) can be rewritten as 
\begin{flalign}
\frac{d}{dt}v_i^{(l)} = \begin{cases}
 \sum _j \left[ w_{ij}^{(l)+} ( 1 - \beta ^+ v_i^{(l)}) + w_{ij}^{(l)-} ( 1 - \beta^- v_i^{(l)}) \right] \theta (t - t_j^{(l-1)}) \\
 \alpha \left( 1 - \beta ^\text{dis} v_i^{(l)}\right)
\end{cases}
\end{flalign}
Note that either $w_{ij}^{(l)+}$ or $w_{ij}^{(l)-}$ is zero.
Thus, by defining 
\begin{flalign}
w_{ij}^{(l)} &= \begin{cases}
    w_{ij}^{(l)+},~\text{for } w_{ij}^{(l)+}\ge0, \\
    w_{ij}^{(l)-},~\text{for } w_{ij}^{(l)+}<0, 
\end{cases}\\
\beta_{ij}^{(l)} &:= \begin{cases}
\beta ^+ \text{ for } w_{ij}^{(l)} \ge 0, \\
\beta ^- \text{ for } w_{ij}^{(l)} < 0, 
\end{cases} 
\end{flalign}
the equation is further simplified as:
\begin{flalign}
\frac{d}{dt}v_i^{(l)} = \begin{cases}
\sum _j w_{ij}^{(l)}\left(1 - \beta_{ij}^{(l)} v_i^{(l)}\right) \theta\left( t-t_j^{(l-1)}\right)  \\
~~~~~~~~~~~~~ \text{for } (l-1) \le t < l \text{ (accumulation phase)} \\
 \alpha \left( 1 - \beta ^\text{dis} v_i^{(l)}\right),~\text{for } l \le t < (l+1) \text{ (firing phase)}.
\end{cases} 
\end{flalign}
This is consistent with the RC-Spike model.

In the model-to-hardware mapping simulations, the RC decay effect was considered even in the firing phase ($\beta ^\text{dis} \neq 0 $). 
If the firing phase spans the interval [0,1] for computational simplicity, 
its solution is 
\begin{align}
v(t) = \frac{1- \left(1 - \beta v(0)\right) e^{-\alpha \beta t}}{\beta}.
\end{align}
Given the condition $v(1) = 1$ if $v(0)=0$ \footnote{Here, $t=0$ and $t=1$ represent the beginning and end time of the firing phase, respectively. 
The state $v(0)=0$, which can be obtained without input spikes, should produce a spike time of $1$, which does not influence the subsequent layers. }, we obtain $\alpha = - \frac{\ln \left(1-\beta\right)}{\beta}$, leading to 
\begin{align}
v(t) = \frac{1 - \left(1-\beta v(0)\right)\left(1-\beta \right)^t}{\beta}.
\end{align}
Finally, by solving $v(t) = 1$ for $t$, the spike time is 
\begin{align}
t &= \frac{\ln \left(\frac{1-\beta}{1-\beta v(0)}\right)}{\ln \left(1-\beta\right)} \\
&= \frac{\ln \left(1-\beta\right)  - \ln \left( 1-\beta v(0) \right) }{\ln \left(1-\beta\right)}.
\end{align}

\section{Proofs} \label{ss:proof}

The differentiable spike-time discretization (DSTD) offers a fast learning process and strong theoretical foundation. 
We demonstrate below that the time evolution obtained with DSTD converges to the rigorous solution of ordinary differential equations (ODEs) as $\Delta _\tau \rightarrow 0$, where $\Delta _\tau$ is the step size of DSTD. 

\subsection{Approximation Error Oder regarding the step size of DSTD} \label{ss:proof1}
~\\
{\bf Definition 1} (Interval Linear ODE)
For random time points $ \{t_0=0,t_1,t_2,\ldots ,t_{n},t_{n+1}=1\}$, the interval linear ODE is defined as
\begin{align}
\begin{aligned}
\dot{x} &= -f_i x + g_i,\quad \forall t \in [t_i , t_{i+1}),\quad  x(0) = 0\\
x_{i} &= x(t_{i}), \quad i \in  \{0,\ldots ,n\}.
\end{aligned}\label{Eq:interval_linear_ODE}
\end{align}

\begin{figure}
    \centering
    \includegraphics[width=0.5\linewidth]{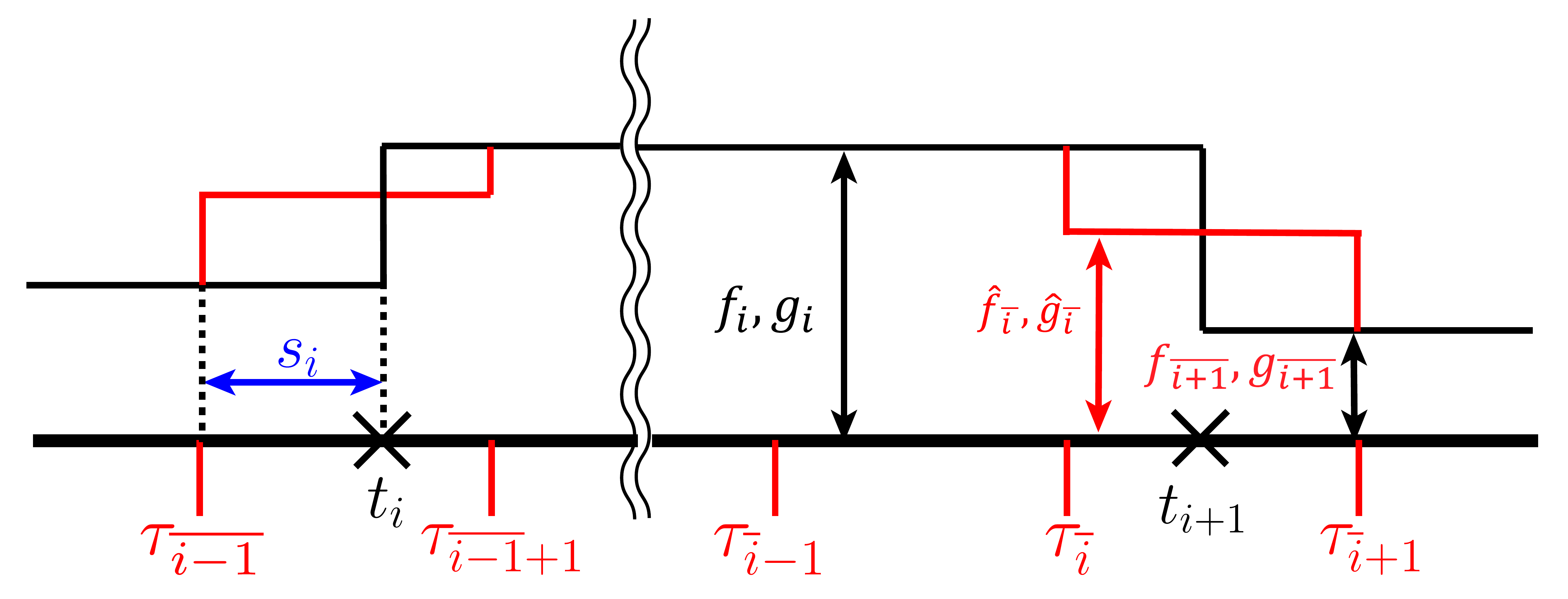}
    \caption{Schematic of random time points $t_i$ and regular time points $\tau_i$. 
    The time interval between two regular time points is constant and is denoted as $\Delta_\tau$. 
    The black solid line represents the values of $f_i$ or $b_i$, whereas the red solid line represents the approximate values of the parameters $\hat{f}_i$ or $\hat{g}_i$.
    }
    \label{fig:random2regular_time}
\end{figure}

\noindent
{\bf Definition 2} (Regular time points)
Suppose that regular time points with $\tau_i \triangleq \tfrac{i}{N} = i \Delta_\tau$ are defined as $\{\tau_0,\tau_1,\ldots ,\tau_N\}$.
The relation of random time points $\{t_0,t_1,\ldots ,t_{n+1}\}$ is denoted as
\begin{align*}
    \overline{i} &\triangleq \arg \max_j \{ \tau_j ~|~ \tau_j \leq t_{i+1} \},\quad
    N_i\triangleq   \overline{i} - \overline{i-1},\quad  s_{i} \triangleq t_i - \tau_{\overline{i-1}}
\end{align*}
\noindent
{\bf Lemma 1.} The following equations satisfy:
\begin{align}
s_0 = s_{n+1} = 0,\quad  t_{i+1} - t_i  =  \Delta_\tau N_i  + s_{i+1} - s_i.
\end{align}
\noindent
{\bf Proof:} The edges of regular time points are given by
\begin{align*}
\tau_{\overline{-1}} &=  \max_j \{ \tau_j ~|~ \tau_j \leq t_0 \} =  \tau_0 = 0, \\
\tau_{\overline{n}} &=  \max_j \{ \tau_j ~|~ \tau_j \leq t_{n+1} \} =  \tau_{N} = 1.
\end{align*}
Therefore, 
\begin{align*}
    s_0 &= t_0 - \tau_{\overline{-1}} = 0 - 0 = 0\\
    s_{n+1} &= t_{n+1} - \tau_{\overline{n}} = 1-1 = 0.
\end{align*} 
Furthermore,
\begin{align*}
t_{i+1} - t_i &= (s_{i+1} + \tau_{\overline{i}}) -  (s_{i} + \tau_{\overline{i -1}}) \\
&= \Delta_\tau N_i + s_{i+1} -  s_{i}
\end{align*}
$\hfill\square$

\noindent
{\bf Definition 3} (Approximated ODE). For regular time points $\{\tau_0,\tau_1,\ldots ,\tau_N\}$, the approximated ODE is defined as
\begin{align}
\begin{aligned}
\dot{\hat{x}} &= -\hat{f}_j \hat{x} + \hat{g}_j,\quad \forall t \in [\tau_j , \tau_{j+1}),\quad \hat{x}(\tau_j) = \hat{x}_j, \\
\quad \hat{x}_{j+1} &= \hat{x}(\tau_{j+1}),
\end{aligned}\label{Eq:Approximated_ODE}
\end{align}
where parameters $\hat{f}_i$ and $\hat{g}_i$ are approximated parameters of $f_i$ and $g_i$, respectively, defined as
\begin{align}
\begin{aligned}
\hat{f}_j &\triangleq 
\begin{cases}
    f_i &    j = \overline{i-1} +1 ,\overline{i} + 2 ,\ldots \overline{i} - 1 \\
    r_{i+1} f_{i} + (1 - r_{i+1}) f_{i+1} &    j = \overline{i}
\end{cases}\\
\hat{g}_j &\triangleq 
\begin{cases}
    g_i &    j = \overline{i-1} +1 ,\overline{i} + 2 ,\ldots \overline{i} - 1\\
    r_{i+1} g_{i} + (1 - r_{i+1}) g_{i+1} &    j = \overline{i}
\end{cases}\\
r_i &\triangleq \frac{ s_{i}}{\Delta_\tau}.
\end{aligned} \label{Eq:hat_coefficients}
\end{align}
\noindent
{\bf Lemma 2} (Solution of INterval Linear ODE). The solution of interval linear ODE (\ref{Eq:interval_linear_ODE}) and (\ref{Eq:Approximated_ODE})  is given by
\begin{align*}
x_{n+1} = \sum_{i = 0}^n \frac{Z_{i+1} - Z_{i}}{f_i} g_i ,\quad \hat{x}_{N} = \sum_{i = 0}^{N-1} \frac{\hat{Z}_{i+1} - \hat{Z}_{i}}{\hat{f}_i} \hat{g}_i,
\end{align*}
respectively, where $Z_i$ and $\hat{Z}_i$ are denoted as
\begin{align*}
Z_i \triangleq \prod_{j=i}^n e^{-f_j(t_{j+1} - t_j)}, \quad \hat{Z}_i \triangleq \prod_{j=i}^{N-1} e^{- \hat{f}_j\Delta_\tau}.
\end{align*}
Note that the product on the empty domain is defined as  1 (e.g.,  $\prod_{n}^{n-1} x_i = 1$).
\noindent
{\bf Proof:} Since these ODEs are of the same form, this study proves only the random interval time points case (\ref{Eq:interval_linear_ODE}).
The solution of linear ODEs for each interval $[t_i,t_{i+1})$ is given by
\begin{align*}
    x(t) &= e^{-f_i(t - t_i)} x_i  + (1 - e^{-f_i(t - t_i)})\frac{g_i}{f_i},\quad \forall t\in  [t_i, t_{i+1} ).
\end{align*}
The solution at each time points $\{t_0,t_1,\ldots ,t_{n+1}\}$ is given by
\begin{align*}
x(t_{i+1})&= x_{i+1} =  e^{-f_i(t_{i+1} - t_i)} x_i  + (1 - e^{-f_i(t_{i+1} - t_i)})\frac{g_i}{f_i}\\
&=  A_i x_i + C_i g_i,\quad  x_0 = x_1 = 0, 
\end{align*}
where $A_i \triangleq e^{-f_i(t_{i+1} - t_i)}$ and $C_i \triangleq \tfrac{1 - e^{-f_i(t_{i+1} - t_i)}}{f_i}$. 
Hence, the endpoint is given by
\begin{align*}
x(1) &= x_{n+1} =  A_n x_n + C_n g_n \\
&= A_n (A_{n-1} x_{n-1} + C_{n-1}g_{n-1}) + C_n g_n\\
&=\sum_{i=0}^n \prod_{j=i+1}^n A_j C_i g_i\\
&=\sum_{i=0}^n \frac{Z_{i+1} - Z_{i}}{f_i} g_i.
\end{align*}
Using the same approach, the solution of the approximate ODE can be calculated.
$\hfill\square$

\noindent
{\bf Lemma 3.} The formula between $Z_i$ and $\hat{Z}_i$ yields the following:
\begin{align}
\hat{Z}_{\overline{i}} &= Z_{i+1} e^{- r_{i+1}f_i\Delta_\tau},\quad \hat{Z}_{\overline{i}+1}= Z_{i+1} e^{(1 - r_{i+1})f_{i+1}\Delta_\tau}. \label{Eq:Z_Zhat_formulation}
\end{align} 
\noindent
{\bf Proof:} Firstly, this study shows the sum of $\hat{f}_i \Delta_\tau$.
Under $[\tau_{\overline{i-1}+1} \tau_{\overline{i}}]$, the sum is calculated as
\begin{align*}
\sum_{j = \overline{i-1}+1}^{\overline{i}} \hat{f}_k \Delta_\tau  &= \sum_{j = \overline{i-1}+1}^{\overline{i}-1} f_i \Delta_\tau + \hat{f}_{\overline{i}}\Delta_\tau\\
&= (N_i - 1) f_i\Delta_\tau  + r_{i+1} f_i\Delta_\tau + (1 - r_{i+1}) f_{i+1}\Delta_\tau\\
&= f_i \big( (t_{i+1} - t_i) - (s_{i+1} - s_{i})\big) - f_i (\Delta_\tau - s_{i + 1})  + f_{i+1}(\Delta_\tau - s_{i + 1}) \\
&= f_i(t_{i+1} - t_i)  - f_i(\Delta_\tau - s_i) +  f_{i+1}(\Delta_\tau - s_{i + 1})
\end{align*}
Hence, the sum of $\hat{f}_i \Delta_\tau$ between $\overline{i} + 1$ to $N$ is given by 
\begin{align*}
\sum_{j = \overline{c}+1}^{N-1}\hat{f}_j \Delta_\tau  &=  \sum_{i = c+1}^{n}\sum_{j = \overline{i-1}+1}^{\overline{i}} \hat{f}_j \Delta_\tau - \hat{f}_N \Delta_\tau\\
&=  \sum_{i = c+1}^{n} \left( f_i(t_{i+1} - t_i)  - f_i(\Delta_\tau - s_i) +  f_{i+1}(\Delta_\tau - s_{i + 1})\right) - f_{n+1} \Delta_\tau\\
&=\sum_{i = c+1}^{n} f_i(t_{i+1} - t_i) + f_{n+1}(\Delta_\tau - s_{n+ 1})  - f_{c+1} (\Delta_\tau - s_{c + 1}) - f_{n+1} \Delta_\tau\\
&=\sum_{i = c+1}^{n} f_i(t_{i+1} - t_i)  - f_{c+1} (\Delta_\tau - s_{c + 1})
\end{align*}
Also, the sum of $\overline{c}$ to $N-1$ is calculated as
\begin{align*}
\sum_{j = \overline{c}}^{N-1}\hat{f}_j \Delta_\tau  &=  \sum_{j = \overline{c}+1}^{N-1}\hat{f}_j \Delta_\tau + \hat{f}_{\overline{c}}\Delta_\tau \\
&=\sum_{i = c+1}^{n} f_i(t_{i+1} - t_i)  - f_{c+1} (\Delta_\tau - s_{c + 1}) +   f_cs_{c+ 1}  + f_{c+1}(\Delta_\tau-s_{c+1}) \\
&=\sum_{i = c+1}^{n} f_i(t_{i+1} - t_i)  +   f_cs_{c+ 1}
\end{align*}
Therefore, $\hat{Z}_{\overline{i}}$ and $\hat{Z}_{\overline{i}+ 1}$ is given by
\begin{align*}
\hat{Z}_{\overline{i}} &= \exp{ - \sum_{j = \overline{c}}^{N-1}\hat{f}_j \Delta_\tau } \\
&= \exp{ - \sum_{i = c+1}^{n} f_i(t_{i+1} - t_i)  -   f_cs_{c+ 1} }\\
&= Z_{c+1} e^{- f_c r_{c+1} \Delta_\tau}.\\
\hat{Z}_{\overline{i}} &= Z_{c+1} e^{f_{c+1} (1 - r_{c+1}) \Delta_\tau}.
\end{align*}
$\hfill\square$

Note that this lemma presents the map approximated ODE solution coefficient to the original ODE solution coefficient.
The following theorem estimates the order of error of the approximate ODE. 

\noindent
{\bf Theorem 1.}  The error between an interval linear ODE~(\ref{Eq:interval_linear_ODE}) and the approximated ODE~(\ref{Eq:Approximated_ODE}) satisfies
\begin{align}
    |x(1) - \hat{x}(1)| = \mathcal{O}(\Delta_\tau^2),
\end{align}
where $\mathcal{O}(\cdot)$ is the big $\mathcal{O}$ notion. 

\noindent
{\bf Proof:} The endpoint of the approximated ODE is written as
\begin{align*}
\hat{x}(1) &= \sum_{j=0}^{N-1} \frac{\hat{Z}_{j+1} - \hat{Z}_{j}}{\hat{f}_j} \hat{g}_j\\
&= \sum_{i=0 }^{n} \bigg(\frac{\hat{Z}_{\overline{i-1}+1} - \hat{Z}_{\overline{i-1}}}{\hat{f}_{\overline{i-1}}} \hat{g}_{\overline{i}} +  \sum_{j = \overline{i-1}+1}^{\overline{i}-1} \frac{\hat{Z}_{j+1} - \hat{Z}_{j}}{\hat{f}_j} \hat{g}_j \bigg)\\
&= \sum_{i=0 }^{n} \bigg(  \frac{\hat{Z}_{\overline{i-1}+1} - \hat{Z}_{\overline{i-1}}}{\hat{f}_{\overline{i-1}}} (r_{i} g_{i-1} + (1 - r_{i}) g_{i})+  \sum_{j = \overline{i-1}+1}^{\overline{i}-1} \frac{\hat{Z}_{j+1} - \hat{Z}_{j}}{f_i}g_i \bigg)\\
&= \sum_{i=0 }^{n} \Big(\frac{\hat{Z}_{\overline{i}} - \hat{Z}_{\overline{i-1}+1}}{f_i} +  r_{i+1}\frac{\hat{Z}_{\overline{i}+1} - \hat{Z}_{\overline{i}}}{\hat{f}_{\overline{i}}} + (1 - r_{i})\frac{\hat{Z}_{\overline{i-1}+1} - \hat{Z}_{\overline{i-1}}}{\hat{f}_{\overline{i-1}}} \Big)g_i.
\end{align*}
Using $Z_i$ and $\hat{Z}_i$ relation formulations~(\ref{Eq:Z_Zhat_formulation}), the approximated solution $\hat{x}(1)$ is given by
\begin{align*}
\hat{x}(1) &=  \sum_{i=0 }^{n} \Bigg[ \left(\frac{e^{-r_{i+1} f_i \Delta_\tau}}{f_i} +  r_{i+1}\frac{ e^{ (1-r_{i+1}) f_{i+1} \Delta_\tau} - e^{-r_{i+1} f_i \Delta_\tau}}{\hat{f}_{\overline{i}}} \right)Z_{i+1} \\
& - \left(\frac{e^{ (1-r_{i}) f_{i} \Delta_\tau}}{f_i} +  (1 - r_{i})\frac{ e^{-r_{i} f_{i-1} \Delta_\tau} - e^{ (1-r_{i}) f_{i} \Delta_\tau} }{\hat{f}_{\overline{i-1}}} \right)Z_{i}\Bigg]g_i.
\end{align*}
Hence, the error of the approximated solution is written as 
\begin{align*}
x(1) - \hat{x} (1) &= \sum_{i=0}^n \frac{Z_{i+1} - Z_{i}}{f_i} g_i - \sum_{j=0}^{N-1} \frac{\hat{Z}_{j+1} - \hat{Z}_{j}}{\hat{f}_j} \hat{g}_j \\ 
&= \sum_{i=0}^n \Big( u_i(\Delta_\tau) Z_{i+1} g_i  - v_i(\Delta_\tau) Z_{i} g_i \Big),
\end{align*}
where $u_i(\cdot)$ and $v_i(\cdot)$ is defined as
\begin{align*}
    u_i(\Delta_\tau) &\triangleq \frac{1 - e^{-r_{i+1} f_i \Delta_\tau}}{f_i} -  r_{i+1}\frac{ e^{ (1-r_{i+1}) f_{i+1} \Delta_\tau} - e^{-r_{i+1} f_i \Delta_\tau}}{\hat{f}_{\overline{i}}}  \\ 
    v_i(\Delta_\tau) &\triangleq \frac{1- e^{ (1-r_{i}) f_{i} \Delta_\tau}}{f_i} -  (1 - r_{i})\frac{ e^{-r_{i} f_{i-1} \Delta_\tau} - e^{ (1-r_{i}) f_{i} \Delta_\tau} }{\hat{f}_{\overline{i-1}}}.
\end{align*}
Here, the order of $u_i(\cdot)$ and $v_i(\cdot)$ are $\mathcal{O}(\Delta_\tau^2)$, respectively, because the limit of each function satisfies 
\footnote{Note that $\{r_i\in [0,1]\}_{i\in \mathcal{N}}$ are random variables that may be sampled differently for each value of $\Delta_\tau$. In the following argument, we treat these variables as independent constants and evaluate the worst case. By doing this, we can estimate the upper bound of the error.}
\begin{align*}
\lim_{\Delta_\tau \rightarrow 0} u_i(\Delta_\tau) &= \frac{1 - e^0}{f_i} - r_{i+1}\frac{e^0 - e^0}{\hat{f}_{\overline{i}}}=0,\\
\lim_{\Delta_\tau \rightarrow 0} v_i(\Delta_\tau) &= \frac{1 - e^0}{f_i} - (1 - r_{i})\frac{e^0 - e^0}{\hat{f}_{\overline{i-1}}}=0,\\ 
\lim_{\Delta_\tau \rightarrow 0} \frac{\partial u_i}{\partial \Delta_\tau}(\Delta_\tau) &= \frac{ f_i r_{i+1}}{f_i} - r_{i+1}\frac{\hat{f}_{\overline{i}}}{\hat{f}_{\overline{i}}} = 0, \\
\lim_{\Delta_\tau \rightarrow 0} \frac{\partial v_i}{\partial \Delta_\tau}(\Delta_\tau) &= \frac{ (1 - r_{i}) f_i}{f_i} - (1 - r_{i})\frac{\hat{f}_{\overline{i-1}}}{\hat{f}_{\overline{i-1}}} = 0. \\
\lim _{\Delta_\tau \rightarrow 0} \frac{\partial^2 u_i}{\partial \Delta_\tau^2}(\Delta_\tau) &= - r_{i+1}^2 f_i - r_{i+1}  \frac{(1-r_{i+1})^2 f_{i+1}^2  - r_{i+1}^2f_i^2 } {\hat{f}_{\overline{i}}}  \\
&= - r_{i+1}^2 f_i - r_{i+1}\left[(1-r_{i+1}) f_{i+1}  - r_{i+1} f_i \right] < \infty \\
\lim _{\Delta_\tau \rightarrow 0} \frac{\partial^2 v_i}{\partial \Delta _\tau^2}(\Delta_\tau) &= -(1-r_i)^2 f_i + (1-r_i) \frac{ -r_{i}^2 f_{i-1}^2  + (1-r_{i})^2 f_{i}^2  }{\hat{f}_{\overline{i-1}}}  \\
&=-(1-r_i)^2 f_i - (1-r_i) \left[ r_{i} f_{i-1}  - (1-r_{i}) f_{i} \right] < \infty
\end{align*}
According to L'Hôpital's rule, the following result is established
\begin{align*}
\lim_{\Delta_\tau \rightarrow 0} \frac{u_i(\Delta_\tau)}{\Delta_\tau^2} = \frac{1}{2} \lim_{\Delta_\tau \rightarrow 0} \frac{\partial^2 u_i}{\partial \Delta_\tau^2}(\Delta_\tau) < \infty \\ 
\lim_{\Delta_\tau \rightarrow 0} \frac{v_i(\Delta_\tau)}{\Delta_\tau^2} = \frac{1}{2} \lim_{\Delta_\tau \rightarrow 0} \frac{\partial^2 v_i}{\partial \Delta_\tau^2}(\Delta_\tau) < \infty.
\end{align*}
The absolute error is written as
\begin{align*}
    \frac{|x(1) - \hat{x} (1)|}{\Delta_\tau^2}  
    &= \frac{\left| \sum_{i=0}^n \Big( u_i(\Delta_\tau) Z_{i+1} g_i  - v_i(\Delta_\tau) Z_{i} g_i \Big)\right|}{\Delta_\tau^2}\\
    &\leq  \sum_{i=0}^n \left|\frac{ u_i(\Delta_\tau)}{\Delta_\tau^2}\right|  |Z_{i+1}| |g_i|  +  \left|\frac{v_i(\Delta_\tau)}{\Delta_\tau^2}\right| |Z_{i}| |g_i| \\
    &\leq (n+1) \max_i | g_i| \left( \max_i \left|\frac{u_i(\Delta_\tau)}{\Delta_\tau^2}\right| +  \max_i \left| \frac{v_i(\Delta_\tau)}{\Delta_\tau^2}\right|\right) < \infty
\end{align*}
Therefore,  the absolute error is $\mathcal{O}(\Delta_\tau^2)$. 
$\hfill\square$

{\bf Remark 1.} In the case where $x$ is multidimensional, the error order is proportional to  the dimension of the variable, but the same property holds for $\Delta_\tau$.

{\bf Remark 2.} This result assumes that the number of random time points $\{t_1,\ldots,t_n\}$ in  each regular time interval $[\tau_i, \tau_{i+1})$ is one or less.
This assumption is satisfactorily satisfied as $N$ increases.
Furthermore, similar results can be obtained by generalizing the constitutive rules of the approximation parameters $\hat{f}_i$ and $\hat{g}_i$.

\subsection{Approximation Error Oder regarding reversal potentials}

We study the case of $\alpha = 0$ and $E_\text{rev}^+ = - E_\text{rev}^-$.
In this case, the RC-Spike model is expressed as 
\begin{align}
\begin{aligned}
\dot{x} &= -\beta f_i x + g_i,\quad \forall t \in [t_i , t_{i+1}),\quad  x(0) = 0\\
x_{i} &= x(t_{i}), \quad i \in  \{0,\ldots ,n\}.
\end{aligned}\label{Eq:interval_linear_ODE_beta}
\end{align}
Here, $\beta = \frac{1}{|E_\text{rev}^\pm|}$.
Given that the analysis in Subsection~\ref{ss:proof1} remains valid when substituting $f_i$ with $\beta f_i$, we redefine the variables accordingly:
\begin{align*}
    u_i(\beta) &\triangleq \frac{1 - e^{-\beta r_{i+1} f_i \Delta_\tau}}{\beta f_i}
    -  r_{i+1}\frac{ e^{ \beta (1-r_{i+1}) f_{i+1} \Delta_\tau} - e^{-\beta r_{i+1} f_i \Delta_\tau}}{\beta \hat{f}_{\overline{i}}}  \\ 
    v_i(\beta) &\triangleq \frac{1- e^{ \beta (1-r_{i}) f_{i} \Delta_\tau}}{\beta f_i} 
    -  (1 - r_{i})\frac{ e^{-\beta r_{i} f_{i-1} \Delta_\tau} - e^{ \beta (1-r_{i}) f_{i} \Delta_\tau} }{\beta \hat{f}_{\overline{i-1}}}, \\
    Z_i(\beta) &\triangleq \prod _{j=i}^n e^{-\beta f_j(t_{j+1}-t_j)}.
\end{align*}
First, the following limit can be derived
\begin{align}
    \lim_{\beta \rightarrow 0} u_i\left(\beta\right) &= r_{i+1}\Delta _t - r_{i+1} \frac{(1-r_{i+1})f_{i+1}\Delta _\tau + r_{i+1}f_i\Delta _\tau}{r_{i+1}f_i + (1-r_{i+1})f_{i+1}}=0, \\
    \lim_{\beta \rightarrow 0} v_i\left(\beta \right) &= (1-r_i)\Delta_\tau - (1-r_i) \frac{r_if_{i-1}\Delta_\tau + (1-r_i)f_i\Delta_\tau}{r_if_{i-1}+(1-r_i)f_i} = 0, \\
    \lim_{\beta \rightarrow 0}Z_i(\beta) &= 1.
\end{align}

Next, we calculate the limits of the derivatives of $u_i$ and $v_i$ using L'Hôpital's rule:
\begin{align}
&\lim_{\beta\rightarrow 0}\frac{\partial u_i}{\partial \beta}(\beta) = \lim_{\beta\rightarrow 0}\Big[  \frac{\beta r_{i+1}f_i\Delta_\tau e^{-\beta r_{i+1}f_i\Delta_\tau} - \left(1 - e^{-\beta r_{i+1} f_i \Delta_\tau}\right)}{ f_i} \nonumber \\
&- r_{i+1}\frac{\beta \Delta _\tau \left[(1-r_{i+1})f_{i+1} e^{ \beta (1-r_{i+1}) f_{i+1} \Delta_\tau} + r_{i+1}f_i e^{-\beta r_{i+1} f_i \Delta_\tau}\right] 
- \left[e^{ \beta (1-r_{i+1}) f_{i+1} \Delta_\tau} - e^{-\beta r_{i+1} f_i \Delta_\tau}\right]}
{\hat{f}_{\overline{i}}} \Big] \beta^{-2} \\
&=\lim_{\beta\rightarrow 0}\Bigg[  \frac{ r_{i+1}f_i\Delta_\tau e^{-\beta r_{i+1}f_i\Delta_\tau}  -  \beta r_{i+1}^2f_i^2\Delta_\tau^2 e^{-\beta r_{i+1}f_i\Delta_\tau} -  r_{i+1} f_i \Delta _\tau e^{-\beta r_{i+1} f_i \Delta_\tau} }{ f_i} \nonumber \\
&- \frac{r_{i+1}}{\hat{f}_{\overline{i}}} \Bigg\{ \Delta_\tau  \left[(1-r_{i+1})f_{i+1} e^{ \beta (1-r_{i+1}) f_{i+1} \Delta_\tau} + r_{i+1}f_i e^{-\beta r_{i+1} f_i \Delta_\tau}\right] \nonumber \\
&+ \beta \Delta_\tau^2  \left[(1-r_{i+1})^2f_{i+1}^2 e^{ \beta (1-r_{i+1}) f_{i+1} \Delta_\tau} - r_{i+1}^2f_i^2 e^{-\beta r_{i+1} f_i \Delta_\tau}\right] \nonumber \\
&- \Delta_\tau\left[(1-r_{i+1})f_{i+1} e^{ \beta (1-r_{i+1}) f_{i+1} \Delta_\tau} + r_{i+1}f_i e^{-\beta r_{i+1} f_i \Delta_\tau} \Delta_\tau \right]\Bigg\} \Bigg] (2\beta)^{-1}  \\
&=\lim_{\beta\rightarrow 0}\frac{  -  \beta r_{i+1}^2f_i\Delta_\tau^2 e^{-\beta r_{i+1}f_i\Delta_\tau} 
- \frac{r_{i+1}\beta \Delta_\tau^2 }{\hat{f}_{\overline{i}}} \left\{(1-r_{i+1})^2f_{i+1}^2 e^{ \beta (1-r_{i+1}) f_{i+1} \Delta_\tau} - r_{i+1}^2f_i^2 e^{-\beta r_{i+1} f_i \Delta_\tau}\right\} }{2\beta} \\
&= \frac{ r_{i+1}^2f_i\Delta_\tau^2  }{2} - \frac{r_{i+1}\Delta_\tau^2}{2} \left[(1-r_{i+1})f_{i+1} - r_{i+1}f_i \right] < \infty.
\end{align}
Similarly, we obtain 
\begin{align}
\lim_{\beta\rightarrow 0}    \frac{\partial v_i}{\partial \beta} &=\lim_{\beta\rightarrow 0} \Bigg[ \frac{-\beta (1-r_i)f_i\Delta_\tau e^{\beta(1-r_i)f_i\Delta_\tau} - (1 - e^{\beta(1-r_i)f_i\Delta_\tau}) }{f_i}  \nonumber \\
&- \frac{1-r_i}{\hat{f}_{\overline{i}}}\bigg\{ \beta \Delta _\tau\left[ -r_if_{i-1}e^{-\beta r_{i} f_{i-1} \Delta_\tau} - (1-r_i)f_i e^{ \beta (1-r_{i}) f_{i} \Delta_\tau}\right] \nonumber \\
&  - \left[  e^{-\beta r_{i} f_{i-1} \Delta_\tau} - e^{ \beta (1-r_{i}) f_{i} \Delta_\tau}\right]  \bigg\}  \Bigg]\beta^{-2} \nonumber \\
& = \lim_{\beta\rightarrow 0} \Bigg[  \frac{-(1-r_i)f_i\Delta_\tau e^{\beta(1-r_i)f_i\Delta_\tau} 
- \beta (1-r_i)^2f_i^2\Delta_\tau^2 e^{\beta(1-r_i)f_i\Delta_\tau} + (1-r_i)f_i\Delta_\tau e^{\beta(1-r_i)f_i\Delta_\tau})}{f_i}  \\
&- \frac{1-r_i}{\hat{f}_{\overline{i}}} \bigg\{ \Delta _\tau\left[ -r_if_{i-1}e^{-\beta r_{i} f_{i-1} \Delta_\tau} - (1-r_i)f_i e^{ \beta (1-r_{i}) f_{i} \Delta_\tau}\right]  \\
&+ \beta \Delta_\tau  \left[ r_i^2f_{i-1}^2e^{-\beta r_{i} f_{i-1} \Delta_\tau} - (1-r_i)^2f_i^2 e^{ \beta (1-r_{i}) f_{i} \Delta_\tau}\right] \nonumber \\
&+ \left[ r_if_{i-1}\Delta_\tau e^{-\beta r_{i} f_{i-1} \Delta_\tau} + (1-r_i)f_i\Delta_\tau e^{ \beta (1-r_{i}) f_{i} \Delta_\tau}\right]  \bigg\}\Bigg] (2\beta)^{-1} \nonumber \\
&= \lim_{\beta \rightarrow 0} \frac{- \beta (1-r_i)^2f_i\Delta_\tau^2 e^{\beta(1-r_i)\Delta_\tau} - \frac{(1-r_i)\beta \Delta_\tau^2}{\hat{f}_{\overline{i}}} \left[ r_i^2f_{i-1}^2e^{-\beta r_{i} f_{i-1} \Delta_\tau} - (1-r_i)^2f_i^2 e^{ \beta (1-r_{i}) f_{i} \Delta_\tau} \right]}{2\beta} \\
&= -\frac{(1-r_i)^2f_i\Delta_\tau^2}{2} - \frac{(1-r_i)\Delta_\tau^2}{2} \left(r_i f_{i-1} - (1-r_i)f_i \right) < \infty.
\end{align}

The absolute error is estimated as
\begin{align*}
\lim_{\beta \rightarrow 0} \frac{|x(1) - \hat{x} (1)|}{\beta}  &= \left| \sum_{i=0}^n \Big( \frac{u_i(\beta)}{\beta} Z_{i+1}(\beta) g_i  - \frac{v_i(\beta)}{\beta} Z_{i}(\beta) g_i \Big)\right|\\
    &\leq  \sum_{i=0}^n \left|\frac{ u_i(\beta)}{\beta}\right|  \left|Z_{i+1}(\beta)\right| |g_i|  +  \left|\frac{g_i(\beta)}{\beta}\right| \left|Z_{i}(\beta)\right| |g_i| \\
    &\leq (n+1) \max_i | g_i| \left( \max_i \left|\frac{u_i(\beta)}{\beta}\right| +  \max_i \left|\frac{g_i(\beta)}{\beta}\right|\right) < \infty.
\end{align*}
Therefore, the absolute error is $\mathcal{O}(\beta)$. Note that $\beta = |E_\text{rev}^\pm|^{-1}$.

\section{Simulation results for TTFS-SNNs}

In time-to-first-spike-coded spiking neural networks (TTFS-SNNs), neurons in each layer fire upon reaching the firing threshold $V_\text{th}$ in response to incoming spike signals, after which the membrane potential is fixed to zero. 
Consequently, each neuron is limited to firing at most once. 
TTFS-SNNs enable highly efficient information processing with minimal spike activity, 
and supports ultra-low-latency inference by allowing classification tasks to be completed once the first spike is generated in the output layer \cite{Mostafa2018supervised,Comsa2020temporal,Goltz2021fast,Sakemi2023supervised}. 
However, neuron models incorporating reversal potentials remain unexplored enough in the context of TTFS-SNNs \cite{Sakemi2023supervised}. 
As demonstrated in RC-Spike models, the use of differentiable spike-time discretization (DSTD) facilitates efficient computation in TTFS-SNNs. 
We applied DSTD to train a TTFS-SNN on the Fashion-MNIST dataset \cite{Xiao2017fashion}. 
The network architecture employed was Conv(16)-Conv(16)-PL-Conv(32)-Conv(32)-PL-128-10, where Conv($n$) denotes a convolutional layer with $n$ channels, a $3\times3$ kernel, and a stride of 1, while PL represents a pooling layer with a $2\times2$ kernel. Training was conducted with a mini-batch size of 32 over 150 epochs. 

In TTFS-SNNs, the synaptic weights must be initialized properly, as learning cannot proceed without neuronal firing.  
For instance, Comsa et al. \cite{Comsa2020temporal} employed a shifted Glorot initial weight distribution for this purpose. 
In the following, we derive an initial weight distribution that minimally affects the firing time distribution of the network when the reversal potential $E_\text{rev}^\pm$ is varied. 
We assumed that $N$ input spikes are uniformly sampled from the time interval $[0,1]$. Additionally, the synaptic weights are assumed to follow a uniform distribution over the interval $[0, w_\text{max}/N]$. 
Under these assumptions, the temporal evolution of the non-leaky membrane potential is described as 
\begin{flalign}
\frac{d}{dt} v(t) &= \left( 1 - \beta v(t)\right) W(t), \\
W(t) &:= \sum_{j \in \Gamma(t)} \frac{w_j}{N}.
\end{flalign}
where we adopted a neuron model that omits leak currents and synaptic decay \cite{Sakemi2023supervised}, and $\Gamma(t)$ denotes the set of spikes received up to time $t$. 
In addition, we assume $\beta=|E_\text{rev}^\pm|^{-1}$. 
In the limit as $N \rightarrow \infty$, $W(t)$ converges to $\frac{w_\text{max} t}{2}$ owing to the law of large numbers
\footnote{The expected number of spikes up to time $t$ is $Nt$. 
Therefore, the expected total weight up to time $t$ is  $\mathbb{E}\left[\sum_{j\in\Gamma(t)}w_j \right] = Nt \times \mathbb{E}\left[w_j\right]=Nt \times \frac{w_\text{max}}{2}$ because input spike timing and weights are independent random variables.  
Thus, we obtain $\mathbb{E}\left[W(t)\right] = \frac{w_\text{max}t}{2}$.
As $N\rightarrow \infty$, the actual sum $W(t)$ converges to its expected value owing to the law of large numbers. 
}.
By replacing $W(t)$ with $\frac{w_\text{max}t}{2}$, solving the differential equation yields the following expression for the membrane potential:
\begin{flalign}
v(t) &= \frac{1 - \exp(-\frac{ w_\text{max} \beta t^2}{4})}{\beta}.
\end{flalign}

To achieve an average delay time per layer of $t^\text{ref}/L$, and with the firing threshold set to 1, the following condition for the maximum synaptic weight is derived:
\begin{flalign}
w_\text{max} &= - 4w^*\frac{L^2}{\beta (t^\text{ref})^2} \ln(1 - \beta),
\end{flalign}
where $w^*$ is a positive constant that adjusts the actual distribution. 
We set $w^* = 2$. 
Note that in TTFS-SNNs, the shape of the firing time distribution undergoes significant changes in subsequent layers, which presents challenges. However, applying a factor of $\frac{\ln(1 - \beta)}{\beta}$ provided a reasonable correction.

\begin{figure*}
\centering
\includegraphics[clip, width=\textwidth]{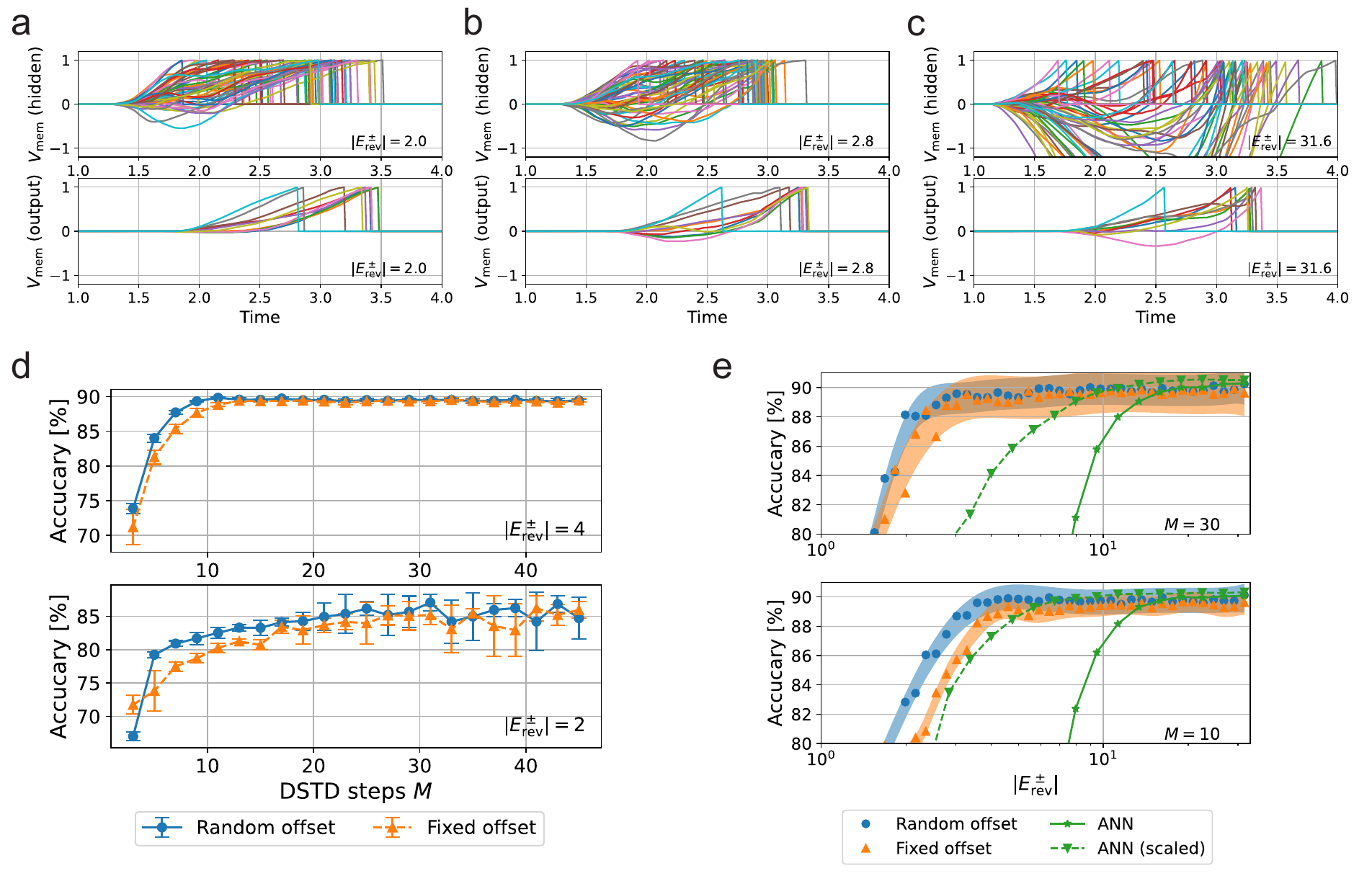}
\caption{{\bf Learning results for convolutional TTFS-SNN models with reversal potentials on Fashion-MNIST dataset.} 
{\bf a-c.} The time evolution of membrane potentials in the trained network. 
The top panels show results from the final hidden layer, while the bottom panels display the output layer. 
The values of $|E_\text{rev}^\pm|$ are set to 2.0 ({\bf a}), 2.8 ({\bf b}), and 31.6 ({\bf c}). 
{\bf d.} Recognition accuracy as a function of the number of discretized time steps $M$ in differentiable spike-time discretization (DSTD). 
The reversal potentials are set to $E_\text{rev}^+ = -E_\text{rev}^-$. 
The top panel corresponds to $E_\text{rev}^+ = 4$, and the bottom panel to $E_\text{rev}^+ = 2$. 
Each case compares the results of randomizing the offset during training (random offset) and those without randomizing the offset (fixed offset).   
{\bf e.} Recognition accuracy for varying $E_\text{rev}^+ (=-E_\text{rev}^-)$ between 1.2 and 31.6. 
The top panel shows results for DSTD steps of 30, while the bottom panel shows results for DSTD steps of 10. 
Both random and fixed offset results are included in each case.  Additionally, for the model trained with $E_\text{rev}^+ = 31.6$ under random offset, the test results for varying $E_\text{rev}^+$ are also shown (ANN). 
Furthermore, the results of model optimization via rescaling of both positive and negative weights by 1\%  are presented for each $E_\text{rev}^+$ (ANN (scaled)).
}
\label{fig:TTFS_VGG}
\end{figure*}

Figure \ref{fig:TTFS_VGG} panels {\bf a-c} display the temporal dynamics of membrane potentials after training. 
The results correspond to three different reversal potentials: $|E_\text{rev}^\pm|=2$ ({\bf a}), $|E_\text{rev}^\pm|=2.8$ ({\bf b}), and $|E_\text{rev}^\pm|=31.6$ ({\bf c}). 
In each figure, the upper and lower panels represent the membrane potential in the final fully connected layer and output layer, respectively.  
For all experiments shown in Figure \ref{fig:TTFS_VGG}, we set $E_\text{rev}^- = - E_\text{rev}^+$. 
The results indicate that a smaller $|E_\text{rev}^\pm|$ prevents the membrane potential in the final fully connected layer from becoming excessively negative, and a similar trend was observed in the output layer. This behavior is likely a result of $E_\text{rev}^+$ and $E_\text{rev}^-$ acting as the upper and lower bounds of the membrane potential, respectively.

Figure \ref{fig:TTFS_VGG} {\bf d} shows the effect of varying DSTD steps $M$ on recognition accuracy. 
The upper and lower panels present the results for $|E_\text{rev}^+|=4$ and $|E_\text{rev}^\pm|=2$, respectively.  
Each panel compares the performance between randomized DSTD offset (random offset) and fixed offset conditions. 
The results are expressed as the mean and standard deviation of five independent trials with random weight initialization. 
For $|E_\text{rev}^+|=4$, convergence was achieved with approximately 20 DSTD steps, whereas for $E_\text{rev}^+=2$, convergence required approximately 40 steps. 
This observation is explained by the stronger nonlinearity associated with smaller $|E_\text{rev}^\pm|$, which degrades approximation of DSTD. 
Additionally, for smaller DSTD steps ($M<10$), the random offset condition consistently yielded better recognition performance than the fixed offset. 
Therefore, random offset may improve the gradient characteristics during DSTD calculations. 
Further investigation into the nature of these gradients will be addressed in future work.

Figure \ref{fig:TTFS_VGG} {\bf e} illustrates the variation in recognition accuracy as a function of $|E_\text{rev}^\pm|$. 
The upper and lower panels display results for DSTD steps $M=30$ and $M=10$, respectively.  
In both panels, the results for random and fixed offsets are depicted. 
For $M=30$, accuracy converged at approximately $|E_\text{rev}^\pm|=3$, whereas for $M=10$, convergence occurred at $|E_\text{rev}^\pm|=4$. 
Furthermore, in both cases, the use of random offsets led to faster convergence (i.e., at smaller $M$ values) compared with fixed offsets. 
When the reversal potential became large, the model behaved similarly to a neuron model that does not account for reversal potentials. 
Accordingly, a model trained with $|E_\text{rev}^\pm|=32$ was employed in the test phase with varying $E_\text{rev}^\pm$ values (ANN)\footnote{This is not an ANN model, but we call this model ANN for convenience.}. 
This procedure is analogous to mapping a model trained without considering hardware non-ideal characteristics onto real hardware. 
In this case, the performance significantly deteriorated when $|E_\text{rev}^\pm|$ fell below approximately 10.
To reduce the impact of these non-ideal characteristics, the positive and negative weight scales were optimized for each $E_\text{rev}^\pm$ (ANN (scaled)) via grid search with magnification increments of 0.01. 
After optimizing these scales, recognition accuracy improved markedly. 
However, the performance remained substantially inferior compared with that of models trained explicitly considering reversal potentials.

\section{Effects of spike timing noise on RC-Spike models}

\begin{figure*}
\centering
\includegraphics[clip, width=\textwidth]{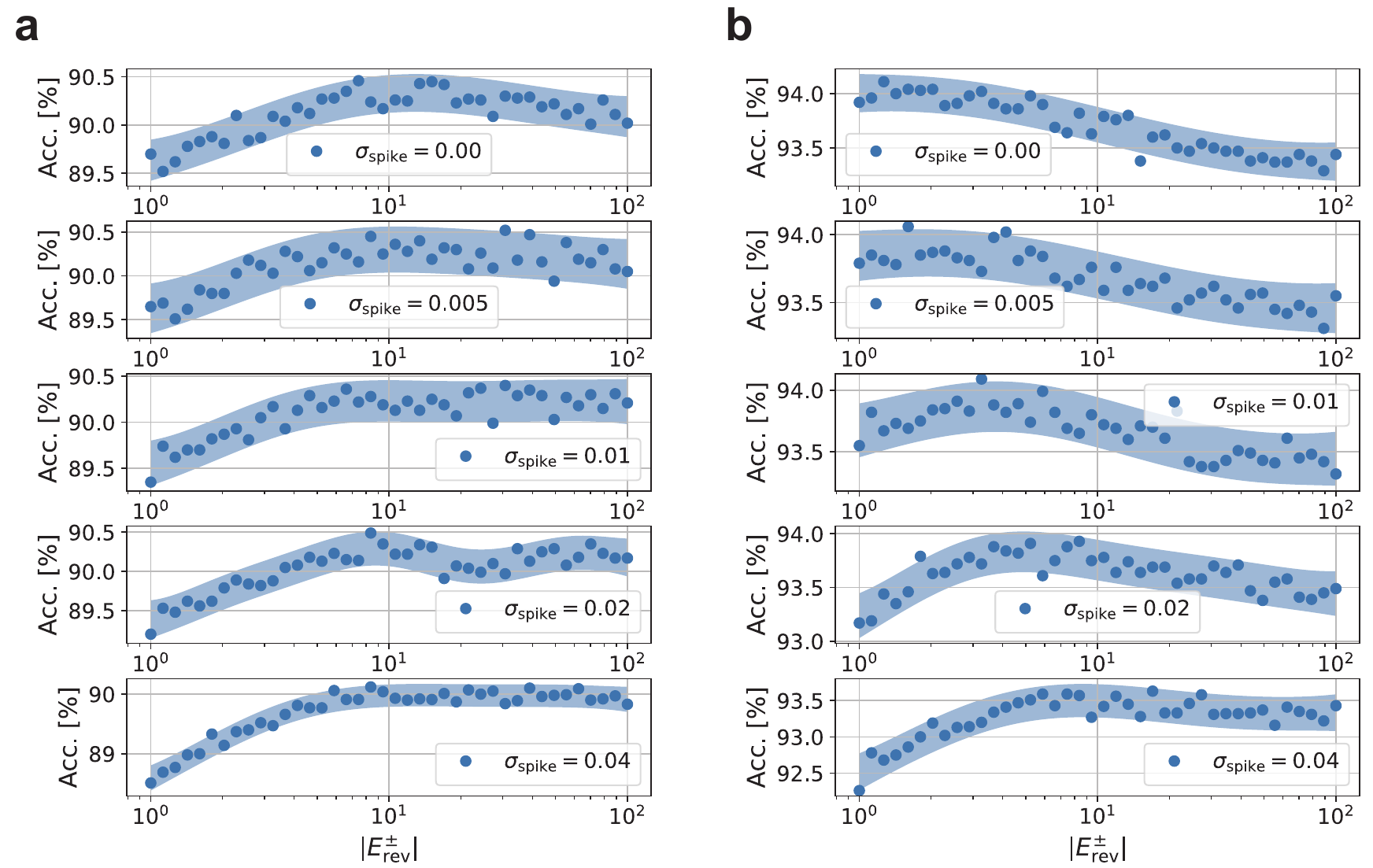}
\caption{Dependence of the recognition performance of the RC-Spike model on $|E_\text{rev}^\pm|$ trained with Fashion-MNIST dataset under various noise intensities $\sigma_\text{spike}$.
{\bf a.} Performance of the fully connected RC-Spike model (784-400-400-10).
{\bf b.} Performance of the convolutional RC-Spike models.
In both {\bf a.} and {\bf b.}, each panel represents results obtained under different noise standard deviations $\sigma_\text{spike}$.
For all cases, we used DSTD steps of $10$ with random offset. 
}
\label{fig:RC_Spike_ANN_noise}
\end{figure*}

The RC-Spike model exhibited a tendency to learn small membrane potentials when the reversal potential $|E_\text{rev}^\pm|$ was small, as a means to mitigate its influence. 
However, in practical hardware implementations, the precision of membrane potentials is inherently limited. 
As a result, using small membrane potentials may lead to significant performance degradation. 
To address this limitation, noise was introduced into the spike timing during performance evaluation.  

Figure \ref{fig:RC_Spike_ANN_noise} illustrates the recognition performance of the RC-Spike model as a function of $|E_\text{rev}^\pm|$ across different spike timing noise.
The noise is modeled as Gaussian, with a mean of 0 and a standard deviation of $\sigma_\text{spike}$. 
We trained the models using DSTD steps of $10$ with random offset. 
Figure \ref{fig:RC_Spike_ANN_noise} {\bf a} presents the results obtained from a fully connected network (784-400-400-10) trained on the Fashion-MNIST dataset. 
Across all noise levels, performance degraded when $|E_\text{rev}^\pm|$ was extremely small ($\sim 1$), which corresponds to pronounced non-ideal characteristics in hardware. 
As $|E_\text{rev}^\pm|$ increased, the model performance improved in an almost monotonic fashion. 
Notably, in the absence of noise ($\sigma_\text{spike}=0.00$), performance peaked at $|E_\text{rev}^\pm|\sim10$.

Figure \ref{fig:RC_Spike_ANN_noise} {\bf b} shows the results for a convolutional network trained on the Fashion-MNIST dataset. 
The network architecture consists of Conv(64)-Conv(64)-PL-Conv(128)-Conv(128)-PL-Conv(256)-Conv(256)-PL-512-10, where Conv($N$) represents a convolutional layer with a $3\times3$ kernel, $N$ output channels, and a stride of 1, while PL denotes a pooling layer with a $2\times 2$ kernel and a stride of 2. 
When the noise intensity was high ($\sigma_\text{spike}=0.04$), the performance, similar to the fully connected network results, improved almost monotonically as $|E_\text{rev}^\pm|$ increased. 
However, under moderate noise conditions, performance was maximized at a specific value of $|E_\text{rev}^\pm|$. 
Moreover, with lower noise intensities, the maximum performance was achieved at smaller values of $|E_\text{rev}^\pm|$.

\begin{figure*}
\centering
\includegraphics[clip, width=\textwidth]{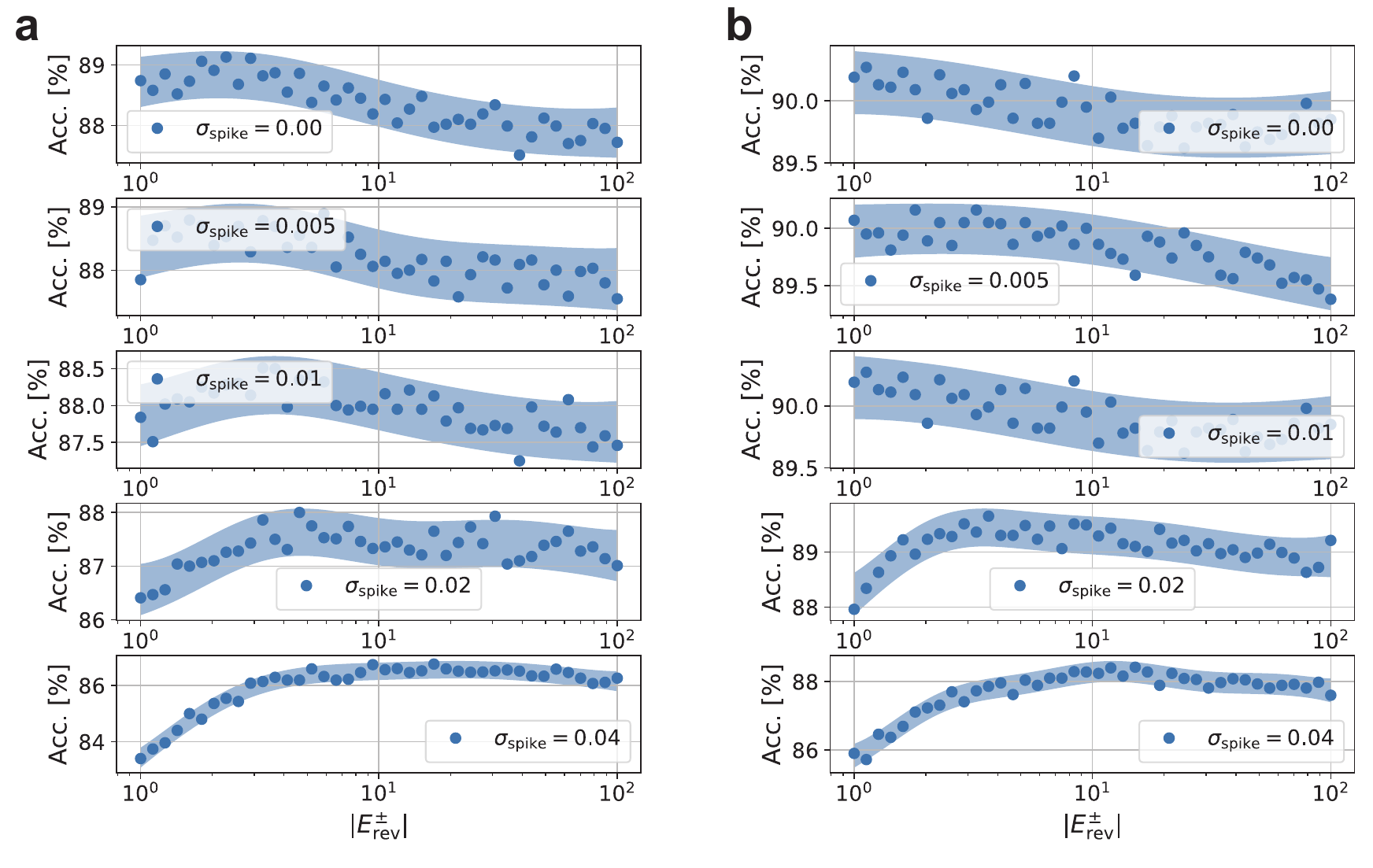}
\caption{Dependence of the recognition performance of the RC-Spike model on $|E_\text{rev}^\pm|$ trained with CIFAR-10 dataset under various noise intensities. We used DSTD steps of $10$ with random offset. 
{\bf a.} The performance results for the VGG7 architecture. {\bf b.} The performance results for large-width VGG7 architecture. 
}
\label{fig:RC_Spike_ANN_noise_2}
\end{figure*}

As shown in Fig. \ref{fig:RC_Spike_ANN_noise_2} {\bf a}, we present an additional simulation using the same convolutional network trained on the CIFAR-10 dataset. Consistent with the findings in Fig. \ref{fig:RC_Spike_ANN_noise}, lower noise intensities tended to yield peak performance at smaller values of $|E_\text{rev}^\pm|$. 
Furthermore, comparable results are observed with a convolutional network of doubled channel width, as illustrated in Fig. \ref{fig:RC_Spike_ANN_noise_2} {\bf b}.

\begin{table} 
\begin{center}
\caption{Performance results for RC-Spike models. CNN architecture refers to Conv(64)-Conv(64)-PL-Conv(128)-Conv(128)-PL-Conv(256)-Conv(256)-PL-512-10, where Conv($N$) represents a convolutional layer with a $3\times3$ kernel, $N$ output channels, and a stride of 1, while PL denotes a pooling layer with a $2\times 2$ kernel and a stride of 2.}
\begin{tabular}{lllll}
Dataset & Architecture & $\sigma_\text{spike}$  & $|E_\text{rev}^\pm|$ & Accuracy \\ \hline
Fashion-MNIST & -400-400-10 & 0.00 & 7.4 & 90.46 \%\\
Fashion-MNIST & -400-400-10 & 0.005 & 30.7 & 90.52 \%\\
Fashion-MNIST & -400-400-10 & 0.01 & 30.7 & 90.40 \%\\
Fashion-MNIST & -400-400-10 & 0.02 & 8.4 & 90.49 \%\\
Fashion-MNIST & -400-400-10 & 0.04 & 8.4 & 90.12 \%\\
Fashion-MNIST & CNN & 0.00 & 1.3 & 94.11 \% \\
Fashion-MNIST & CNN & 0.005 & 1.6 & 94.06 \% \\
Fashion-MNIST & CNN & 0.01 & 3.3 & 94.09 \% \\
Fashion-MNIST & CNN & 0.02 & 8.4 & 93.93 \% \\
Fashion-MNIST & CNN & 0.04 & 17.0 & 93.63 \% \\ 
CIFAR-10 & CNN & 0.00 & 2.3 & 89.13 \% \\
CIFAR-10 & CNN & 0.005 & 5.9 & 88.89 \% \\
CIFAR-10 & CNN & 0.01  & 3.3 & 88.51 \% \\
CIFAR-10 & CNN & 0.02 & 4.6 & 88.00 \& \\
CIFAR-10 & CNN & 0.04 & 17.0 & 86.76 \% \\
CIFAR-10 & CNN (width doubled) & 0.00 & 1.1 & 90.27 \% \\
CIFAR-10 & CNN (width doubled) & 0.005 & 1.8 & 90.16 \% \\
CIFAR-10 & CNN (width doubled) & 0.01  & 2.0 & 90.25 \% \\
CIFAR-10 & CNN (width doubled) & 0.02 & 3.7 & 89.65 \& \\
CIFAR-10 & CNN (width doubled) & 0.04 & 15.1 & 88.42 \%
\end{tabular}
\label{tab:performance_summary} 
\end{center}
\end{table}

This analysis highlights the interplay between reversal potential $|E_\text{rev}^\pm|$ and spike timing noise $\sigma_\text{spike}$, which could play a critical role in designing systems that integrate both models and hardware.
In Table \ref{tab:performance_summary}, we summarize the performance results. 
We show the best cases over the range of reversal potential values  $|E_\text{rev}^\pm|$.

\section{Effects of DSTD steps  in test phase}

\begin{figure*}
\centering
\includegraphics[clip, width=10cm]{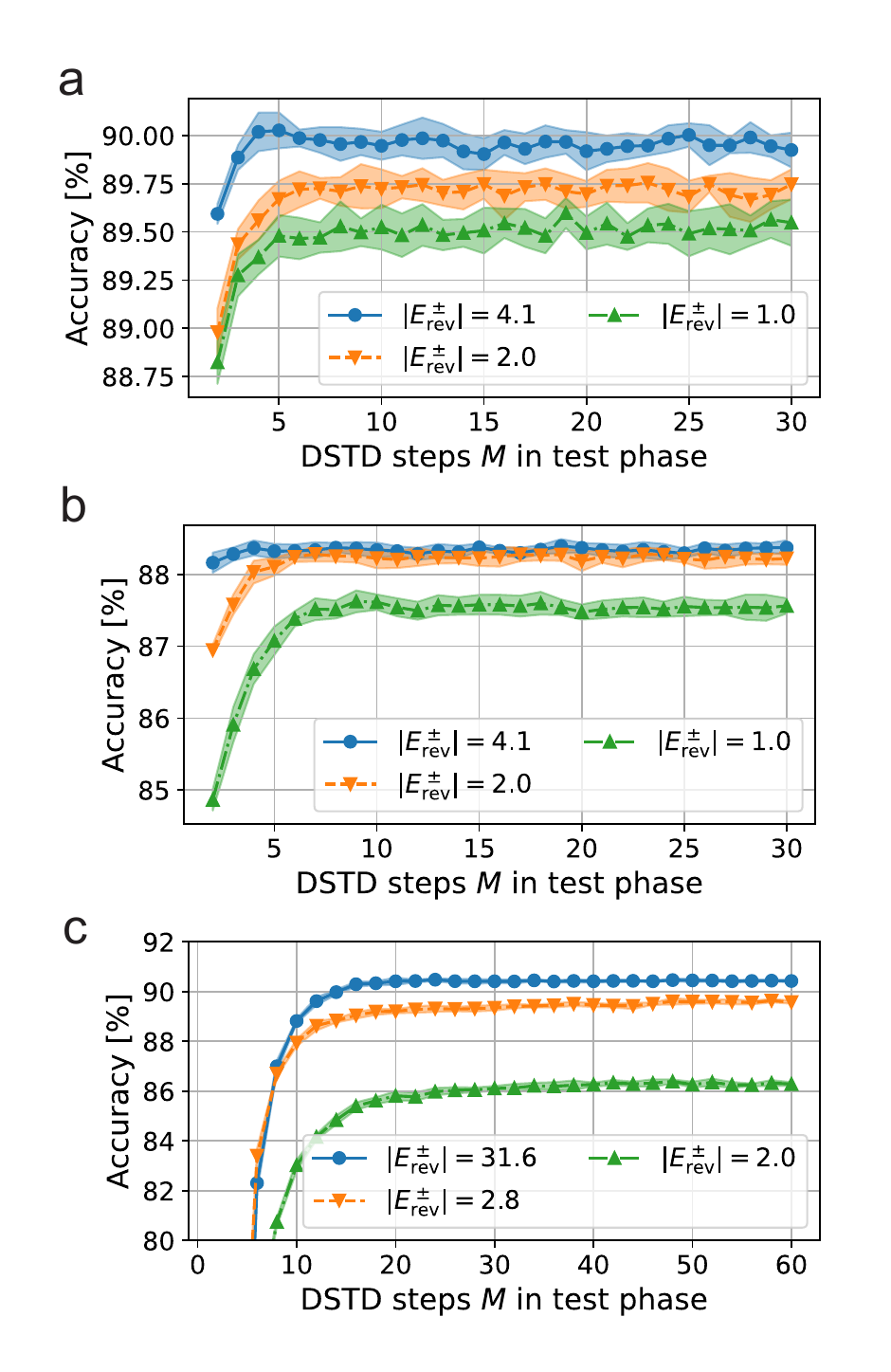}
\caption{{\bf Effect of DSTD steps $M$ during the test phase.} 
Each figure presents the variation in recognition accuracy as a function of the number of DSTD steps $M$ during the test phase using trained models. 
Specifically, a fully connected RC-Spike model (784-400-400-10) trained on the Fashion-MNIST dataset ({\bf a}), convolutional RC-Spike model trained on the CIFAR-10 dataset ({\bf b}), and convolutional TTFS-SNN model ({\bf c}) were evaluated. 
For the RC-Spike models, the number of steps during training was set to 15, while for the TTFS-SNN model, it was set to 30. 
In all cases, random offsets were applied during training. 
The error bars indicate standard deviations over 10 evaluations.
}
\label{fig:steps_in_test}
\end{figure*}

Neuron models incorporating reversal potentials impose high computational costs, making it challenging to evaluate such models with ODE solvers or exact solutions, even during the test phase. 
Consequently, we employed DSTD even for the test phase as well. 
To mitigate the impact of approximation errors introduced by DSTD, we set the number of DSTD steps $M$ to a sufficiently large value. Fig. \ref{fig:steps_in_test} illustrates the results of re-evaluating the trained model under different DSTD steps $M$ during the test phase. 
We selected $M=30$ for the RC-Spike model and $M=60$ for the TTFS-SNN model, both of which were found to be adequate for the convergence of recognition accuracy.

\section{Weight scaling optimization in test phase}

The positive reversal potential ($E_\text{rev}^+$) reduces the current as the membrane potential increases, while the negative reversal potential ($E_\text{rev}^-$) increases the current under the same conditions. 
This behavior suggests that, under a rough approximation, the reversal potentials modulate the  total effective current. 
Consequently, the impact of the reversal potentials 
may be partially mitigated without training a PNN model through weight scaling adjustments.


\begin{figure*}
\centering
\includegraphics[clip, width=\textwidth]{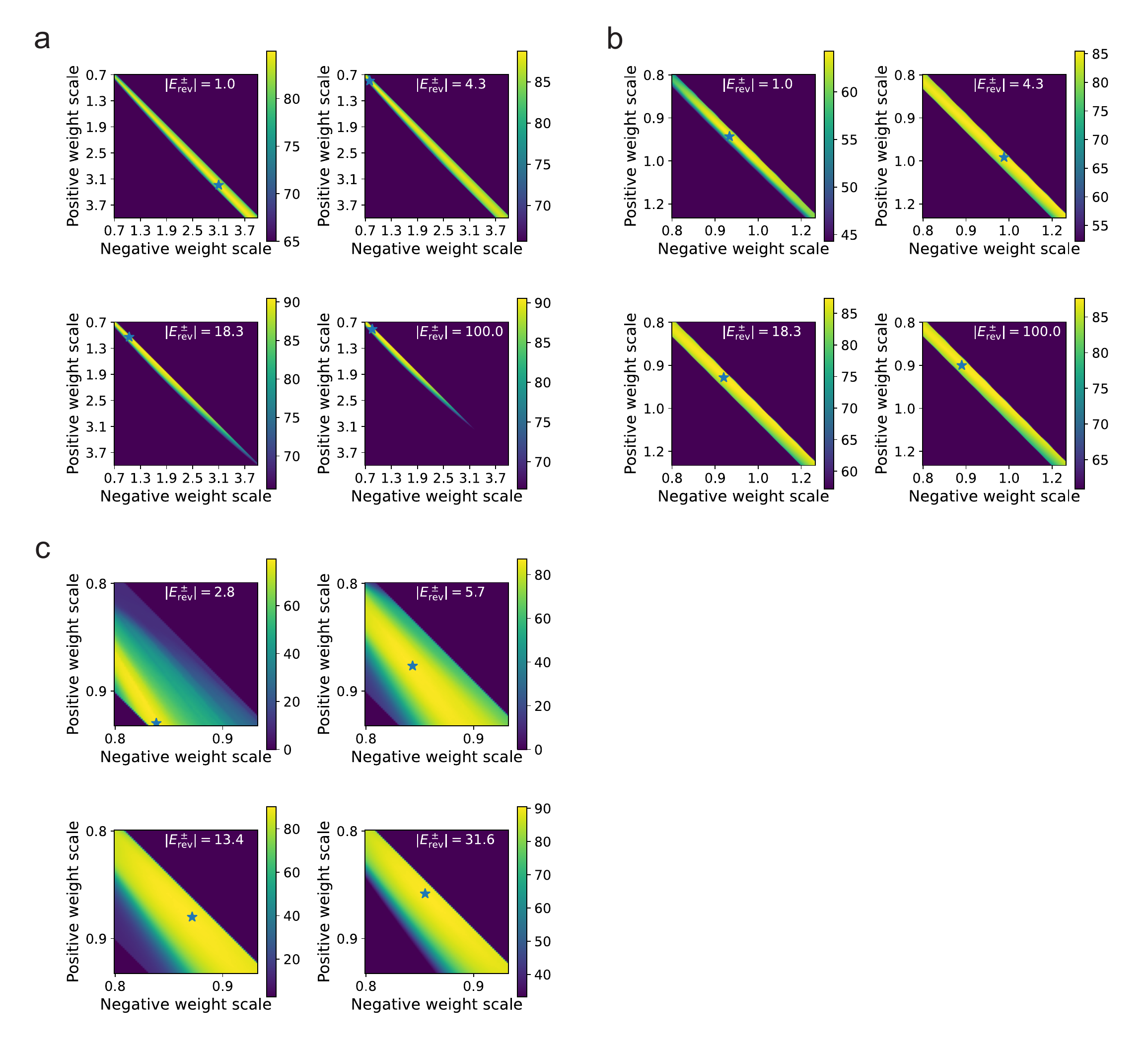}
\caption{
We present the results of optimizing the scaling of both positive and negative weights during the test phase for various values of $|E_\text{rev}^\pm|$. 
For the ANN models tested, $|E_\text{rev}^\pm|$ was set to 100 for the RC-Spike models and 31.6 for the TTFS-SNN models. 
The best scaling point is plotted with a blue star-shaped symbol. 
{\bf a.} Results for the RC-Spike model with two hidden layers using the Fashion-MNIST dataset. 
{\bf b.} Results for the RC-Spike model with a convolutional architecture using the CIFAR-10 dataset. 
{\bf c.} Results for the TTFS-SNN model using the Fashion-MNIST dataset. 
}
\label{fig:weight_scaling}
\end{figure*}

Figure \ref{fig:weight_scaling} presents the recognition performance obtained for specific values of reversal potentials during the test phase using the following ANN models.  
The ANN models were trained with $|E_\text{rev}^\pm|=100$ for the case of RC-Spike models and $|E_\text{rev}^\pm|=31.6$ for the case of TTFS-SNN models. 
The positive and negative weights of the network were uniformly scaled in increments of 0.1. 
In each plot, the scaling factor that achieved the highest recognition performance is highlighted with blue star-shaped symbols. 
These results demonstrate that, across all scenarios, weight scaling consistently enhances recognition performance.


\begin{figure*}
\centering
\includegraphics[clip, width=\textwidth]{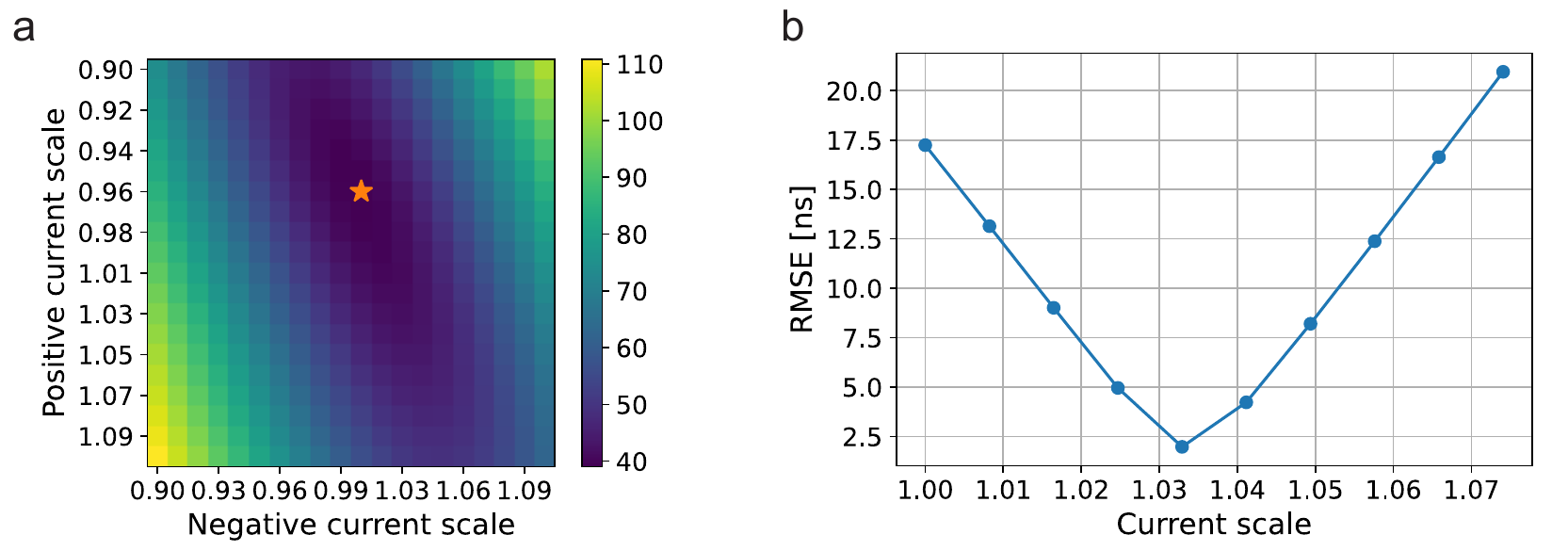}
\caption{{\bf a.} Optimization of current values during ANN-to-IMC mapping.
A grid search was conducted on positive and negative currents with a step size of 1\%. 
The evaluation metric is the root mean square error (RMSE) between the spike timing of the output layer from the RC-Spike model and that from  the SPICE simulation. 
The optimal point is marked by an orange star. 
{\bf b.} Optimization of current values during SNN-to-IMC mapping.
The absolute value of the current is scaled uniformly in approximately 1\% increments, without distinguishing between positive and negative currents. 
}
\label{fig:weight_scaling_HW}
\end{figure*}

To map model weights onto hardware, it is necessary to scale the current values on the hardware to mitigate the effects of parasitic capacitance inherent in IMC circuits. 
In addition , for the case of ANN-to-IMC mapping, the effects of $E_\text{rev}^\pm$ must be mitigated. 
Therefore, tor ANN-to-IMC mapping, both positive and negative currents were optimally scaled.  
By contrast, for SNN-to-IMC mapping, the absolute value of the current is scaled uniformly without distinguishing between positive and negative currents. 
The results of current scaling for ANN-to-IMC mapping are presented in Fig. \ref{fig:weight_scaling_HW} {\bf a}, while those for SNN-to-IMC mapping are shown in Fig. \ref{fig:weight_scaling_HW} {\bf b}.

\section{Circuit details}

The RC-Spike circuit introduced in this study was designed using the SkyWater technology node (sky130) processing design kit (PDK) \cite{sky130}. 
The sky130 process represents the 8th generation SONOS technology node, featuring a minimum transistor gate length of 150 nm and a five-layer metal interconnect system. 
We employed entirely open-source tools, with schematic design performed using xschem \cite{xschem}, layout design and parasitic extraction using magic \cite{magic}, layout-versus-schematic check using netgen \cite{netgen}, and circuit simulations using ngspice \cite{ngspice}. 
Note that all circuit simulations presented in this study are post-layout simulations carried out by extracting parasitic capacitances and resistances from the circuit layout \cite{tutorial_rcx}. 
The RC-Spike circuit comprises two major components: a synapse circuit and neuron circuit. 
Detailed descriptions of these circuits are provided in the following subsections.

\subsection{Synapse circuits}

\begin{figure*}
\centering
\includegraphics[clip, width=18cm]{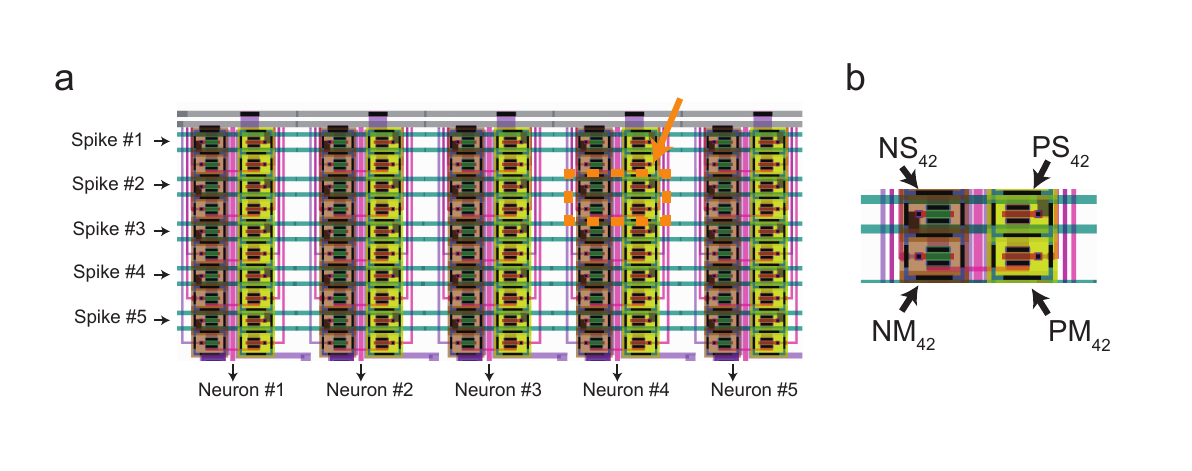}
\caption{{\bf a.} Layout of the array of synapse circuit blocks. In this schematic, five spike signals from the preceding layer are transferred from the left side. 
These input spike signals trigger the generation of synaptic currents, which are subsequently summed and directed toward the neuron circuit. 
{\bf b.} Enlarged view of a single synapse block, highlighted by orange dashed lines and arrows in {\bf a}. 
Each synapse block comprises two N-type MOSFETs (NMOS transistors) and two P-type MOSFETs (PMOS transistors). 
The $(i,j)$ synaptic block connects the $j$th spike signal to the $i$th neuron circuit, where NS$_{ij}$ and PS$_{ij}$ serve as the NMOS and PMOS selectors, respectively, 
and NM$_{ij}$ and PM$_{ij}$ function as the corresponding memory elements. 
The selectors activate upon receiving a spike signal, enabling current flow, while the memory elements modulate the magnitude of the current.
}
\label{fig:synapse_circuit}
\end{figure*}

Figure \ref{fig:synapse_circuit} {\bf a}  presents the synapse circuit layout result, where five spike signals (Spike \#1–5) originating from the preceding layer are input from the left. 
Each spike signal consists of two components: the original binary signal and an inverted binary signal. 
These signals propagate through the fourth metal layer (horizontal green lines). 
Upon receiving a spike signal, the synapse circuit generates a current that flows into the dendrites. 
The dendrites are implemented using the second metal layer (vertical pink lines) and are connected to the corresponding neuron circuits.
Figure \ref{fig:synapse_circuit} {\bf b} depicts a single synapse block, specifically the one enclosed by orange dashed lines and arrows in Fig. \ref{fig:synapse_circuit} {\bf a}. 
Each synapse block is composed of two N-type MOSFETs (NMOS) transistors and two P-type MOSFETs (PMOS) transistors. 
In the synapse block connecting Spike \#$j$ and Neuron \#$i$, NS$_{ij}$ represents the selector NMOS transistor, NM$_{ij}$ the memory NMOS transistor, PS$_{ij}$ the selector PMOS transistors, and PM$_{ij}$ the memory PMOS transistors. 
The selector NMOS and PMOS transistors are activated upon receiving a spike signal, allowing current to flow, while the memory NMOS and PMOS transistors modulate the magnitude of the current. 
The current magnitude of the memory MOSFETs is controlled by the bias voltage applied to their gate terminals; however, in this circuit, for simplicity, the bias voltage was explicitly applied in the simulation. 
In actual in-memory computing (IMC) circuits, SRAM circuits \cite{Yamaguchi2021energy} or floating-gate technology \cite{Bavandpour2019energy} may be employed to control the magnitude of currents. 
The MOSFETs used in each synapse block have a gate length of 250 nm and a gate width of 1 $\mu$m.

\begin{figure*}
\centering
\includegraphics[clip, width=16cm]{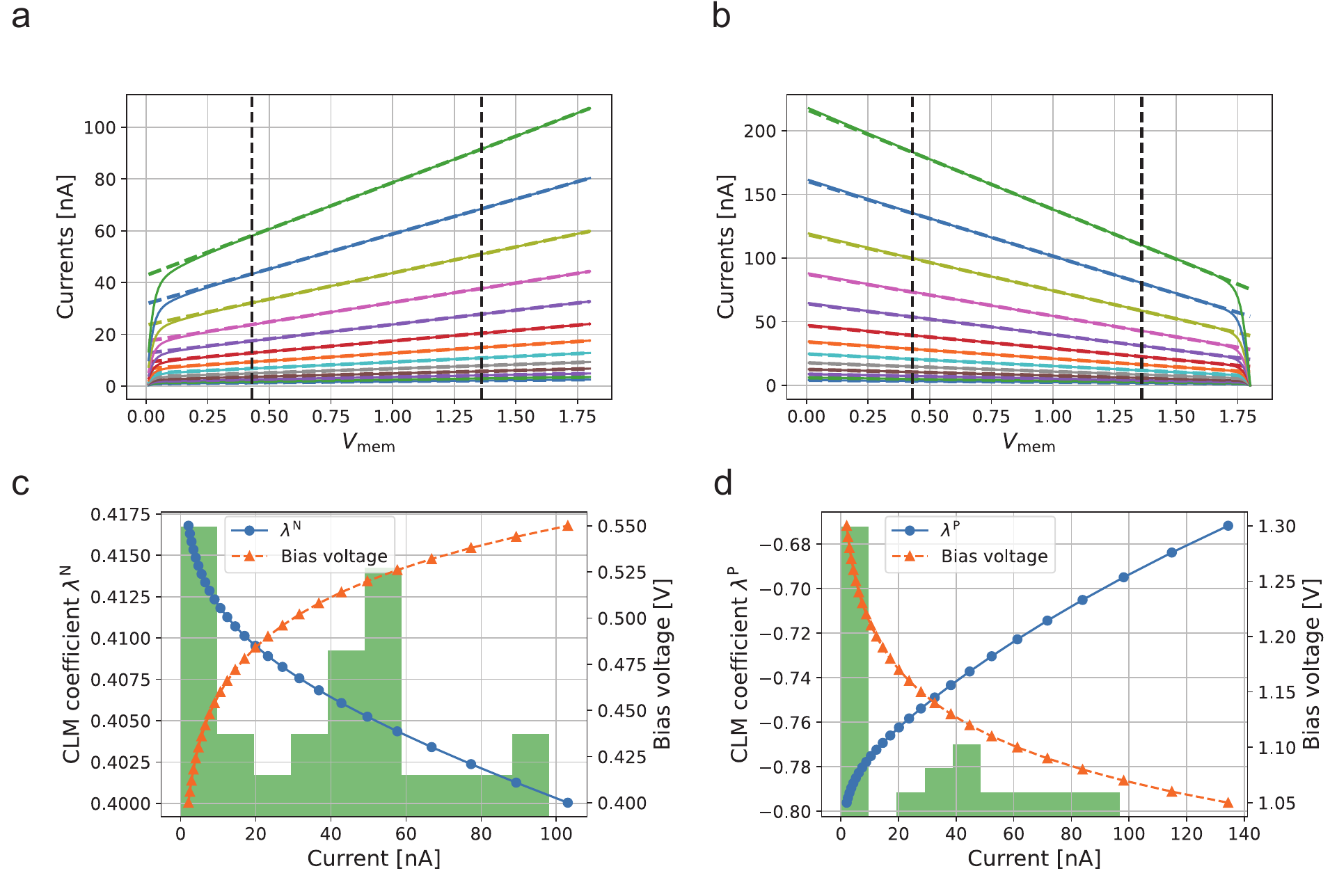}
\caption{{\bf Current characteristics of synapse circuits.}
{\bf a.} Dependence of drain current on drain voltage for an N-type MOSFET (NMOS) for gate voltages varying in 0.2 V increments from 0.4 V to 0.65 V.
{\bf b.} Dependence of drain current on drain voltage for a P-type MOSFET (PMOS) for gate voltages varying in 0.2 V increments from 1.3 V to 1.05 V.
In {\bf a} and {\bf b}, both NMOS and PMOS devices have a gate length of 250 nm and a gate width of 1 $\mu$m. 
The vertical black dashed lines in the figures indicate the fitting range of $[0.43\text{ V}, 1.36 \text{ V}]$, where a linear fit is applied to determine the non-ideal characteristic factor for each gate voltage. 
{\bf c.} Gate voltage dependence of the non-ideal characteristics. 
The horizontal axis represents the drain current, while the two vertical axes indicate the corresponding gate voltage and the non-ideal factor, $\lambda^\text{N}$, for the NMOS. 
Additionally, the green histogram illustrates the distribution of currents (corresponding to weights) across the entire network.
d. Same as {\bf c} but illustrates the non-ideal factor $\lambda^\text{P}$ for the PMOS. 
}
\label{fig:CLM}
\end{figure*}

Memory MOS transistors (NM$_{ij}$ and PM$_{ij}$) do not operate as an ideal current source, as the output current is dependent on the drain voltage, which corresponds to the membrane potential in our circuit. 
Figs. \ref{fig:CLM} {\bf a} and {\bf b} present the simulation results illustrating the membrane potential dependence of synaptic currents in memory MOS transistors under various gate voltages. 
Both NMOS and PMOS devices exhibit significant current variations in response to changes in membrane potential. 
This behavior is primarily attributed to channel-length modulation (CLM). 
The simulation demonstrates that the design achieves an excellent linear fit with the data for membrane potentials ranging from 0.46V to 1.36V. 
From this fit, the CLM coefficients, $\lambda^N$ and $\lambda^P$, are obtained (Eq. (\ref{eq:CLM_NMOS}) and Eq. (\ref{eq:CLM_PMOS})).  

Figs. \ref{fig:CLM} {\bf c} and {\bf d} show the relationship between synaptic current and the CLM coefficient at a resting (or initial) membrane potential of 1.3 V, under various gate voltage conditions. 
In the figure, the synaptic current is plotted along the horizontal axis, 
while the corresponding gate voltage and CLM coefficient are depicted on the vertical axis. 
While the CLM coefficient is dependent on the synaptic current at the resting membrane potential, the variation remains relatively small, approximately 5\%.
In this study, for the sake of simplicity, we have modeled the reversal potential as independent of the membrane potential. 
However, for more accurate modeling, it would be beneficial to introduce a reversal potential $E_\text{rev}^\pm(\cdot)$ that depends on the membrane potential.

\subsection{Neuron circuits}

\begin{figure*}
\centering
\includegraphics[clip, width=16cm]{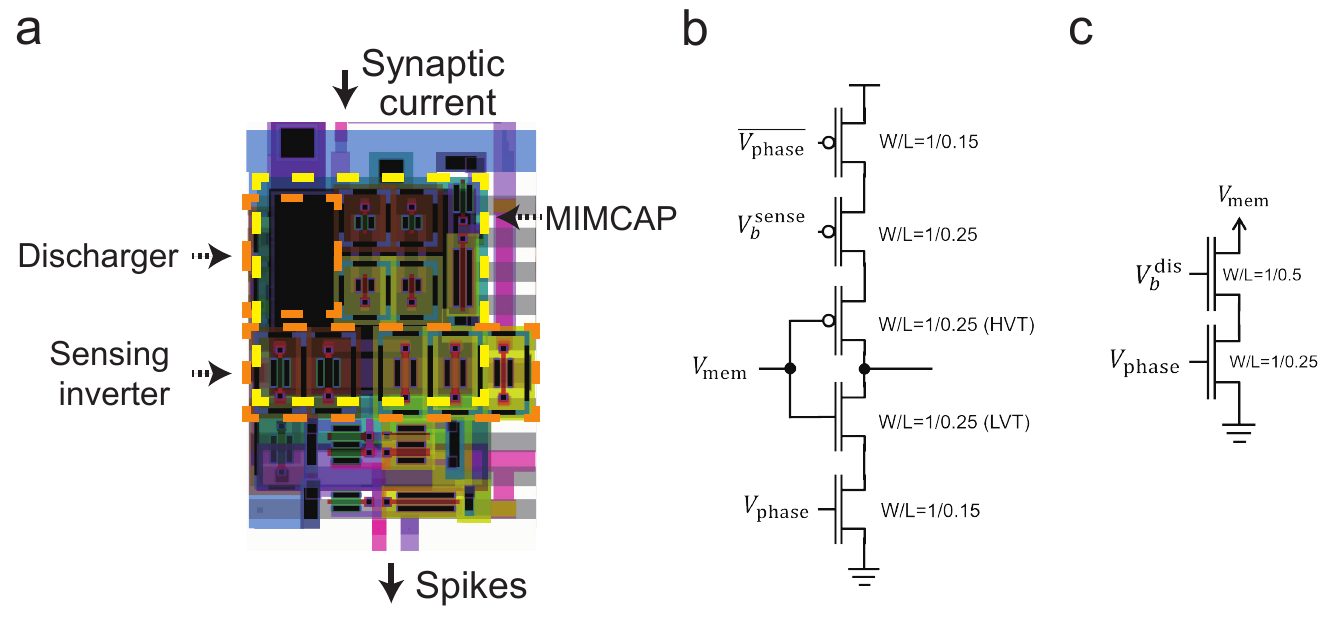}
\caption{{\bf a.} Neuron circuit layout. 
Synaptic currents are introduced from the top via the synapse circuit. 
The membrane capacitance is implemented using a metal-insulator-metal  capacitor (MIMCAP) between the fourth and fifth metal layers, with dimensions of 7 $\mu$m $\times$ 7 $\mu$m and a capacitance of 103.3 fF. 
The discharger circuit and sensing inverter circuit are outlined by the orange dashed lines. 
The discharger circuit regulates the discharge of the charge stored in the MIMCAP at a defined rate, 
while the sensing inverter detects when the membrane potential drops below the firing threshold. 
Digital control signals and bias signals for the neuron circuit propagate through the third metal layer (indicated by horizontal gray lines). 
{\bf b.} Schematic of sensing inverter circuit. 
To lower the firing threshold, an NMOS transistor with a low threshold voltage (LVT) is connected at the input, and a PMOS transistor with a high threshold voltage (HVT) is used. 
Additionally, a PMOS transistor connected to $V_b^\text{sense}$ allows the threshold voltage and current levels to be fine-tuned. MOSFETs controlled by the $V_\text{phase}$ signal and its inverted logic $\overline{V_\text{phase}}$ ensure that the sensing inverter is active only when the $V_\text{phase}$ signal is ON. 
{\bf c.} Schematic of discharger circuit. 
When the $V_\text{phase}$ signal is ON, the charge stored in the membrane capacitance is discharged to the ground. 
The discharge current is controlled by the $V_b^\text{dis}$ signal.
}
\label{fig:neuron_circuit}
\end{figure*}

Figure \ref{fig:neuron_circuit} {\bf a} presents the layout of the neuron circuit. 
The current from the synaptic circuit enters from the top and is stored in the metal-insulator-metal capacitor (MIMCAP), represented by the yellow dashed line in the figure. 
MIMCAP is implemented using the fourth and fifth metal layers, with a capacitance of 103.3 fF. 
Digital control signals and bias signals for the neuron circuit propagate along the third metal layer (denoted by the horizontal gray lines). In the MIMCAP configuration, the fourth metal layer is grounded, while the fifth metal layer corresponds to the membrane potential, effectively suppressing capacitive coupling between the digital signals and the metal line for the membrane potential.

The components enclosed by the orange dashed lines correspond to the sensing inverter and discharger circuits, with their respective schematics depicted in Fig. \ref{fig:neuron_circuit} {\bf b} and Fig. \ref{fig:neuron_circuit} {\bf c}, respectively. 
The sensing inverter is designed with a low firing threshold voltage, achieved by using high-threshold-voltage (HVT) PMOS transistors and low-threshold-voltage (LVT) NMOS transistors. It is important to note that the membrane potential is inverted, and in this circuit, the firing threshold voltage is set lower than the resting membrane potential. 
Furthermore, by applying the bias voltage $V_b^\text{sense}$ to the PMOS transistor, the firing threshold voltage can be reduced even further, while simultaneously minimizing the leakage current. 
This circuit is active only when the $V_\text{phase}$ signal is in the ON state. 

The discharger circuit discharges the charge accumulated in MIMCAP to the ground line when the $V_\text{phase}$ signal is ON. 
The discharge current is controlled by an NMOS transistor, which is biased by the voltage $V_b^\text{dis}$. 
This NMOS transistor also plays a role in suppressing capacitive coupling between the $V_\text{phase}$ signal and the metal line for the membrane potential.

\begin{figure*}
\centering
\includegraphics[clip, width=16cm]{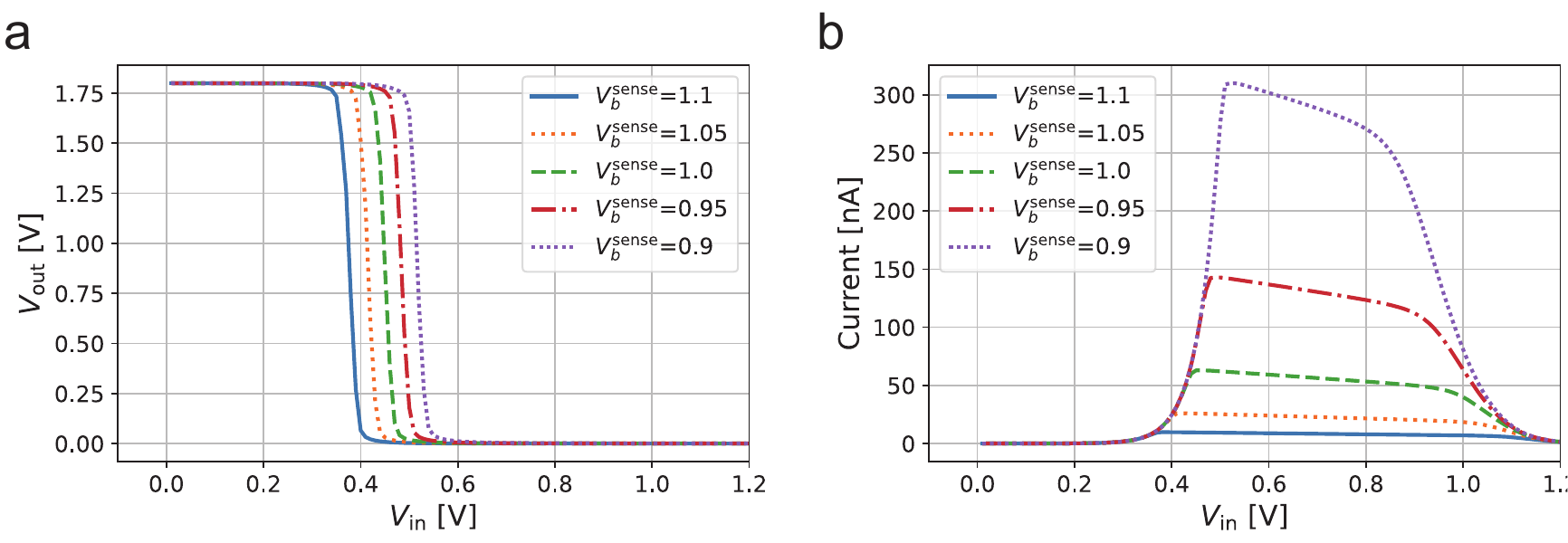}
\caption{{\bf a.} Input-output voltage characteristics of the sensing inverter circuit under various bias voltages $V_b^\text{sense}$. 
{\bf b.} Power dissipation characteristics of the sensing inverter circuit under various bias voltages $V_b^\text{sense}$.
}
\label{fig:sensor_sim}
\end{figure*}

Fig. \ref{fig:sensor_sim} {\bf a} presents the input-output characteristics of the sensing inverter under various bias voltages $V_b^\text{sense}$. 
As described earlier, the bias voltage $V_b^\text{sense}$ modulates the firing threshold. 
Fig. \ref{fig:sensor_sim} {\bf b} depicts the variation in through-current levels as a function of $V_b^\text{sense}$. 
A significant reduction in through-current led to a slower rise in the output voltage of the sensing inverter, which, in turn, increased the total amount of through-current of the subsequent NAND gate. 
To mitigate excessive power dissipation in the subsequent NAND gate, the $V_b^\text{sense}$ bias voltage was set to 0.95 V.

\begin{figure*}
\centering
\includegraphics[clip, width=16cm]{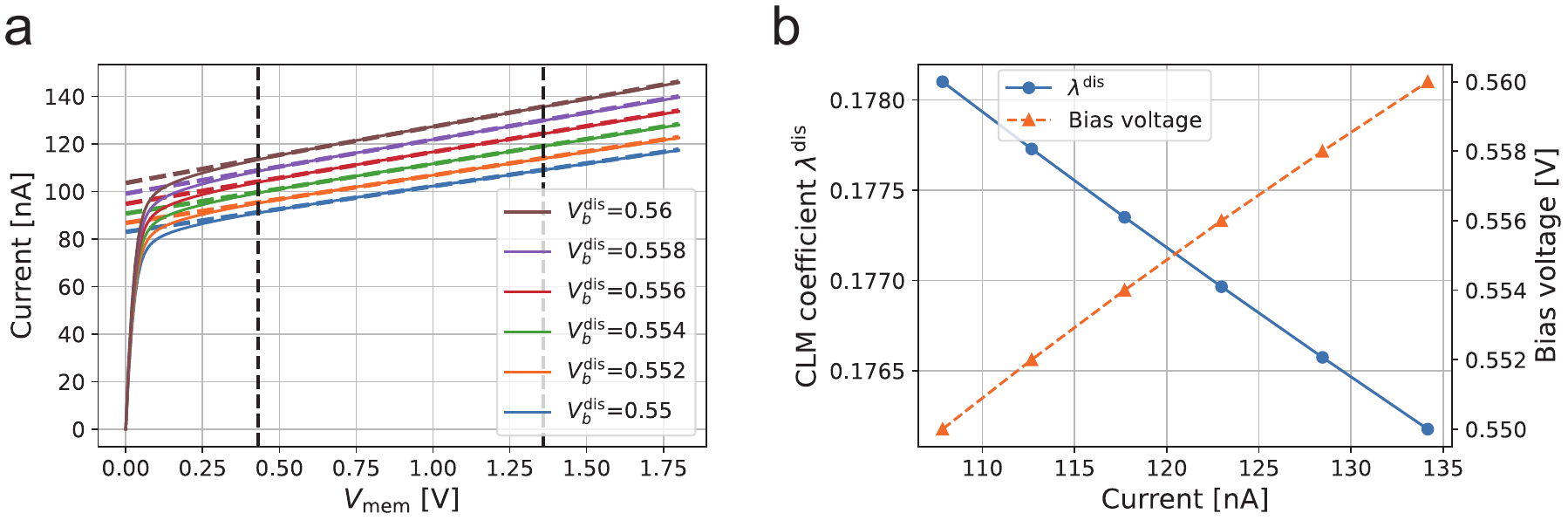}
\caption{{\bf a.} Dependence of the discharger circuit current on membrane potential under various bias voltages of $V_b^\text{dis}$. 
The vertical black dashed lines in the figures indicate the fitting range of $[0.43\text{ V}, 1.36 \text{ V}]$, where a linear fit is applied to determine the non-ideal characteristic factor $\lambda^\text{dis}$ for each gate voltage. 
{\bf b.} Dependence of the CLM coefficient on current in the discharger circuit.
The horizontal axis represents the drain current, while the two vertical axes indicate the corresponding gate voltage and the non-ideal factor, $\lambda^\text{dis}$, for the NMOS. 
}
\label{fig:charger}
\end{figure*}

The discharger circuit utilizes an NMOS as a current source, leading to a current output that is dependent on the membrane potential.
Fig. \ref{fig:charger} {\bf a} illustrates the membrane potential dependence of the current in the discharger circuit. 
Similar to the behavior observed in memory cells, a nearly perfect linear fit can be achieved within the potential range of 0.46 to 1.36 V.
Fig. \ref{fig:charger} {\bf b} presents the CLM coefficient $\lambda^\text{dis}$ corresponding to the current at the resting membrane potential.
The bias voltage $V_b^\text{dis}$ was configured at 0.554 V, which ensures that no firing occurs during the firing phase in the absence of input spikes.

\bibliographystyle{MSP}
\bibliography{myBib}

\end{document}